\documentclass[opre,nonblindrev]{informs3} % current default for manuscript submission

\DoubleSpacedXI

\usepackage{natbib}
 \bibpunct[, ]{(}{)}{,}{a}{}{,}%
\TheoremsNumberedThrough     % Preferred (Theorem 1, Lemma 1, Theorem 2)
\ECRepeatTheorems

\EquationsNumberedThrough    % Default: (1), (2), ...
\MANUSCRIPTNO{}

\usepackage{psfrag,ifthen,multirow,mathrsfs}
\usepackage{epsfig,psfrag,subcaption,float,texnansi,color,tikz,ifthen,array}
\usepackage{algorithm}
\usepackage{algpseudocode}
\usepackage{microtype}
\usepackage{graphicx}
\usepackage{latexsym,url,afterpage,enumerate,lmodern}
\usepackage[normalem]{ulem}
\usepackage{amsmath,amssymb,amsfonts}%amsthm
\usepackage[margin=10pt,font=small,labelfont=bf]{caption}
\usepackage{booktabs,xcolor}
\usepackage{IEEEtrantools}
\usepackage{bm}
\usepackage{enumitem}
\usepackage{tabularx}
\usepackage{dsfont}
\usepackage{rotating}
\usepackage{lscape}
\usepackage{wrapfig}
\usepackage{mathtools}
\usepackage[hidelinks]{hyperref}

\DeclarePairedDelimiter\floor{\lfloor}{\rfloor}

\newcommand{\field}[1]{\ensuremath{\mathbb{#1}}}
\newcommand{\sets}[1]{\ensuremath{\mathcal{#1}}}

\newcommand{\naturals}{\ensuremath{\field{N}}} % natural numbers
\newcommand{\R}{\ensuremath{\mathbb{R}}} % expectation

\newcommand{\conv}{\ensuremath{\text{conv}}}

\newcommand{\newsa}[1]{{\color{magenta} #1}}
\newcommand{\revision}[1]{{\color{black} #1}}

\newcommand{\flow}{\texttt{FlowOCT}} 
 
\newcommand{\benders}{\texttt{BendersOCT}} 
\newcommand{\bendersMIPSolAlg}{\texttt{BendersOCT-MIPSol-Alg2}}
\newcommand{\bendersMIPSolLP}{\texttt{BendersOCT-MIPSol-LP}} 
\newcommand{\bendersAllSolLP}{\texttt{BendersOCT-AllSol-LP}} 
\newcommand{\FlowOCTworst}{\texttt{FlowOCT-worst}} 

\newcommand{\bendersFiveBuckets}{\texttt{BendersOCT-5}} 
\newcommand{\bendersTenBuckets}{\texttt{BendersOCT-10}} 
\begin{document}

%\setcounter{page}{1}
%\MONTH{00}% (month or a similar seasonal id)
%\YEAR{2020}% e.g., 2005
%\SHORTYEAR{20}% shortened year (two-digit)
%\DOI{10.1287/xxxx.0000.0000}%

\RUNAUTHOR{Aghaei, G\'omez, and Vayanos}

\RUNTITLE{Strong Optimal Classification Trees}

% \TITLE{Bundling Information Goods of Decreasing Value}
\TITLE{Strong Optimal Classification Trees}

\ARTICLEAUTHORS{%
\AUTHOR{Sina Aghaei}
\AFF{Center for Artificial Intelligence in Society, University of Southern California, Los Angeles, CA 90089, USA, \EMAIL{saghaei@usc.edu}} %, \URL{}}
\AUTHOR{Andr\'es G\'omez }
\AFF{Department of Industrial and Systems Engineering, Los Angeles, CA 90089, USA, \EMAIL{gomezand@usc.edu}}
\AUTHOR{Phebe Vayanos}
\AFF{Center for Artificial Intelligence in Society, University of Southern California, Los Angeles, CA 90089, USA, \EMAIL{phebe.vayanos@usc.edu}} %, \URL{}}
} % end of the block

\ABSTRACT{
Decision trees are among the most popular machine learning models and are used routinely in applications ranging from revenue management and medicine to bioinformatics. In this paper, we consider the problem of learning \emph{optimal} binary classification trees \revision{with univariate splits}. Literature on the topic has burgeoned in recent years, motivated both by the empirical suboptimality of heuristic approaches and the tremendous improvements in mixed-integer optimization (MIO) technology. Yet, existing MIO-based approaches from the literature do not leverage the power of MIO to its full extent: they rely on \emph{weak} formulations, resulting in slow convergence and large optimality gaps. To fill this gap in the literature, we propose an intuitive flow-based MIO formulation for learning optimal binary classification trees.
Our formulation can accommodate side constraints to enable the design of interpretable and fair decision trees. Moreover, we show that our formulation has a \emph{stronger} linear optimization relaxation than existing methods \revision{in the case of binary data}. We exploit the decomposable structure of our formulation and max-flow/min-cut duality to derive a Benders' decomposition method to speed-up computation. We propose a tailored procedure for solving each decomposed subproblem that provably generates \emph{facets} of the feasible set of the MIO as constraints to add to the main problem. We conduct extensive computational experiments on standard benchmark datasets on which we show that our proposed approaches are\revision{~{29} times} faster than {state-of-the-art} MIO-based techniques and improve {out-of-sample} performance by up to~{8}\%. 
% We have developed an R package for our methods which can be found online at {\url{https://github.com/pashew94/StrongTree}} along with instructions and is freely distributed for academic and non-profit use.% doxygen instructions
}


\maketitle

%%%%%%%%%%%%%%%%%%%%%%%%%%%%%%%%%%%%%%%%%%%%%%%%%%
%%%%%%%%%%%%%%%%%%%%%%%%%%%%%%%%%%%%%%%%%%%%%%%%%%

%%%%%%%%%%%%%%%%%%%%%%%%%%%%%%%%%%%%%%%%%%%%%%%%%%
%%%%%%%%%%%%%%%%%%%%%%%%%%%%%%%%%%%%%%%%%%%%%%%%%%
\section{Introduction}
\label{sec:Introduction}
%%%%%%%%%%%%%%%%%%%%%%%%%%%%%%%%%%%%%%%%%%%%%%%%%%
%%%%%%%%%%%%%%%%%%%%%%%%%%%%%%%%%%%%%%%%%%%%%%%%%%

%%%%%%%%%%%%%%%%%%%%%%%%%%%%%%%%%%%%%%%%%%%%%%%%%%
\subsection{Motivation \& Related Work}
%%%%%%%%%%%%%%%%%%%%%%%%%%%%%%%%%%%%%%%%%%%%%%%%%%

Since their inception over 30 years ago, see~\citet{breiman1984classification}, decision trees have become among the most popular techniques for interpretable machine learning \revision{(ML)}{, see~\citet{rudin2019stop}}. 
\revision{Typically, a} decision tree takes the form of a binary tree.
% \revision{In this paper, we are interested in learning decision trees that take the form of a \emph{binary tree}, which is commonly found in literature, see~\citet{carrizosa2021mathematical}.} 
In each \emph{branching} node of the tree, a binary test is performed on a specific feature. Two branches emanate from each branching node, with each branch representing the outcome of the test. If a datapoint passes (resp.\ fails) the test, it is directed to the left (resp.\ right) branch. A predicted label is assigned to all \emph{leaf} nodes. Thus, each path from root to leaf represents a classification rule that assigns a unique label to all datapoints that reach that leaf. The goal in the design of optimal decision trees is to select the tests to perform at each branching node and the labels to assign to each leaf to maximize prediction accuracy (classification) or to minimize prediction error (regression). {In practice, shallow trees are preferred as they are easier to understand and interpret. Thus, we focus on designing trees of bounded depth.}
Not only are decision trees popular in their own right; they also form the backbone for more sophisticated machine learning models. For example, they are the building blocks for ensemble methods (such as random forests) which combine several decision trees and constitute some of the most popular and stable machine learning techniques available, see e.g.,~\citet{breiman1996bagging, breiman2001random} and~\citet{liaw2002classification}. They have also proved useful to provide explanations for the solutions to optimization problems, see e.g.,~\citet{bertsimas2018voice}.

Decision trees are used routinely in applications ranging from revenue management to chemical engineering, medicine, and bioinformatics. For example, they are used for the management of substance-abusing psychiatric patients~\citep{intrator1992decision}, to manage Parkinson's disease~\citep{olanow2001algorithm}, to predict a mortality on liver transplant waitlists~\citep{bertsimas2019development}, to learn chemical concepts such as octane number and molecular substructures~\citep{blurock1995automatic}, and to evaluate housing systems for homeless youth~\citep{chan2018empirical}. Moreover, tree ensembles can be used to reveal associations between micro RNAs and human diseases~\citep{chen2019ensemble} and to predict outcomes in antibody incompatible kidney transplantation~\citep{shaikhina2019decision}.

The problem of learning optimal decision trees is an $\mathcal{NP}$-hard problem, see~\citet{hyafil1976constructing} and~\citet{breiman1984classification}. It can intuitively be viewed as a combinatorial optimization problem with an exponential number of decision variables: at each branching node of the tree, one can select which feature to branch on (and potentially the level of that feature), guiding each datapoint to the left or right using logical constraints.

\paragraph{Traditional Methods.} Motivated by these hardness results, traditional algorithms for learning decision trees have relied on heuristics that employ very intuitive, yet ad-hoc, rules for constructing the decision trees. For example, CART uses the Gini Index to decide on the splitting, see~\citet{breiman1984classification}; ID3 employs entropy, see~\citet{quinlan1986induction}; and C4.5 leverages normalized information gain, see~\citet{quinlan2014c4}. The high quality and speed of these algorithms combined with the availability of software packages in many popular languages such as \texttt{R} or \texttt{Python} has facilitated their popularization, see e.g.,~\citet{therneau2015package} and~\citet{kuhn2018package}.

\paragraph{Mathematical Optimization Techniques.}  Motivated by the heuristic nature of traditional approaches which provide no guarantees on the quality of the learned tree, several researchers have proposed algorithms for learning provably \emph{optimal} trees based on techniques from mathematical optimization. Approaches for learning optimal decision trees rely on enumeration coupled with rules to prune-out the search space. For example, \citet{nijssen2010optimal} use itemset mining algorithms and~\citet{narodytska2018learning} use satisfiability (SAT) solvers. \citet{verhaeghe2019learning} propose a more elaborate implementation combining several ideas from the literature, including branch-and-bound, itemset mining techniques, and caching. \citet{hu2019optimal} use analytical bounds (to aggressively prune-out the search space) combined with a tailored bit-vector based implementation. {\citet{lin2020generalized} extend the approach of~\citet{hu2019optimal} to produce optimal decision trees over a variety of objectives such as F-score and area under the receiver operating characteristic curve (AUROC)}. \citet{demirovic2020murtree} and \citet{demirovic2020optimal} use dynamic programming and \citet{nijssen2020learning} use caching branch-and-bound search to compute optimal decision trees.
% {~\citet{bertsimas2019machine} propose a heuristic local search algorithm, based on coordinate descent, to find optimal or near-optimal solutions in practical times. Despite their great performance in practice, they are not able to provide any optimality guarantee for their solutions.
{\citet{blanquero2021optimal} propose a continuous-based randomized approach for learning optimal classification trees with oblique cuts. }
Incorporating constraints into the above approaches is usually a challenging task, see e.g., 
\citet{detassis2020teaching}.

\paragraph{The Special Case of MIO.} As an alternative approach to conducting the search, \citet{bertsimas2017optimal} recently proposed to use mixed-integer optimization (MIO) to learn optimal classification trees. Following this work, using MIO to learn decision trees gained a lot of traction in the literature with the works of~\citet{gunluk2018optimal}, \citet{aghaei2019learning}, and~\citet{verwer2019learning}. This is no coincidence. First, there exists a suite of MIO {off-the-shelf} solvers and algorithms that can be leveraged to effectively prune-out the search space. Indeed, solvers such as \citet{cplex2009v12} and \citet{gurobi2015gurobi}  have benefited from decades of research, see~\citet{bixby2012brief}, and have been very successful at solving broad classes of MIO problems. Second, MIO comes with a highly expressive language that can be used to tailor the objective function of the problem or to augment the learning problem with constraints of practical interest. For example, \citet{aghaei2019learning} leverage the power of MIO to learn fair and interpretable classification and regression trees by augmenting their model with additional constraints. They also show how MIO technology can be exploited to learn decision trees with more sophisticated structure (linear branching and leafing rules). Similarly, \citet{gunluk2018optimal} use MIO {to solve the problem of learning classification trees} by taking into account the special structure of categorical features and allowing combinatorial decisions (based on subsets of values of features) at each node.
MIO formulations have also been leveraged to design decision trees for decision- and policy-making problems, see e.g.,~\citet{azizi2018designing} and \citet{ciocan2018interpretable}, for optimizing decisions over tree ensembles, see~\citet{mivsic2017optimization} and~\citet{biggs2018optimizing}{, and also for developing heuristic approaches for learning classification trees, see~\citet{bertsimas2019machine}}. {We refer the interested reader to the paper of \citet{carrizosa2021mathematical} for an in-depth review of the field.}

%%%%%%%%%%%%%%%%%%%%%%%%%%%%%%%%%%%%%%%%%%%%%%%%%%
\subsection{Discussion}  %\& Motivation
%%%%%%%%%%%%%%%%%%%%%%%%%%%%%%%%%%%%%%%%%%%%%%%%%%

The works of \citet{bertsimas2017optimal}, \citet{aghaei2019learning}, and \citet{verwer2019learning} have served to showcase the modeling power of using MIO to learn decision trees and the potential suboptimality of traditional algorithms. Yet, we argue that they have not leveraged the power of MIO to its full extent. 

A critical component for efficiently solving MIOs is to pose good formulations, but determining such formulations is no simple task. The standard approach for solving MIO problems is the branch-and-bound method, which partitions the search space recursively and solves Linear Optimization (LO) relaxations for each partition to {produce bounds} for fathoming sections of the search space. Thus, since solving an MIO requires solving a large sequence of LO problems, small and compact formulations are desirable as they enable the LO relaxation to be solved faster. Moreover, formulations with tight LO relaxations, referred to as \emph{strong} formulations, are also desirable as they produce higher {quality bounds} which lead to a faster pruning of the search space, ultimately reducing the number of LO problems to be solved. Indeed, a recent research thrust focuses on devising strong formulations for inference problems \revision{\citep{dong2015regularization,atamturk2018sparse,bienstock2018principled,xie2018scalable,gomez2019outlier,anderson2020strong,bertsimas2020sparse,hazimeh2020sparse}}. Unfortunately, these two goals are at odds with one another: stronger formulations often require more variables and constraints than \emph{weak} ones. For example, in the context of decision trees, \citet{verwer2019learning} propose an MIO formulation with significantly fewer variables and constraints than the formulation of \citet{bertsimas2017optimal}, but in the process weaken the LO relaxation. As a consequence, neither method consistently outperforms the other. 

We note that in the case of MIO problems with large numbers of decision variables and constraints, classical decomposition techniques from the Operations Research literature may be leveraged to break the problem up into multiple tractable subproblems of benign complexity, see e.g., \citet{gade2014decomposition,liu2016decomposition,guo2019logic,macneil2019integer,gangammanavar2020stochastic,liu2020asymptotic}. A notable example of a decomposition algorithm is  Benders' decomposition, see \citet{benders1962partitioning}. This decomposition approach exploits the structure of mathematical optimization problems with so-called \emph{complicating variables} which couple constraints with one another and which, once fixed, result in an attractive decomposable structure that is leveraged to speed-up computation and alleviate memory consumption, allowing the solution of large-scale MIO problems. 

To the best of our knowledge, existing approaches from the literature have not sought explicitly strong formulations, neither have they attempted to leverage the potentially decomposable structure of the problem. This is precisely the gap we fill with the present work.

%%%%%%%%%%%%%%%%%%%%%%%%%%%%%%%%%%%%%%%%%%%%%%%%%%
\subsection{Proposed Approach \& Contributions}
%%%%%%%%%%%%%%%%%%%%%%%%%%%%%%%%%%%%%%%%%%%%%%%%%%

%\paragraph{Proposed Approach \& Contributions. }
Our approach and main contributions in this paper are:
\begin{enumerate}[label=\emph{(\alph*)}]\setlength\itemsep{0em}
    \item We propose a flow-based MIO formulation for learning optimal classification trees with binary features. In this model, correctly classified datapoints can be seen as \emph{flowing} from the root of the tree to a suitable leaf while incorrectly classified datapoints are not allowed to flow through the tree.  Our formulation can easily be augmented with constraints (e.g., imposing fairness), regularization penalties and conveniently be adjusted to cater for imbalanced datasets.
    \item  We demonstrate that our proposed formulation has a stronger LO relaxation than existing alternatives. Notably, it does not involve big-$M$ constraints. It is also amenable to Benders' decomposition. In particular, binary tests are selected in the main problem and each subproblem guides each datapoint through the tree via a max-flow subproblem. We leverage the max-flow structure of the subproblems to solve them efficiently via a tailored min-cut procedure. Moreover, we prove that all the constraints generated by this Benders' procedure are facet-defining{, i.e., they are required to describe the (convex hull of the) projection of the feasible region into the space of variables appearing in the main problem.}
    \item We conduct extensive computational studies on benchmark datasets from the literature, showing that our formulations improve upon the state-of-the-art MIO algorithms, both in terms of in-sample solution quality (and speed) and out-of-sample performance.
\end{enumerate}
The proposed modeling and solution paradigm can act as a building block for the faster and more accurate learning of more sophisticated trees and tree ensembles. %Continuous data can be discretized and binarized to address problems with continuous labels, see \citet{breiman2017classification}. %Fairness and interpretability constraints can naturally be incorporated into the problem, see~\citet{aghaei2019learning}. \notepv{refer to what we do for imbalanced data here withe any necessary refs?}

The rest of the paper is organized as follows. We introduce our flow-based formulation and our Benders' decomposition method in Section \ref{sec:DT_Formulation} and Section \ref{sec:Benders}, respectively. Several generalizations of our core formulation are discussed in Section \ref{sec:Generalization}. We report on computational experiments with popular benchmark datasets in Section \ref{sec:Experiments}. Most proofs and detailed computational results are provided in the electronic companion.

%%%%%%%%%%%%%%%%%%%%%%%%%%%%%%%%%%%%%%%%%%%%%%%%%%
%%%%%%%%%%%%%%%%%%%%%%%%%%%%%%%%%%%%%%%%%%%%%%%%%%
 \section{Learning Balanced Classification Trees}
\label{sec:DT_Formulation}
%%%%%%%%%%%%%%%%%%%%%%%%%%%%%%%%%%%%%%%%%%%%%%%%%%
%%%%%%%%%%%%%%%%%%%%%%%%%%%%%%%%%%%%%%%%%%%%%%%%%%

In this section, we describe our MIO formulation for learning  optimal \emph{balanced} classification trees of a given depth, i.e., trees wherein the distance between all nodes where a prediction is made and the root node is equal to the tree depth. Our MIO formulation relies on the observation that once the structure of the tree is fixed, determining whether a datapoint is correctly classified or not reduces to checking whether the datapoint can, based on its features and label, flow from the root of the tree to a leaf where the prediction made matches its label. Thus, we begin this section by formally defining balanced trees and their associated \emph{flow graph} in Section~\ref{sec:DTandFlowGraph}, before introducing our proposed flow-based formulation in Section~\ref{sec:DTFlowFormulation}. We discuss variants of this basic model that can be used to learn sparse, possibly imbalanced, trees in Section~\ref{sec:Generalization}. \revision{Throughout the paper, we represent vectors using bold fonts and sets using capital calligraphic fonts.}

%%%%%%%%%%%%%%%%%%%%%%%%%%%%%%%%%%%%%%%%%%%%%%%%%%
\subsection{Decision Tree and Associated Flow Graph}
\label{sec:DTandFlowGraph}
%%%%%%%%%%%%%%%%%%%%%%%%%%%%%%%%%%%%%%%%%%%%%%%%%%

A key step towards our flow-based MIO formulation of the problem consists in converting the decision tree of fixed depth that we wish to train to a directed acyclic graph where all arcs are directed from the root of the tree to the leaves. We detail this conversion together with the basic terminology that we use in our paper in what follows.

\begin{definition}[Balanced Decision Tree] \label{def:balanced_trees} 
  A balanced decision tree of depth $d \in \naturals$ is a perfect binary tree, i.e., a binary tree in which all interior nodes have two children and all leaves have the same depth. We number the nodes in the tree in the same order they appear in the breadth-first search traverse such that 1 is the root node and $2^{d+1}-1$ is the bottom right node. We define the set of first~$2^{d}-1$ nodes $\sets B:=\{1,\ldots,2^d-1\}$ as \emph{branching nodes} and the remaining $2^d$ nodes $\sets L:=\{2^d,\ldots, 2^{d+1}-1\}$ as the \emph{leaf nodes} of the decision tree.
\end{definition}

\begin{figure}[t!]
%\vskip 0.2in
\begin{center}
\centerline{
\includegraphics[width=0.35\textwidth]{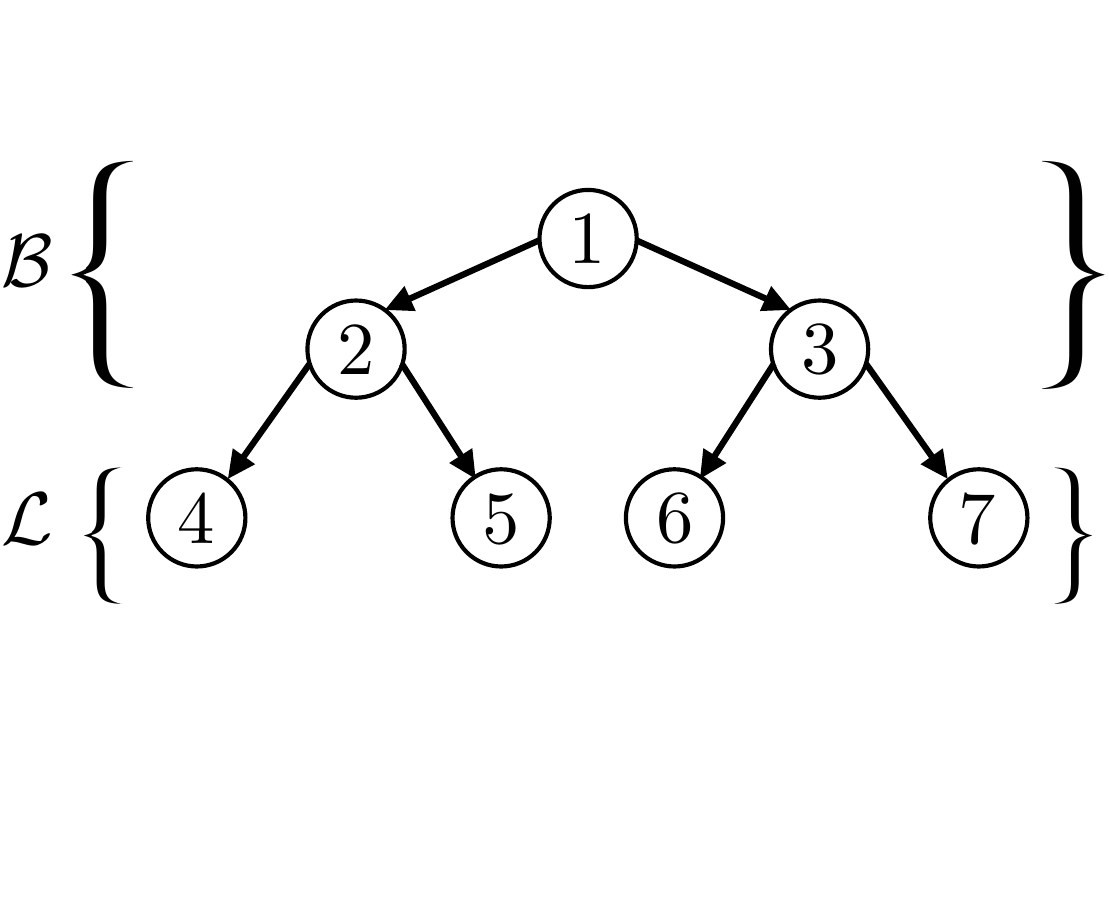}
\hspace{2cm}
\includegraphics[width=0.35\textwidth]{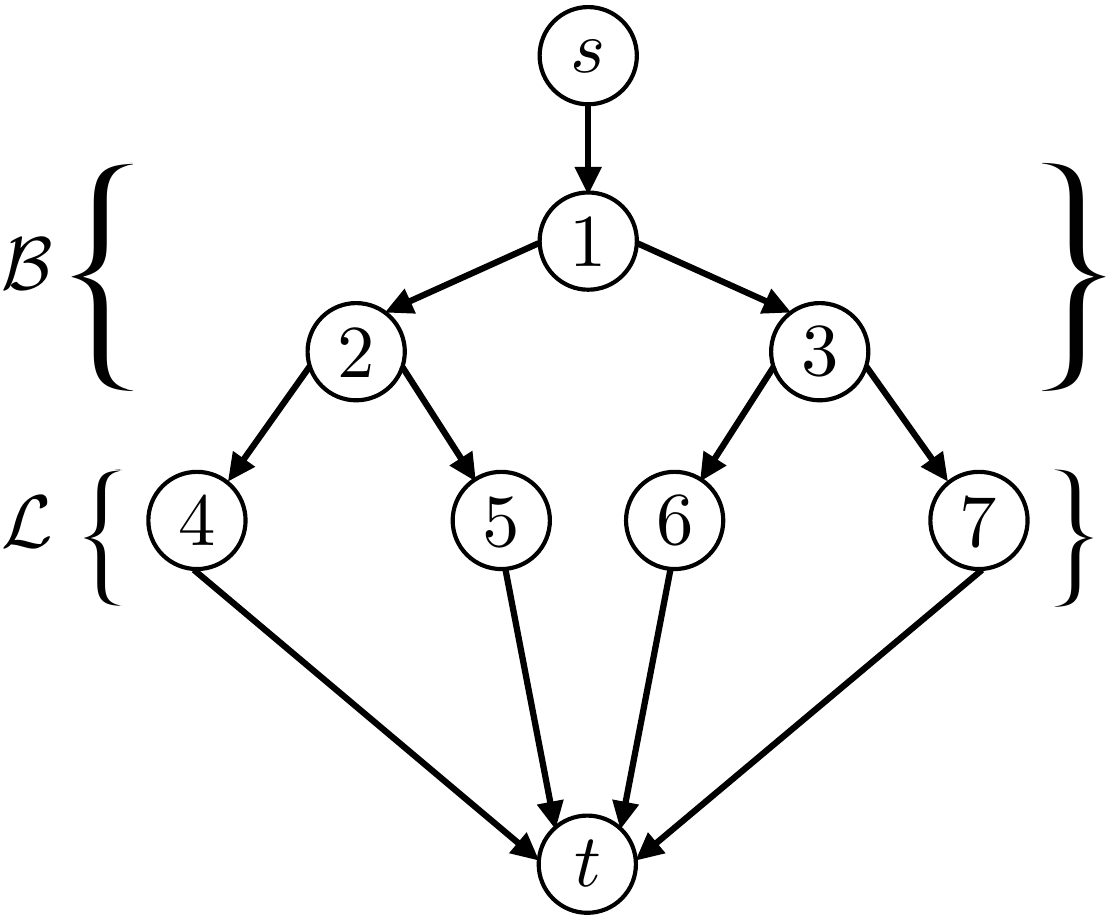}
}
\caption{A decision tree of depth 2 (left) and its associated flow graph (right). Here, $\sets B=\{1,2,3\}$ and $\sets L=\{4,5,6,7\}$, while $\sets V=\{s,1,2,\ldots,7,t\}$ and $\sets A = \{(s,1),(1,2),\ldots,(7,t)\}$.}
\label{fig:sample_tree}
\end{center}
%\vskip -0.2in
\end{figure}

An illustration of a balanced decision tree is provided in Figure~\ref{fig:sample_tree} (left). The key idea behind our model is to convert a balanced decision tree to a \emph{directed acyclic graph} by augmenting it with a single source node~$s$ that is connected to the root node (node 1) of the tree and a single sink node~$t$ connected to all leaf nodes of the tree. We refer to this graph as the \emph{flow graph} of the decision tree. An illustration of these concepts on a decision tree of depth $d=2$ is provided on Figure~\ref{fig:sample_tree}.

A formal definition for the flow graph associated with a balanced decision tree is as follows.
\begin{definition}[Flow Graph of a Balanced Decision Tree] \label{def:terminology}
Given a {balanced} decision tree of depth $d$, define the directed \emph{flow graph}  $\sets G=(\sets V,\sets A)$ associated with the tree as follows. Let $\sets V:=\{s,t\}\cup \sets B \cup \sets L$ be the vertices of the flow graph.  Given $n \in \sets B$, let $\ell(n):=2n$ be the \emph{left descendant of $n$,} $r(n):=2n+1$ be the \emph{right descendant of $n$}, 
and  
$$
\sets A :=\Big\{(n,\ell(n)): n\in \sets B\} \Big\}\cup \Big\{(n,r(n)): n\in \sets B\} \Big\} \cup \{(s,1)\}\cup \Big\{(n,t): n\in \sets  L\Big\}
$$
be the arcs of the graph. Also, given $n\in \sets B \cup \sets L$, let $a(n)$ be the \emph{ancestor} of $n$, defined through $a(1):=s$ and $a(n):=\lfloor n/2\rfloor$ if $n\neq 1$.
\end{definition}

%%%%%%%%%%%%%%%%%%%%%%%%%%%%%%%%%%%%%%%%%%%%%%%%%%
\subsection{Problem Formulation}
\label{sec:DTFlowFormulation}
%%%%%%%%%%%%%%%%%%%%%%%%%%%%%%%%%%%%%%%%%%%%%%%%%%

Let $\mathcal D:=\{ {\bm x}^i, y^i \}_{i\in \mathcal I}$ be a training dataset consisting of datapoints indexed in the set $\sets I \subseteq{\naturals}$. Each row $i\in \sets I$ consists of $F$ binary features indexed in the set $\sets F \subseteq{\naturals}$, which we collect in the vector ${\bm x}^i \in \{0,1\}^F$, and a label $y^i$ drawn from the finite set $\sets K$ of classes. In this section we formulate the problem of learning a multi-class classification tree of fixed finite depth 
$d \in \naturals$ that minimizes the misclassification rate (or equivalently, maximizes the number of correctly classified datapoints) based on MIO technology. 

In our formulation, the classification tree is described through the branching variables $\bm b$ and the prediction variables $\bm w$. In particular, we use the variables $b_{nf} \in \{0,1\}$, $f \in \sets F$, $n \in \sets B$, to indicate if the tree branches on feature $f$ at branching node $n$ (i.e., it equals 1 if and only if the binary test performed at~$n$ asks ``is $x^i_f=0$''?). Accordingly, we employ the variables $w^n_k \in \{0,1\}$, $n \in \sets L$, $k \in \sets K$, to indicate that at leaf node $n$ the tree predicts class~$k$. We use the auxiliary routing/flow variables $\bm z$ to decide on the flow of data through the flow graph associated with the decision tree. Specifically, for each node $n \in \sets B\cup \sets L$ and for each datapoint $i \in \sets I$, we introduce a decision variable $z^i_{a(n),n} \in \{0,1\}$ which equals 1 if and only if the $i$th datapoint is correctly classified and {its flow} traverses the arc $(a(n),n)$ on its way to the sink $t$. We let $z^i_{n,t}$
be defined accordingly for each {arc} between node $n \in  \sets L$ and sink $t$. Datapoint $i \in \sets I$ is correctly classified if and only if {its corresponding flow} passes through some leaf node $n \in \sets L$ such that $w^n_{y_i}=1$, i.e., where the class predicted coincides with the class of the datapoint. If {the flow of} a {data}point $i$ arrives at such a leaf node~$n$ and {the datapoint} is correctly classified, it{s corresponding flow} is directed to the sink, i.e., $z^i_{n,t}=1${; otherwise, the corresponding flow is not initiated from the source at all.}
With these variables, the flow-based formulation reads 
\begin{subequations}
\begin{align}
\text{maximize} \;\; & \displaystyle \sum_{i \in \mathcal I} \sum_{n \in \sets L } z^i_{n,t} \label{eq:flow_obj}\\
%%%%%%%%%%%%%%%%%%%%%%%%%%%%%%%%%binary constraints
%%%%%%%%%%%%%%%%%%%%%%%%%%%%%%%%%The flow conservation constraints
\text{subject to}\;\;
& \displaystyle \sum_{f \in \sets F}b_{nf} = 1   &\hspace{-5cm}  \forall n \in \sets B \label{eq:flow_internal_branch}\\
& \displaystyle z^i_{a(n),n} =  z^i_{n,\ell(n)} + z^i_{n,r(n)}  &\hspace{-5cm}  \forall n \in \sets B, i \in \mathcal I \label{eq:flow_conservation_internal}\\
&  \displaystyle z^i_{a(n),n} = z^i_{n,t} &\hspace{-5cm}   \forall n \in \sets L, i \in \mathcal I \label{eq:flow_conservation_terminal}\\
%%%%%%%%%%%%%%%%%%%%%%%%%%%%%%%%%capacity constraints
& \displaystyle z^i_{s,1} \leq 1 &\hspace{-5cm} \forall i \in \mathcal I\label{eq:flow_source}\\
&  \displaystyle z^i_{n,\ell(n)}\leq \sum_{f \in \sets F: x_{f}^i=0}b_{nf} &\hspace{-5cm} \forall n \in \sets B, i \in \mathcal I \label{eq:flow_branch_left}\\
&  \displaystyle z^i_{n,r(n)}\leq \sum_{f \in \sets F: x_{f}^i=1}b_{nf}  &\hspace{-5cm} \forall n \in \sets B, i \in \mathcal I \label{eq:flow_branch_right}\\
&  \displaystyle z^i_{n,t} \leq  w^n_{y^i} &\hspace{-5cm} \forall n \in  \sets L, i \in \mathcal I \label{eq:flow_sink}\\
%%%% DVs
&  \displaystyle \sum_{k \in \sets K}w^n_{k} = 1  &\hspace{-5cm}  \forall n \in  \sets L \label{eq:flow_leaf_prediction}\\
&  \displaystyle w^n_{k} \in \{0,1\}  &\hspace{-5cm}   \forall n \in  \sets L,k \in \sets K \\
&  \displaystyle b_{nf} \in \{0,1\}  &\hspace{-5cm}   \forall n \in \sets B,f \in \sets F \\
&  \displaystyle z^i_{a(n),n}\in \{0,1\}  &\hspace{-5cm}  \forall n \in \sets B \cup \sets L,i \in \sets I\\
&  \displaystyle z^i_{n,t}\in \{0,1\}  &\hspace{-5cm}  \forall n \in \sets L,i \in \sets I.
\end{align}
\label{eq:flow}
\end{subequations}

An interpretation of the constraints and objective is as follows. Constraint{s}~\eqref{eq:flow_internal_branch} ensure that at each branching node $n \in \sets B$ we branch on exactly one feature $f \in \sets F$.
Constraint{s}~\eqref{eq:flow_conservation_internal} {are} flow conservation constraints for each datapoint $i$ and node $n \in \sets B$: {they} ensure that if a datapoint arrives at a node, {then} it must also leave the node through one of its descendants. Similarly, constraint{s}~\eqref{eq:flow_conservation_terminal} enforce flow conservation for each node $n \in \sets L$.
The inequality constraint{s}~\eqref{eq:flow_source} imply that at most one unit of flow can enter the graph through the source {for each datapoint}. Constraint{s}~\eqref{eq:flow_branch_left} (resp.\ \eqref{eq:flow_branch_right}) ensure that if {the flow of }a datapoint is routed to the left (resp.\ right) at node~$n$, then one of the features such that $x^i_f=0$ (resp.\ $x^i_f=1$) must have been selected for branching at the node. Constraint{s}~\eqref{eq:flow_sink} guarantee that datapoints {whose flow is} routed to the sink node~$t$ are correctly classified. Constraint{s}~\eqref{eq:flow_leaf_prediction} make sure that each leaf node is assigned a predicted class $k \in \sets K$. The objective \eqref{eq:flow_obj} maximizes the total number of correctly classified datapoints.%  $\sum_{i \in \mathcal I} \sum_{n \in \sets L} z^i_{n,t}$.

Formulation~\eqref{eq:flow} has several distinguishing features relative to existing MIO formulations for training decision trees: {First,} it does not use big-$M$ constraints. {Second,} it includes \emph{flow variables} indicating whether each datapoint is directed to the left or right at each branching node, {which resembles the well-known multi-commodity flow concept, see~\citet{hu1963multi}.} 
% \emph{(c)}~It only tracks datapoints that are correctly classified. {\citet{gunluk2018optimal} also propose not to track incorrectly classified datapoints throughout their formulation, as their objective function only cares about the correctly classified points, as such the variables associated with the incorrectly classified points can be projected out from the formulation}. 
{Third, incorrectly classified datapoints are associated with a flow of zero. In contrast, previous formulations \revision{\citep{bertsimas2017optimal,verwer2019learning}} and more sophisticated formulations we propose later in Section~\ref{sec:Generalization} include binary decision variables that indicate the leaf each datapoint lands on. The advantage of modeling \revision{misclassified} points with a flow of zero is that the resulting formulation is smaller. A similar idea of ``projecting out'' variables associated with misclassified datapoints was proposed by~\citet{gunluk2018optimal}.} 

The number of variables and constraints in Problem~\eqref{eq:flow} is $\sets O\big(2^{d}(|\sets I|+|\sets F| + {|\sets K|})\big)$, where $d$ is the tree depth. Thus, its size is of the same order as the \revision{univariate splits formulation of~\citet{bertsimas2017optimal} (formulation (24) in their paper)} while being of higher order than that of~\citet{verwer2019learning}. Nonetheless, the LO relaxation of formulation \eqref{eq:flow} is tighter than {those} of~\citet{bertsimas2017optimal} and~\citet{verwer2019learning}, as demonstrated in the following theorem {which we prove in Electronic Companion~\ref{appendix_sec:oct_comparison}}, and therefore results in a more aggressive pruning of the {branch-and-bound tree}. 

%\notepv{here it is a little strange that we do not say anything about the size of the formulation of verwer (even if we have briefly discussed this in the intro)}\notesa{In case if we want to add it: Number of decision variables for~\citet{verwer2019learning} is $\sets O\big(2^{d}(|\sets F| + |\sets K| + log(T_{max}))\big)$ where $T_{max}$ is the maximum number of thresholds for any feature.}\noteag{This does not sound right: in this case, Verwer's would have a larger order than Bertsimas'.}.\notesa{You're right. The $|\sets I|$ term was there by mistake. Verwer's does not depend on the number of rows.}

\begin{theorem}\label{theo:OCT}
Problem~\eqref{eq:flow} has a stronger LO relaxation than the formulations of~\citet{bertsimas2017optimal} and~\citet{verwer2019learning}. 
\end{theorem}

\begin{remark}
Formulation~\eqref{eq:flow} assumes that all features $f \in \sets F$ are binary.
However, this formulation can also be applied to datasets involving categorical or {bounded} integer features by first preprocessing the data as follows. For each categorical feature, we encode it as a one-hot vector, i.e., for each level of the feature, we create a new binary column which has value 1 if and only if the original column has the corresponding level. We follow a similar approach for encoding integer features with a slight change. The new binary column has value 1 if and only if the main column has the corresponding value or any value smaller than it, see e.g.,\revision{~\citet{verwer2019learning} and~\citet{okada2019efficient}}. {This discretization of the features increases the size of the dataset linearly with the number of possible values of each categorical/integer feature.}
{We also note that formulation~\eqref{eq:flow} only considers univariate splits. It can however be generalized to allow for multivariate splits. Indeed, in the case of binary features, having multivariate (or oblique) splits is equivalent to ``combinatorial'' branching, which can be done by creating additional artificial features, see \cite{gunluk2018optimal} for \revision{additional} discussion.~\revision{We can also extend formulation~\eqref{eq:flow} for the case of non-binary trees, where each splitting node can have more than two children. This can be achieved by introducing additional nodes and edges to the flow graph while adjusting the formulation accordingly.}}

%. In this way if we split on level $l > 0$ of an integer feature~$f$, any datapoint with $x^i_f \leq l$ would take the left branch. This is a common approach in the literature
%\noteag{I would shorten the discussion for integer features as follows: ``A similar approach can be used for integer features, see~\citet{verwer2019learning,okada2019efficient} for a further discussion." }
\end{remark}           

%%%%%%%%%%%%%%%%%%%%%%%%%%%%%%%%%%%%%%%%%%%%%%%%%%
%%%%%%%%%%%%%%%%%%%%%%%%%%%%%%%%%%%%%%%%%%%%%%%%%%
\section{Benders' Decomposition via Facet-defining Cuts }
\label{sec:Benders}
%%%%%%%%%%%%%%%%%%%%%%%%%%%%%%%%%%%%%%%%%%%%%%%%%%
%%%%%%%%%%%%%%%%%%%%%%%%%%%%%%%%%%%%%%%%%%%%%%%%%%

In Section~\ref{sec:DT_Formulation}, we proposed a formulation for designing optimal classification trees that is provably stronger than existing {approaches} from the literature. Our {model} presents an attractive decomposable structure that we leverage in this section to speed-up computation.

%%%%%%%%%%%%%%%%%%%%%%%%%%%%%%
\subsection{Main Problem, Max-Flow Subproblems, and Benders' Decomposition}
%%%%%%%%%%%%%%%%%%%%%%%%%%%%%%

A classification tree is uniquely characterized by the branching decisions~{$\bm b$} and predictions~{$\bm w$} made at the branching nodes and leaves, respectively. Given a choice of {$\bm b$} and {$\bm w$}, each datapoint in problem~\eqref{eq:flow} is allotted one unit of flow {that can be guided} from the source node to the sink node through the flow graph associated with the decision tree. If the datapoint cannot be correctly classified, the flow that will reach the sink (and by extension enter the source) will be zero.  
In particular, once the branching variables~$\bm b$ and prediction variables~$\bm w$ have been fixed, optimization of the auxiliary flow variables $\bm z$ can be done separately for each datapoint. In particular, we can decompose problem~\eqref{eq:flow} into a \emph{main problem} involving the variables $({\bm b},{\bm w})$, and $|\sets I|$ \emph{subproblems} indexed by $i\in \sets I$ each involving the flow variables $\bm z^i$ associated with datapoint $i$. Additionally, each subproblem is a maximum flow problem for which specialized polynomial-time methods exist. Due to these characteristics, formulation~\eqref{eq:flow} can be naturally tackled using Benders' decomposition, see~\citet{benders1962partitioning}. In what follows, we describe the Benders' decomposition approach.

Problem~\eqref{eq:flow} can be written equivalently as:
\begin{subequations}
\begin{align}
\text{maximize}\;\;&   \displaystyle \sum_{i \in \sets I}g^i({\bm b},{\bm w})  \label{eq:master_obj}\\
\text{subject to} \; \; & \displaystyle \sum_{f \in \sets F}b_{nf} = 1   &\hspace{-5cm}  \forall n \in \sets B \label{eq:master_internal_branch}\\
&  \displaystyle \sum_{k \in \sets K}w^n_{k} = 1  &\hspace{-5cm}  \forall n \in  \sets L \label{eq:master_leaf_prediction}\\
%%%%%%%%%%%%%%%%%%%%%%%%%%%%%%%%%binary constraints
&  \displaystyle b_{nf} \in \{0,1\}  &\hspace{-5cm} \forall n \in \sets B,f \in \sets F \\
&  \displaystyle w^n_{k} \in \{0,1\}  &\hspace{-5cm}   \forall n \in  \sets L,k \in \sets K,
\end{align} 
\label{eq:master}%
\end{subequations}
where, for any fixed $i\in \sets I$, ${\bm w}$ and ${\bm b}$, the quantity $g^i({\bm b},{\bm w})$ is defined as the optimal objective value of the problem 
\begin{subequations}
\begin{align}
g^i({\bm b},{\bm w})=\text{maximize} \;\; & \displaystyle \sum_{n \in \sets L } z^i_{n,t} \label{eq:slave_obj}\\
%%%%%%%%%%%%%%%%%%%%%%%%%%%%%%%%%binary constraints
%%%%%%%%%%%%%%%%%%%%%%%%%%%%%%%%%The flow conservation constraints
\text{subject to}\;\;
& \displaystyle z^i_{a(n),n} =  z^i_{n,\ell(n)} + z^i_{n,r(n)}   &\hspace{-5cm}  \forall n \in \sets B \label{eq:slave_conservation_internal}\\
&  \displaystyle z^i_{a(n),n} = z^i_{n,t} &\hspace{-5cm}   \forall n \in \sets L \label{eq:slave_conservation_terminal}\\
%%%%%%%%%%%%%%%%%%%%%%%%%%%%%%%%%capacity constraints
& \displaystyle z^i_{s,1} \leq 1 \label{eq:slave_source}\\
&  \displaystyle z^i_{n,\ell(n)}\leq \sum_{f \in \sets F: x_{f}^i=0}b_{nf} &\hspace{-5cm} \forall n \in \sets B \label{eq:slave_branch_left}\\
&  \displaystyle z^i_{n,r(n)}\leq \sum_{f \in \sets F: x_{f}^i=1}b_{nf}  &\hspace{-5cm} \forall n \in \sets B \label{eq:slave_branch_right}\\
&  \displaystyle z^i_{n,t} \leq  w^n_{y^i} &\hspace{-5cm} \forall n \in  \sets L \label{eq:slave_sink}\\
%%%% DVs
&  \displaystyle z^i_{a(n),n}\in \{0,1\}  &\hspace{-5cm}  \forall n \in \sets B \cup \sets L \label{eq:slave_integral_1}\\
&  \displaystyle z^i_{n,t}\in \{0,1\}  &\hspace{-5cm}  \forall n \in \sets L \label{eq:slave_integral_2}.
\end{align}
\label{eq:slave}
\end{subequations}
Problem \eqref{eq:slave} is a maximum flow problem on the flow graph $\sets G$, see Definition~\ref{def:terminology}, whose arc capacities are determined by $({\bm b},{\bm w})$ and datapoint $i\in \sets I$, as formalized next.

\begin{definition}[Capacitated flow graph] Given the flow graph $\sets G=(\sets V,\sets A)$, vectors~$({\bm b},{\bm w})$, and datapoint $i \in \sets I$, define arc capacities $c^i({\bm b},{\bm w})$ as follows. Let $c^i_{s,1}({\bm b},{\bm w}):=1$, $c^i_{n,\ell(n)}({\bm b},{\bm w}):= \sum_{f \in \sets F: x_{f}^i=0}b_{nf}$ and $c^i_{n,r(n)}({\bm b},{\bm w}):=\sum_{f \in \sets F: x_{f}^i=1}b_{nf}$ for all $n \in \sets B$, and $c^i_{n,t}({\bm b},{\bm w}):= w^n_{y^i}$ for $n\in \sets L$. Define the \emph{capacitated flow graph} $\sets G^i({\bm b},{\bm w})$ as the flow graph $\sets G$ augmented with capacities $c^i({\bm b},{\bm w})$.
\end{definition}

Observe that the arc capacities in $\sets G^i({\bm b},{\bm w})$ are affine functions of $({\bm b},{\bm w})$. %Since~$b$ and~$w$ are feasible in problem~\eqref{eq:master}
An interpretation of the arc capacities and capacitated flow graph for{~$\bm b$ and~$\bm w$ } feasible in problem~\eqref{eq:master} is as follows. Constraints~\eqref{eq:master_internal_branch} imply that the weights~$c^i({\bm b},{\bm w})$, $i\in \sets I$, in the above definition are binary. Indeed, the capacity of the {arc} incoming into node 1 is 1. Moreover, for each node $n \in \sets B$ and datapoint $i\in \sets I$, exactly one of the outgoing {arcs} of node $n$ in graph $\sets G^i({\bm b},{\bm w})$ has capacity 1: the left {arc} if the datapoint passes the test ($c^i_{n,\ell(n)}({\bm b},{\bm w})=1$) or the right {arc} if it fails it ($c^i_{n,r(n)}({\bm b},{\bm w})=1$). Finally, for each node $n \in \sets L$ in graph~$\sets G^i({\bm b},{\bm w})$, its outgoing {arc} has capacity one if and only if the datapoint has the same label~$y^i$ as that predicted at node~$n$. Thus, for each datapoint, the set of {arc}s with capacity 1 in the graph~$\sets G^i({\bm b},{\bm w})$ forms a path from the source node to the leaf where this datapoint is assigned in the decision tree. This path is connected to the sink via an arc of capacity~1 if and only if the datapoint is correctly classified. {We note that as all the arc capacities in the flow graph are binary, the integrality constraints~\eqref{eq:slave_integral_1} and~\eqref{eq:slave_integral_2} can be relaxed.}

%\notepv{I think it would be nice to give an explanation in words of the above arc capacities: in particular, that for b and w binary, these upper bounds are either 0 or 1; also explain when they are zero and when they are one}
%\newsa{More specifically, at any branching node $n \in  \sets B$, $c^i_{n,\ell(n)}({\bm b},{\bm w})=1$ ($c^i_{n,r(n)}({\bm b},{\bm w})=1$) if the datapoint pass (fail) the test at that node. The arc that connect leaf node $n \in\sets L$ to the sink has a capacity  of one ($c^i_{n,t}({\bm b},{\bm w})=1$) if and only if datapoint $i$ has the same class label as the predicted class at node $n$. So $b$ and $w$ induce a specific graph for datapoint $i$ in Problem~\Eqref{eq:slave}. }
 
Problem~\eqref{eq:slave} is equivalent to a maximum flow problem on the capacitated flow graph~$\sets G^i({\bm b},{\bm w})$. From the well-known max-flow/min-cut duality, see e.g.,~\citet{vazirani2013approximation}, it follows that~$g^i({\bm b},{\bm w})$ is the capacity of a minimum~$(s,t)$ cut of graph~$\sets G^i({\bm b},{\bm w})$. In other words, it is the greatest value smaller than or equal to the value of all cuts in graph~$\sets G^i({\bm b},{\bm w})$. Given a set $\sets S\subseteq \sets V$, we let $\sets C(\sets S):=\{(n_1,n_2)\in \sets A: n_1\in \sets S, n_2\not\in \sets S\}$ denote the cut-set corresponding to the source set~$\sets S$. With this notation, problem~\eqref{eq:master} can be reformulated as
\begin{subequations}
	\begin{align}
	\displaystyle \mathop \text{maximize}_{\bm g,\bm b,\bm w} \;\;& \displaystyle \sum_{i \in \sets I}g^i \label{eq:master2_obj}\\
	\text{subject to}\;\;&  
	%%%%%%%%%%%%%%%%%%%%%%%%%%%%%%%%% constraints
	g^i \leq  \sum_{(n_1,n_2)\in \sets C(\sets S)}c_{n_1,n_2}^i({\bm b},{\bm w}) \quad\quad\qquad&\hspace{-5cm}  \forall i\in \sets I, \sets S\subseteq \sets V\setminus \{t\}: s\in \sets S \label{eq:master2_benders_cut}\\
	%%%%%%%%%%%%%%%%%%%%%%%%%%%%%%%%%binary constraints
	&  \displaystyle \sum_{f \in \sets F}b_{nf}=1   &\hspace{-5cm} \forall n \in \sets B  \label{eq:master2_internal_branch}\\
&  \displaystyle \sum_{k \in \sets K}w^n_{k} = 1  &\hspace{-5cm}  \forall n \in  \sets L \label{eq:master2_leaf_prediction}\\
	&  \displaystyle b_{nf} \in \{0,1\}  &\hspace{-5cm} \forall n \in \sets B,f \in \sets F \label{eq:master2_b_sign}\\
	&  \displaystyle w^n_{k} \in \{0,1\}  &\hspace{-5cm}   \forall n \in  \sets L,k \in \sets K \label{eq:master2_w_sign}\\
	&  \displaystyle g^i \leq 1   & \forall i \in \sets I \label{eq:master2_g_upperbound},
	\end{align}
	\label{eq:master2}%
\end{subequations}
where, at an optimal solution, $g^i$ represents the value of a minimum $(s,t)$ cut of graph~$\sets G^i({\bm b},{\bm w})$. Indeed, the objective function~\eqref{eq:master2_obj} maximizes the $g^i$, while constraints \eqref{eq:master2_benders_cut} ensure that the value of~$g^i$ is no greater than the value of a minimum cut.

Formulation~\eqref{eq:master2} contains an exponential number of inequalities~\eqref{eq:master2_benders_cut} and is implemented using row generation, whereby constraints~\eqref{eq:master2_benders_cut} are initially dropped and added on the fly during optimization as Benders' {cuts}. Note that we added the redundant constraint{s}~\eqref{eq:master2_g_upperbound} to ensure problem \eqref{eq:master2} is bounded even if all constraints~\eqref{eq:master2_benders_cut} are dropped. Row generation can be implemented in modern MIO solvers via callbacks, by adding lazy constraints at relevant nodes of the branch-and-bound tree. Identifying which constraint~\eqref{eq:master2_benders_cut} to add can in general be done by solving a minimum cut problem, and could in principle be solved via well-known algorithms, such as~\citet{goldberg1988new} and~\citet{hochbaum2008pseudoflow}.

The Benders' method we propose is reminiscent of the approach proposed by~\citet{lozano2018binary} to tackle two-stage problems in which the subproblem corresponds to a ``tractable'' 0-1 problem (i.e., a problem that admits an exact LO reformulation). In~\citet{lozano2018binary}, this subproblem is a decision diagram; in our case, it is a maximum flow problem. Thus, inequalities~\eqref{eq:master2_benders_cut} correspond to usual Benders' cuts, that replace the discrete subproblem with its (equivalent) continuous LO reformulation. As also pointed out by~\citet{lozano2018binary}, at each iteration of Benders' method, there are often multiple optimal dual solutions, each corresponding to a candidate inequality that can be added. Cuts obtained from most of these solutions might be weak in general, and thus they discuss how to strengthen them. In the next section, we propose a method that, among all potential optimal dual solutions, finds one that is guaranteed to result in a facet-defining cut for the convex hull of the feasible region given by inequalities~\eqref{eq:master2_benders_cut}-\eqref{eq:master2_g_upperbound}, that is, a cut that is already the best possible and admits no further strengthening. In addition, the method is faster than using a general purpose method to find minimum cuts or solve LO problems which we show in Electronic Companion~\ref{appendix_sec:benders_variants}. 
% \notesa{Do we need to update this part? To mention that we add some cuts at node 0? No, in section EC.4 we mention how we implement \benders and in section 5 we say we implement \benders as explained in EC4.}

%%%%%%%%%%%%%%%%%%%%%%%%%%%%%%
\subsection{Generating Facet-Defining Cuts via a Tailored Min-Cut Procedure}
%%%%%%%%%%%%%%%%%%%%%%%%%%%%%%

Row generation methods for integer optimization problems such as Benders' decomposition may require a long time to converge to an optimal solution if each added inequality is weak for the feasible region of interest. It is therefore of critical importance to add strong non-dominated cuts at each iteration of the algorithm, e.g., see~\citet{magnanti1981accelerating}. We now argue that not all inequalities~\eqref{eq:master2_benders_cut} are facet-defining for the convex hull of the feasible set of problem~\eqref{eq:master2}. To this end, let $\sets H_=$ represent the ({mixed-binary}) feasible region defined by constraints \eqref{eq:master2_benders_cut}-\eqref{eq:master2_g_upperbound}, and denote by $\text{conv}(\sets H_=)$ its convex hull.
Example~\ref{ex:counterexample} below shows that some of inequalities \eqref{eq:master2_benders_cut} are not facet-defining for $\text{conv}(\sets H_=)$ --even if they correspond to a minimum cut for a given value of $({\bm b},{\bm w})$-- and are in fact dominated. 

\iffalse
\newsa{
\begin{remark}\label{rem:ineq}
	Observe that in formulations~\eqref{eq:flow} and~\eqref{eq:master2}, apart from constraint~$\sum_{f\in \sets F}b_{nf}=1$, variable~$b$ appears only on the right-hand side of ``less than or equal to'' constraints. Therefore, it is possible to relax the equality constraint to~$\sum_{f\in \sets F}b_{nf}\leq 1$, as there always exists an optimal solution of the resulting problem where the relaxed constraint is active.
\end{remark}
}
\fi

%
\begin{example}\label{ex:counterexample}
		Consider an instance of problem~\eqref{eq:master2} with a depth $d=1$ decision tree (i.e., $\sets B=\{1\}$ and $\sets L=\{2,3\}$) and a dataset involving a single feature ($\sets F =\{1\}$). Consider datapoint~$i$ such that $x_1^i=0$ and $y^i=1$. Suppose that the solution to the main problem is such that we branch on (the unique) feature at node 1 and predict class $0$ at nodes 2 and 3. Then, datapoint $i$ is routed left at node 1 and is misclassified. A valid min-cut for the resulting graph includes all arcs incoming into the sink, i.e., $\sets S=\{s,1,2,3\}$ and $\sets C(\sets S)=\{(2,t), (3,t)\}$. The associated Benders' inequality~\eqref{eq:master2_benders_cut} reads 
		\begin{equation}\label{eq:weak}
		    g^i\leq w^2_{1}+w^3_{1}. 
		\end{equation}
		Intuitively, \eqref{eq:weak} states that datapoint $i$ can be correctly classified if its class label is assigned to at least one node, and is certainly valid for $\text{conv}(\sets H_=)$. However, since datapoint $i$ cannot be routed to node $3$, the stronger inequality
		\begin{equation}\label{eq:strong}
		    g^i\leq w^2_{1}
		\end{equation} is valid for $\text{conv}(\sets H_=)$ and dominates \eqref{eq:weak}.
\end{example}

Example~\ref{ex:counterexample} implies that an implementation of formulation \eqref{eq:master2} using general purpose min-cut algorithms to identify constraints to add may perform poorly.  This motivates us to develop a tailored algorithm that exploits the structure of each capacitated flow graph $\sets G^i({\bm b},{\bm w})$ to return inequalities that are never dominated, thus resulting in faster convergence of the Benders' decomposition approach.

\begin{figure}[ht]
\vskip 0.2in
\begin{center}
\centerline{\includegraphics[width=0.5\textwidth]{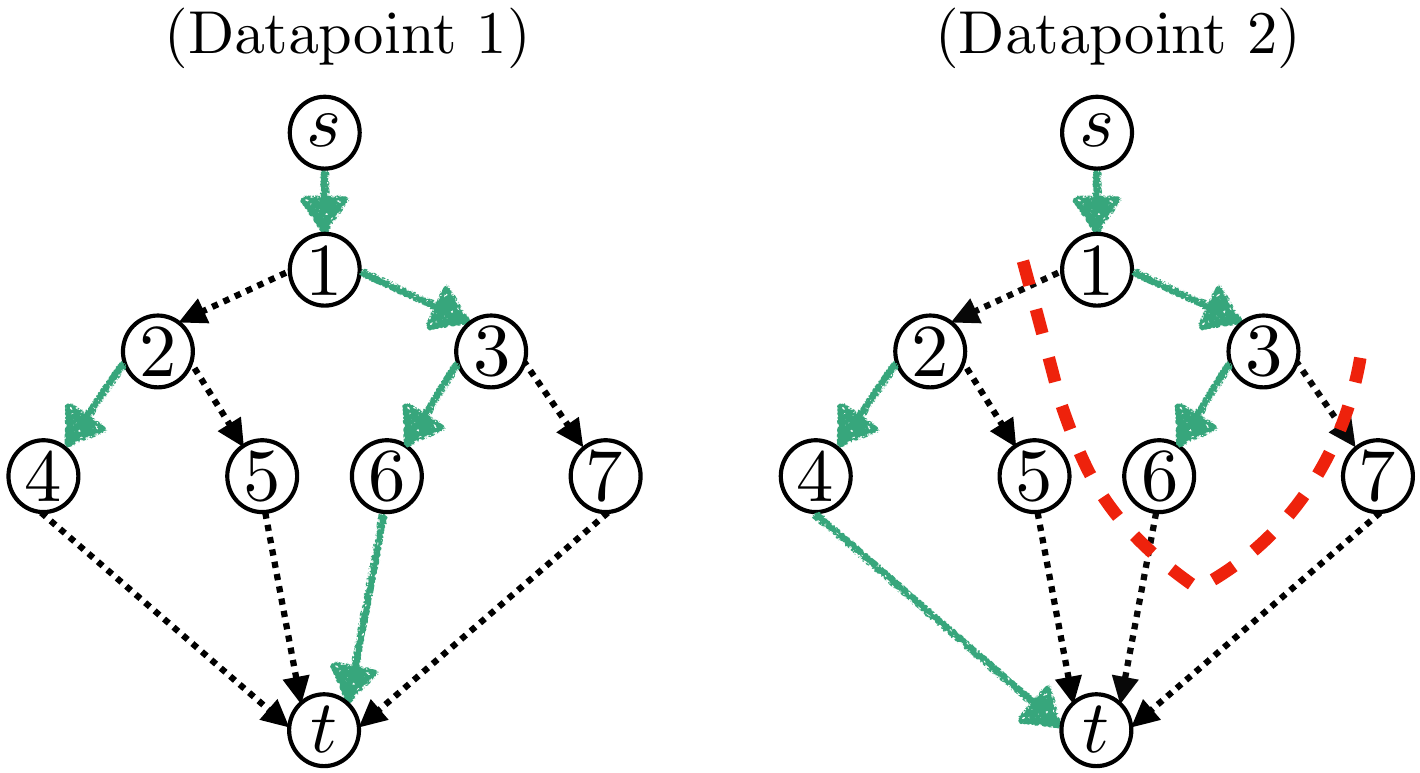}}
\caption{Illustration of Algorithm~\ref{alg:cut} on two datapoints that are correctly classified (datapoint 1, left) and incorrectly classified (datapoint 2, right). Unbroken (green) arcs~$(n,n')$ have capacity~$c_{n,n'}^i({\bm b},{\bm w})=1$ (and others capacity 0). In the case of datapoint 1 which is correctly classified since there exists a path from source to sink, Algorithm~\ref{alg:cut} terminates on line~\ref{line:-1} and returns~$-1$. In the case of datapoint 2 which is incorrectly classified, Algorithm~\ref{alg:cut} returns set~$\sets S = \{s,1,3,6\}$ on line~\ref{line:return}. The associated minimum cut consists of arcs $(1,2)$, $(6,t)$, and $(3,7)$ and is represented by the thick (red) dashed line.}
\label{fig:subproblem_tree}
\end{center}
\vskip -0.2in
\end{figure}

Algorithm~\ref{alg:cut} shows the proposed procedure, which can be called at \emph{integer nodes} of the branch-and-bound tree, using for example callback procedures available in most commercial and open-source solvers. 
{Algorithm~\ref{alg:cut} traverses the flow graph associated with a datapoint using depth-first search to investigate the existence of a path from the source to the sink. If there is such a path, it means that the datapoint is correctly classified and there are no violating inequalities. Otherwise, it outputs the set of all observed nodes in the traverse as the source set of the min-cut.} Figure~\ref{fig:subproblem_tree} illustrates Algorithm~\ref{alg:cut}. We now prove that Algorithm~\ref{alg:cut} is indeed a valid \emph{separation algorithm}.

\begin{algorithm}[h]
	\OneAndAHalfSpacedXI
		\caption{Separation procedure}
		\label{alg:cut}
		 \textbf{Input:} ${(\bm b,\bm w,\bm g)}\in \{0,1\}^{\sets B\times\sets F}\times \{0,1\}^{\sets L\times \sets K}\times \R^{\sets I} \text{ satisfying~\eqref{eq:master2_internal_branch}-\eqref{eq:master2_g_upperbound}; } \newline
		 \hspace*{\algorithmicindent} i~\in~\sets I:  \text{ datapoint used to generate the cut.}$ \newline
		 \textbf{Output:} $-1$ if all constraints \eqref{eq:master2_benders_cut} corresponding to $i$ are satisfied;  \newline 
		 \hspace*{\algorithmicindent} source set $\sets S$ of min-cut otherwise.
	    \begin{algorithmic}[1]
 		\State \textbf{if }$g^i=0$ \textbf{ then return } $-1$\label{line:simple}
 		\State \textbf{Initialize} $n\leftarrow 1$ \hfill \Comment{Current node $=$ root}
 		\State \textbf{Initialize} $\sets S\leftarrow \{s\}$ \hfill \Comment{$\sets S$ is in the source set of the cut}
 		\While{$n\in \sets B$}\label{line:ini}
 		\State $\sets S\leftarrow \sets S\cup\{n\}$
 		\If{$c_{n,\ell(n)}^i({\bm b},{\bm w})=1$}  \label{lin:sub-start}
 		\State $n\leftarrow \ell(n)$ \hfill \Comment{Datapoint $i$ is routed left}
 		\ElsIf{$c_{n,r(n)}^i({\bm b},{\bm w})=1$} \label{lin:sub-start2}
 		\State $n\leftarrow r(n)$ \hfill \Comment{Datapoint $i$ is routed right} \label{lin:sub-end}
 		\EndIf
 		\EndWhile \label{line:end}\Comment{At this point, $n\in \sets L$}
 		\State $\sets S\leftarrow \sets S\cup\{n\}$ \label{line:terminal_start}
 		\If{$g^i > c_{n,t}^i({\bm b},{\bm w})$} \Comment{Minimum cut $\sets S$ with capacity 0 found}\label{line:separation}
 		\State \textbf{return }$\sets S \label{line:return}$
 		\Else \Comment{Minimum cut $\sets S$ has capacity 1, constraints \eqref{eq:master2_benders_cut} are satisfied}\label{line:satisfaction}
 		\State \textbf{return} $-1 \label{line:-1}$
 		\EndIf\label{line:terminal_end}
		%\State 
	\end{algorithmic}
\end{algorithm}

\begin{proposition}\label{prop:algorithm}
 Given $i\in \sets I$ and $({\bm b,\bm w,\bm g})$ satisfying~\eqref{eq:master2_internal_branch}-\eqref{eq:master2_g_upperbound} (in particular, $\bm b$ and $\bm w$ are integral), Algorithm~\ref{alg:cut} either finds a violated inequality \eqref{eq:master2_benders_cut} or proves that all such inequalities are satisfied.
\end{proposition}
\proof{Proof.}
Note that the right{-}hand side of \eqref{eq:master2_benders_cut}, which corresponds to the capacity of a cut in the graph, is nonnegative. Therefore, if $g^i=0$ (line~\ref{line:simple}), all inequalities are automatically satisfied. Since $({\bm b},{\bm w})$ is integer, all arc capacities are either~0 or~1. {We assume that the arcs with zero capacity are removed from the flow graph.} Moreover, since $g^i\leq 1$, we find that either the value of a minimum cut is~$0$ and there exists a violated inequality, or the value of a minimum cut is at least~$1$ and there is no violated inequality. Finally, there exists a 0-capacity cut if and only if $s$ and~$t$ belong to different connected components in the graph $\sets G^i({\bm b},{\bm w})$.

The connected component $s$ belongs to, can be found using depth-first search.
For any fixed $n \in \sets B$, {constraints~\eqref{eq:master2_internal_branch}} and the definition of $c^i({\bm b},{\bm w})$ imply that either arc $(n,\ell(n))$ or arc $(n,r(n))$ has capacity 1 (but not both). If arc $(n,\ell(n))$ has capacity 1 (line~\ref{lin:sub-start}), then $\ell(n)$ can be added to the component connected to~$s$ (set $\sets S$); the case where arc $(n,r(n))$ has capacity 1 (line~\ref{lin:sub-start2}) is handled analogously. This process continues until a leaf node is reached (line~\ref{line:terminal_start}). If the capacity of the arc to the sink is 1 (line~\ref{line:satisfaction}), then an {$s-t$ path} is found and no cut with capacity $0$ exists. Otherwise (line~\ref{line:separation}), $\sets S$ is the connected component of $s$ and $t\not \in \sets S$, thus $\sets S$ is the source of a minimum cut with capacity~$0$.\halmos
\endproof

Observe that Algorithm~\ref{alg:cut}, which exploits the specific structure of the network for binary $({\bm b},{\bm w})$ feasible in \eqref{eq:master}, is much faster than general purpose minimum-cut methods. Indeed, since at each iteration in the main loop (lines~\ref{line:ini}-\ref{line:end}), the value of $n$ is updated to a descendant of $n$, the algorithm terminates in a{t} most $\sets O(d)$ iterations, where~$d$ is the depth of the tree. As $|\sets B\cup \sets L|$ is $\sets O(2^d)$, the complexity is logarithmic in the size of the tree. 
In addition to providing a very fast method for generating Benders' inequalities at integer nodes of a branch-and-bound tree, Algorithm~\ref{alg:cut} is guaranteed to generate strong non-dominated inequalities. 

Define 
$$
    \sets H_\leq  :=  \left\{ 
    \begin{array}{lcl}
    \bm b\in \{0,1\}^{\sets B\times \sets F},  \bm w\in \{0,1\}^{\sets L\times \sets K}, \bm g\in \R^{\sets I} & : & \displaystyle \sum_{f\in \sets F}b_{nf}\leq 1 \;\; \forall n\in \sets B \\
    && \displaystyle \sum_{k\in \sets K}w^n_k\leq 1 \;\; \forall n\in \sets L \\
    && \displaystyle g^i \leq  \sum_{(n_1,n_2)\in \sets C(\sets S)}c_{n_1,n_2}^i({\bm b},{\bm w}) \;  \forall i\in \sets I, \sets S\subseteq \sets V\setminus \{t\}: s\in \sets S 
    \end{array} 
    \right\}.
$$
Our main motivation for introducing $\sets H_\leq$ is that $\conv(\sets H_\leq)$ is full dimensional, whereas $\conv(\sets H_=)$ is not. Moreover, in formulation \eqref{eq:master2}, apart from constraints \eqref{eq:master2_internal_branch}-\eqref{eq:master2_leaf_prediction}, variables {$\bm b$ and $\bm w$ }only appear in the right{-}hand side of inequalities \eqref{eq:master2_benders_cut} with non-negative coefficients. Therefore, replacing constraints \eqref{eq:master2_internal_branch}-\eqref{eq:master2_w_sign} with $({\bm b},{\bm w})\in \sets H_\leq$ still results in valid formulation for problem \eqref{eq:master2}, since there exists an optimal solution where the inequalities are tight. Theorem~\ref{theo:facet} below formally states that Algorithm~\ref{alg:cut} generates non-dominated inequalities.

\begin{theorem}\label{theo:facet}
All violated inequalities generated by Algorithm~\ref{alg:cut} are facet-defining for $\rm{conv}(\sets H_\leq)$.
\end{theorem}
%\delsa{We defer the proof of Theorem~\ref{theo:facet} to the supplemental material~\ref{appendix_sec:polyhedral}.}

\begin{example}[Example~\ref{ex:counterexample} Continued]
	In the instance considered in Example~\ref{ex:counterexample}, if $b_{1f}=1$ and $w^2_{1}=0$, then the cut generated by Algorithm~\ref{alg:cut} has source set~$\sets S=\{s,1,2\}$ and results in inequality %\notepv{i would prefer it slightly if we could explain how the inequality is obtained (just write down the expression on the right hand side for this source set)}
	$$
	\displaystyle g^i \; \leq \;  \sum_{(n_1,n_2)\in \sets C(\sets S)}c_{n_1,n_2}^i({\bm b},{\bm w}) \; = \; c_{2,4}^i({\bm b},{\bm w}) + c_{1,3}^i({\bm b},{\bm w}) \; = \; w^2_1,
	$$ 
	which is precisely the stronger inequality~\eqref{eq:strong}.
\end{example}

\revision{\begin{remark}
\label{remark:benders_implementation}
Algorithm~\ref{alg:cut} can only be invoked at integer nodes of the branch-and-bound tree. However, there is a slight advantage in including as many cut-set inequalities as possible at the root node of the branch-and-bound tree by utilizing Gurobi to solve the corresponding linear optimization problem for each subproblem. This approach allows us to enhance the LO relaxation at integer nodes. A detailed investigation on this matter is reported in Electronic Companion~\ref{appendix_sec:benders_variants}.
\end{remark}}

%%%%%%%%%%%%%%%%%%%%%%%%%%%%%%%%%%%%%%%%%%%%%%%%%%
%%%%%%%%%%%%%%%%%%%%%%%%%%%%%%%%%%%%%%%%%%%%%%%%%%
\section{Generalizations}
\label{sec:Generalization}
%%%%%%%%%%%%%%%%%%%%%%%%%%%%%%%%%%%%%%%%%%%%%%%%%%
%%%%%%%%%%%%%%%%%%%%%%%%%%%%%%%%%%%%%%%%%%%%%%%%%%

In Section~\ref{sec:DT_Formulation}, we proposed a flow-based MIO formulation for designing optimal \emph{balanced} decision trees, see problem~\eqref{eq:flow}. In this section, we generalize this core formulation to design regularized (i.e., not necessarily balanced) classification trees wherein the distance from root to leaf may vary across leaves. We also discuss a variant that tracks all datapoints, even those that are not correctly classified, making it suitable to learn from imbalanced datasets, and to design fair decision trees.

%%%%%%%%%%%%%%%%%%%%%%%%%%%%%%%%%%%%%%%%%%%%%%%%%%
\subsection{Imbalanced Decision Trees}
\label{sec:regularization}
%%%%%%%%%%%%%%%%%%%%%%%%%%%%%%%%%%%%%%%%%%%%%%%%%%

Formulation~\eqref{eq:flow} outputs a balanced decision tree as defined in Definition~\ref{def:balanced_trees}. Such trees may result in over-fitting of the data and poor out-of-sample performance, in particular if $d$ is large. To this end, we propose a variant of formulation~\eqref{eq:flow} which allows for the design of trees that are not necessarily balanced and that have the potential of performing better out-of-sample. To this end, we introduce the following terminology pertaining to \emph{imbalanced} trees.
\begin{definition}[Imbalanced Decision Trees] \label{def:imbalanced_trees}
An imbalanced decision tree of (maximum) depth $d \in \naturals$ is a full binary tree, i.e., a tree in which every node has either~0 or~2 children and where the largest depth of a leaf is~$d$. We let~$\sets B:=\{1,\ldots,2^d-1\}$ denote the set of all \emph{candidate} branching nodes and~$\sets T:=\{2^d,\ldots, 2^{d+1}-1\}$ represent the set of \emph{terminal nodes}. We will refer to a node $n\in \sets B \cup \sets T$ as a \emph{leaf} if no branching occurs at the node.
\end{definition}

Note that in a \emph{balanced} decision tree, see Definition~\ref{def:balanced_trees}, branching occurs at all nodes in $\sets B$, and leaves of the decision tree correspond precisely to all nodes of maximum depth, i.e., $\sets L=\sets T$. In contrast, in \emph{imbalanced} decision trees  nodes in $\sets B$ can be leaf nodes and terminal nodes in $\sets T$ need not be part of the decision tree.

\begin{figure}[t!]
%\vskip 0.2in
\begin{center}
\centerline{
\includegraphics[width=0.35\textwidth]{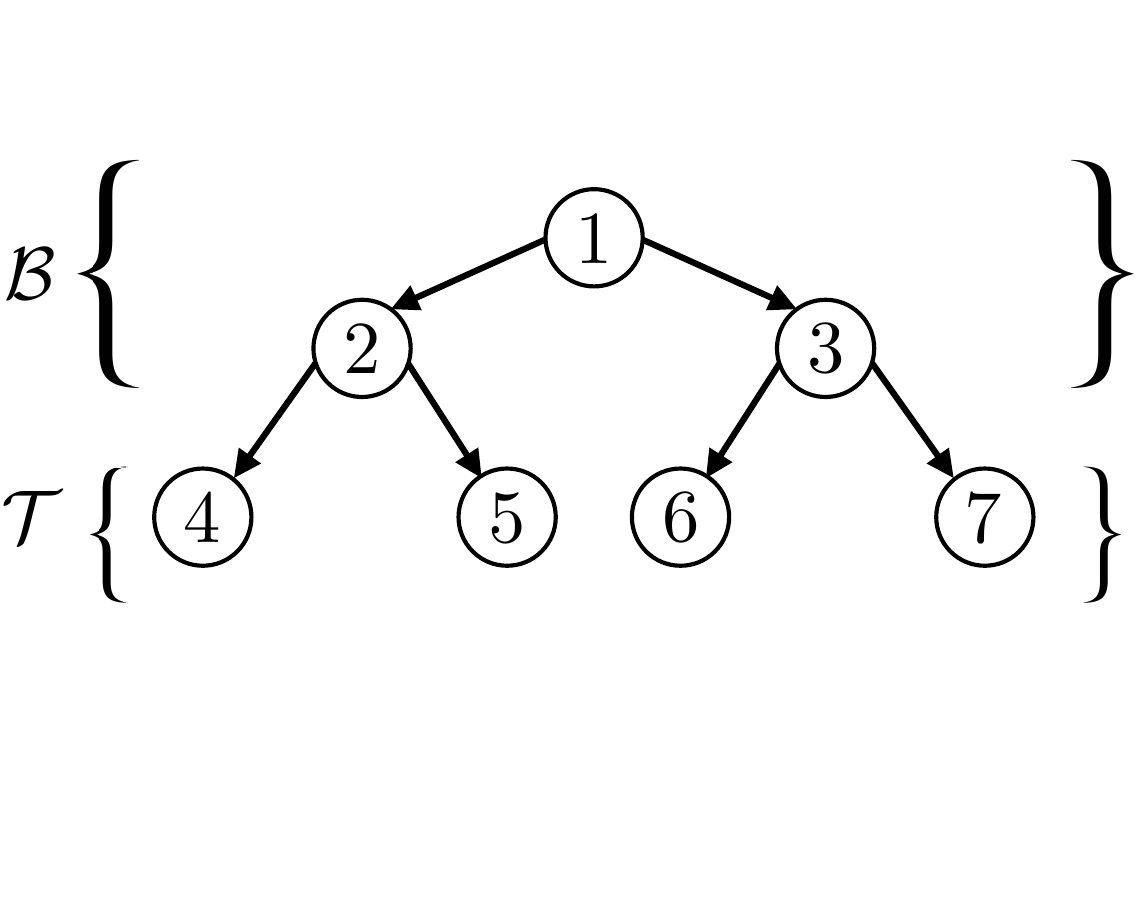}
\hspace{2cm}
\includegraphics[width=0.35\textwidth]{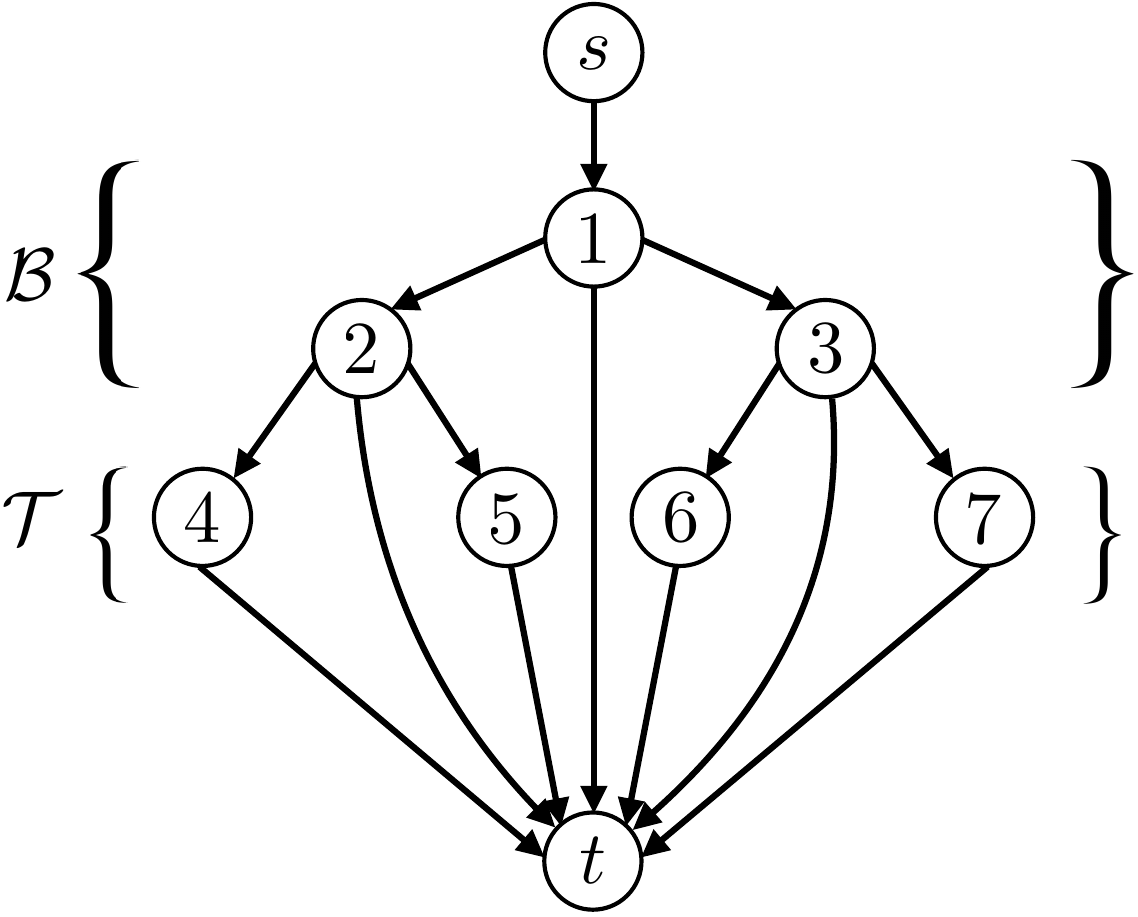}
}
\caption{A decision tree of depth 2 (left) and its associated flow graph (right) that can be used to train imbalanced decision trees of maximum depth $2$. Here, $\sets B=\{1,2,3\}$ and $\sets T=\{4,5,6,7\}$, while $\sets V=\{s,1,2,\ldots,7,t\}$ and $\sets A = \{(s,1),(1,2),(1,t),\ldots,(7,t)\}$. The additional {arc}s that connect the branching nodes to the sink allow branching nodes~$n \in \sets B$ to be converted to leaves where a prediction is made. Correctly classified datapoints that reach a leaf are directed to the sink. Incorrectly classified datapoints are not allowed to flow in the graph.}
\label{fig:sample_tree_penalty}
\end{center}
%\vskip -0.2in
\end{figure}

Akin to Definition~\ref{def:terminology}, we associate with an imbalanced decision tree a directed acyclic graph by augmenting the tree with a single source node~$s$ that is connected to the root node of the tree and a single sink node~$t$ that is connected to \emph{all nodes}~$n \in \sets B \cup \sets T$ of the tree, allowing correctly classified datapoints to flow to the sink from any node where a prediction is made (leaf of the learned tree). An illustration of these concepts on an imbalanced decision tree of depth $d = 2$ is provided on Figure~\ref{fig:sample_tree_penalty}. A formal definition for the flow graph associated with a imbalanced decision tree of maximum depth~$d$ is as follows. 

\begin{definition}[Flow Graph of an Imbalanced Decision Tree] \label{def:terminology_imbalanced}
Given an imbalanced decision tree of depth~$d$, we define its associated directed \emph{flow graph}~$\sets G=(\sets V,\sets A)$ as follows. Let $\sets V:=\{s,t\}\cup \sets B \cup \sets T$ be the vertices of the flow graph.  Given $n \in \sets B$, let $\ell(n):=2n$ be the \emph{left descendant of $n$,} $r(n):=2n+1$ be the \emph{right descendant of $n$}, 
and  
$$
\sets A :=\Big\{(n,\ell(n)): n\in \sets B\} \Big\}\cup \Big\{(n,r(n)): n\in \sets B\} \Big\} \cup \{(s,1)\}\cup \Big\{(n,t): n\in \sets  B \cup \sets T\Big\}
$$
be the arcs of the graph. Also, given $n\in \sets B \cup \sets T$, let $a(n)$ be the \revision{\textit{parent}} of $n$, defined through $a(1):=s$ and $a(n):=\lfloor n/2\rfloor$ if $n\neq 1$.
\end{definition}

%The updated form of the directed acyclic graph~\eqref{fig:sample_tree} is shown in~\eqref{fig:sample_tree_penalty}. 
%\newsa{The new edges connecting the branching node $n \in \sets B$ to the sink, as introduced in Definition~\ref{def:terminology_imbalanced}, are meant to allow the branching nodes not to split. In such a scenario, the branching node become a leaf of the tree and make a prediction and the datapoints arrived in that node can only get routed to the sink if they have the same label as the predicted label of the leaf node.}
 
We are now ready to formulate the variant of problem~\eqref{eq:flow} that allows for the design of imbalanced classification trees. In addition to the decision variables from formulation~\eqref{eq:flow} we introduce, for every node~$n \in \sets B \cup \sets T$, the binary decision variable~$p_n$  which has a value of one {if and only if} node~$n$ is a leaf node of the tree, i.e., if we make a prediction at node~$n$. The auxiliary routing/flow variables $\bm z$ now account for all {arc}s in the flow graph introduced in Definition~\ref{def:terminology_imbalanced}. The problem of learning optimal imbalanced classification trees is then expressible as% \noteag{I added the regularization to the objective here: otherwise, the problem makes no sense as $p_n=1$ {if and only if} $n\in \sets T$, which may confuse readers.}
%\newpage
\begin{subequations}
\begin{align}
\text{maximize} \;\; & \displaystyle (1-\lambda) \sum_{i \in \mathcal I} \sum_{n \in \sets B \cup \sets T } z^i_{n,t}  - \lambda \sum_{n \in \sets B}\sum_{f\in \sets F}b_{nf} \label{eq:flow_reg_obj}\\
%%%%%%%%%%%%%%%%%%%%%%%%%%%%%%%%%binary constraints
%%%%%%%%%%%%%%%%%%%%%%%%%%%%%%%%%The flow conservation constraints
\text{subject to}\;\;
& \displaystyle \sum_{f \in \sets F}b_{nf} + p_n + \sum_{m \in \sets P(n)}p_m = 1   &\hspace{-5cm}  \forall n \in \sets B \label{eq:flow_reg_branch_or_predict}\\
&  p_n+\sum_{m \in \sets P(n)}p_{m} =1   &  \forall n \in \sets T \label{eq:flow_reg_terminal_leaf}\\
& \displaystyle z^i_{a(n),n} =  z^i_{n,\ell(n)} + z^i_{n,r(n)} + z^i_{n,t}  &\hspace{-5cm}  \forall n \in \sets B, i \in \mathcal I \label{eq:flow_reg_conservation_internal}\\
&  \displaystyle z^i_{a(n),n} = z^i_{n,t} &\hspace{-5cm}   \forall n \in \sets T, {i \in \mathcal I} \label{eq:flow_reg_conservation_terminal}\\
%%%%%%%%%%%%%%%%%%%%%%%%%%%%%%%%%capacity constraints
& \displaystyle z^i_{s,1} \leq 1 &\hspace{-5cm} \forall i \in \mathcal I\label{eq:flow_reg_source}\\
&  \displaystyle z^i_{n,\ell(n)}\leq \sum_{f \in \sets F: x_{f}^i=0}b_{nf} &\hspace{-5cm} \forall n \in \sets B, i \in \mathcal I \label{eq:flow_reg_branch_left}\\
&  \displaystyle z^i_{n,r(n)}\leq \sum_{f \in \sets F: x_{f}^i=1}b_{nf}  &\hspace{-5cm} \forall n \in \sets B, i \in \mathcal I \label{eq:flow_reg_branch_right}\\
&  \displaystyle z^i_{n,t} \leq  w^n_{y^i} &\hspace{-5cm} \forall  n \in \sets B \cup \sets T, {i \in \mathcal I} \label{eq:flow_reg_sink}\\
%%%% DVs
&  \displaystyle \sum_{k \in \sets K}w^n_{k} = p_n  &\hspace{-5cm}  \forall n \in \sets B \cup \sets T \label{eq:flow_reg_leaf_prediction}\\
&  \displaystyle w^n_{k} \in \{0,1\}  &\hspace{-5cm}   \forall n \in \sets B \cup \sets T,k \in \sets K \\
&  \displaystyle b_{nf} \in \{0,1\}  &\hspace{-5cm}   \forall n \in \sets B,f \in \sets F \\
&  \displaystyle p_{n} \in \{0,1\}  &\hspace{-5cm}   \forall n \in \sets B \cup \sets T \\
&  \displaystyle z^i_{a(n),n}, z^i_{n,t}\in \{0,1\}  &\hspace{-5cm}  \forall n \in \sets B \cup \sets T,i \in \sets I,
\end{align}
\label{eq:flow_reg}%
\end{subequations}
where \revision{$\sets P(n)$} is the set of all ancestors  of node \revision{$n \in \mathcal B \cup \mathcal T$}{, i.e., the set of all nodes lying on the unique path from the root node to node $n$,} and $\lambda\in [0,1]$ is a regularization parameter. An explanation of the new/modified problem constraints is as follows. Constraint{s}~\eqref{eq:flow_reg_branch_or_predict} {imply} that at any node $n \in \sets B$  we either branch on a feature $f$ (if~$\sum_{f \in \sets F}b_{nf}=1$), predict a label (if~$p_n=1$), or get pruned if a prediction is made at one of the node ancestors (i.e., if~$\sum_{m \in \sets P(n)}p_m=1$). Similarly constraint{s}~\eqref{eq:flow_reg_terminal_leaf} {ensure} that any node $n \in \sets T$ is either a leaf node of the tree or is pruned. Constraints~\eqref{eq:flow_reg_conservation_internal}-\eqref{eq:flow_reg_leaf_prediction} exactly mirror constraints~\eqref{eq:flow_conservation_internal}-\eqref{eq:flow_leaf_prediction} in problem~\eqref{eq:flow}. They are slight modifications of the original constraints accounting for the new {arc}s added to the flow graph and for the possibility of making predictions at branching nodes $n \in \sets B$. In particular, constraint{s}~\eqref{eq:flow_reg_leaf_prediction} {imply} that if a node~$n$ gets pruned we do not predict any class at the node, i.e., $w^n_k=0$ for all~$k \in \sets K$. A penalty term is added to \eqref{eq:flow_reg_obj}, to encourage sparser trees with fewer branching decisions. 
{Note that while it is feasible to design a decision tree with the same branching decisions in multiple nodes on a single path from root to sink, such solutions are never optimal for \eqref{eq:flow_reg} if $\lambda>0$, as a simpler tree would result in the same misclassification.}
% {
% Problem~\eqref{eq:flow} may output an optimal decision tree with the same branching decisions in multiple nodes on a single path from root to sink. However, the regularization term in~\eqref{eq:flow_reg_obj}, prevents such optimal solutions.} 

Problem~\eqref{eq:flow_reg} allows for the design of regularized decision trees by augmenting this nominal formulation with additional regularization constraints. These either limit or penalize the number of nodes that can be used for branching or place a lower bound on the number of datapoints that land on each leaf. We detail these variants in the following. All of these variants either explicitly or implicitly restrict the tree size, thereby mitigating the risk of overfitting and resulting in more interpretable trees.

%%%%%%%%%%%%%%%%%%%%%%%%%%%%%%%%%%%%%%%%%%%%%%%%%%

\paragraph{Sparsity.} A commonly used approach to restrict tree size is to add sparsity constraints which restrict the number of branching nodes, see e.g., \citet{breiman1984classification,quinlan2014c4,bertsimas2017optimal},~\citet{aghaei2019learning}{, and \citet{blanquero2020sparsity}}. Such constraints are usually met in traditional methods by performing a post-processing \emph{pruning} step on the learned (unrestricted) tree. We enforce this restriction by adding the constraint
\begin{equation}\label{eq:branching_limit}
    \sum_{n\in \sets B}\sum_{f\in \sets F}b_{nf} \leq C
\end{equation}
to problem~\eqref{eq:flow_reg}, and possibly setting $\lambda=0$. This constraint ensures that the learned tree has at most~$C$ branching nodes (thus at most $C+1$ leaf nodes), where~$C$ is a hyper-parameter that can be tuned using cross-validation, see e.g., \citet{bishop2006pattern}. We note that due to the non-convexity introduced by the integer variables in problem~\eqref{eq:flow_reg}, the constrained sparsity formulation provides more options than the penalized version, see e.g.,~\citet{lombardi2020analysis}. In other words, for any choice of $\lambda \in [0,1]$ in the penalized version, there exists a choice of~$C \in \{0,\ldots,2^d\}$ in the constrained version that yields the same solution, but the converse is not necessarily true. The penalized version is more common in the machine learning literature and we will thus use it for benchmarking purposes in our numerical results, see Section~\ref{sec:Experiments}.

%%%%%%%%%%%%%%%%%%%%%%%%%%%%%%%%%%%%%%%%%%%%%%%%%%

%%%%%%%%%%%%%%%%%%%%%%%%%%%%%%%%%%%%%%%%%%%%%%%%%%

\paragraph{Maximum Number of Features to Use.} One can also constrain the total number~$C$ of features branched on in the decision tree as in~\citet{aghaei2019learning} by adding the constraints
\begin{equation}\label{eq:flow_regularization_max_num_features}
\begin{array}{ll}
\displaystyle \sum_{f \in \sets F} b_f \leq C \text{ and } b_f \geq b_{nf} \;\; \forall n \in \sets B, \; f\in \sets F
\end{array}
\end{equation}
to problem~\eqref{eq:flow_reg}, where $b_f \in \{0,1\}$, $f \in \sets F$, are decision variables that indicate if feature~$f$ is used in the tree. Parameter~$C$ can be a user-specified input or tuned via cross-validation. Note that neither integrality nor bound constraints on variable $b_f$ need to be enforced in the formulation.

% \delsa{We point out that the Benders' decomposition approach described in Section~\ref{sec:Benders} can be straightforwardly extended to any of the aforementioned variants. The modifications needed to account for the additional variables $z_{n,t}$, $n\in \sets B$, are detailed in Electronic Companion~\ref{appendix_sec:Benders}.}

\paragraph{Minimum Number of Datapoints in each Leaf Node.} Another popular regularization approach, see e.g., \citet{bertsimas2017optimal}, consists in placing a lower bound on the number of datapoints that land in each leaf. To ensure that each node contains at least $N_{\text{min}}$ datapoints, we impose
\begin{equation}\label{eq:flow_regularization_min_datapoints_per_leaf}
     \displaystyle \sum_{i\in \sets I}z^i_{a(n),n} \geq N_{\text{min}} p_n \quad \forall n \in \sets B \cup \sets T
\end{equation}
in problem~\eqref{eq:flow_reg}.

%%%%%%%%%%%%%%%%%%%%%%%%%%%%%%%%%%%%%%%%%%%%%%%%%%
\subsection{Imbalanced Datasets}
\label{sec:imbalanced_datasets}
%%%%%%%%%%%%%%%%%%%%%%%%%%%%%%%%%%%%%%%%%%%%%%%%%%

A dataset is called \emph{imbalanced} when the class distribution is not uniform, i.e., when the number of datapoints in each class varies significantly from class to class. In the case when a dataset consists of two classes, i.e., $\sets K:=\{k,k'\}$, we say that it is imbalanced if
$$
| \{ i \in \sets I \; : \; y^i = k  \} | \; \gg \; | \{ i \in \sets I \; : \; y^i = k'  \} |,
$$
in which case $k$ and $k'$ are referred to as the majority and minority classes, respectively. In imbalanced datasets, predicting the majority class for all datapoints results in high accuracy, and thus decision trees that maximize prediction accuracy without accounting for the imbalanced nature of the data perform poorly on the minority class. Imbalanced datasets occur in many important domains, e.g., to predict landslides, to detect fraud, or to predict if a patient has cancer, e.g., see,\revision{~\citet{kirschbaum2009evaluation,wei2013effective}, and~\citet{khalilia2011predicting}}. Naturally, being able to predict the minority class(es) accurately is crucial in such settings.

In the following we propose adjustments to the core formulation~\eqref{eq:flow_reg} to ensure meaningful decision trees are learned even in the case of imbalanced datasets. 
These adjustments require calculating metrics such as true positives, true negatives, false positives, and false negatives or some function of them such as recall (fraction of all datapoints from the positive class that are correctly identified) and precision (the portion of correctly classified datapoints from the positive class out of all datapoints with positive predicted class). Thus, learning meaningful decision trees requires us to track \emph{all datapoints}, not only the correctly classified ones as done in formulations~\eqref{eq:flow} and~\eqref{eq:flow_reg}.

To this end, we propose to replace the single sink in the flow graph associated with a imbalanced decision tree, see Definition~\ref{def:terminology_imbalanced}, 
with~$|\sets K|$ sink nodes denoted by~$t_k$, one for each class $k \in \sets K$. Each sink node $t_k$, $k\in \sets K$, is connected to all nodes~$n \in \sets B \cup \sets T$ and collects all datapoints (both correctly and incorrectly classified) with predicted class~$k$.

A formal definition for the flow graph associated with a decision tree that can be used to train imbalanced decision trees and that can track all datapoints is as follows.
\begin{definition}[Complete Flow Graph of an Imbalanced Decision Tree] \label{def:terminology_imbalanced_all} Given a decision tree of depth~$d$, we define the directed \emph{complete flow graph}  $\sets G=(\sets V,\sets A)$ associated with the tree that can be used to learn imbalanced decision trees of maximum depth~$d$ and that tracks all datapoints through the graph as follows. Let $\sets V:=\{s\}\cup \{t_k: k \in \sets K\}\cup \sets B \cup \sets T$ be the vertices of the flow graph.  Given $n \in \sets B$, let $\ell(n)$, $r(n)$, and $a(n)$ be as in Definition~\ref{def:terminology_imbalanced} and let
$$
\begin{array}{l}
\sets A :=\Big\{(n,\ell(n)): n\in \sets B\} \Big\}\cup \Big\{(n,r(n)): n\in \sets B\} \Big\} \cup \{(s,1)\} \\ 
\qquad \qquad \qquad \cup \Big\{(n,t_k): n\in \sets  B \cup \sets T, k \in \sets K\Big\}
\end{array}
$$
be the arcs of the graph.
\end{definition}

We are now ready to formulate the variant of problem~\eqref{eq:flow_reg} that tracks all datapoints through the flow graph with sink nodes $t_k$, $k\in \sets K$. In a way that parallels formulation~\eqref{eq:flow_reg}, we introduce auxiliary variables~$z^i_{n,t_k} \in \sets \{0,1\}$ for each node~$n\in \sets B \cup \sets T$ and each class $k \in \sets K$ to track the flow of datapoints (both correctly classified and missclassified) to the sink node $t_k$. Our formulation reads
\begin{subequations}
\begin{align}
\text{maximize} \;\; & \displaystyle \sum_{i \in \mathcal I} \sum_{n \in \sets B \cup \sets T } z^i_{n,t_{y^i}}  \label{eq:flow_reg_all_obj}\\
%%%%%%%%%%%%%%%%%%%%%%%%%%%%%%%%%binary constraints
%%%%%%%%%%%%%%%%%%%%%%%%%%%%%%%%%The flow conservation constraints
\text{subject to}\;\;
& \displaystyle \sum_{f \in \sets F}b_{nf} + p_n + \sum_{m \in \sets P(n)}p_m = 1   &\hspace{-5cm}  \forall n \in \sets B \label{eq:flow_reg_all_branch_or_predict}\\
&  p_n+\sum_{m \in \sets P(n)}p_{m} =1   &  \forall n \in \sets T \label{eq:flow_reg_all_terminal_leaf}\\
& \displaystyle z^i_{a(n),n} =  z^i_{n,\ell(n)} + z^i_{n,r(n)} + \sum_{k \in \sets K}z^i_{n,t_k}   &\hspace{-5cm}  \forall n \in \sets B, i \in \mathcal I \label{eq:flow_reg_all_conservation_internal}\\
&  \displaystyle z^i_{a(n),n} = \sum_{k \in \sets K}z^i_{n,t_k}  &\hspace{-5cm}   \forall n \in \sets T, {i \in \mathcal I} \label{eq:flow_reg_all_conservation_terminal}\\
%%%%%%%%%%%%%%%%%%%%%%%%%%%%%%%%%capacity constraints
& \displaystyle z^i_{s,1} = 1 &\hspace{-5cm} \forall i \in \mathcal I\label{eq:flow_reg_all_source}\\
&  \displaystyle z^i_{n,\ell(n)}\leq \sum_{f \in \sets F: x_{f}^i=0}b_{nf} &\hspace{-5cm} \forall n \in \sets B, i \in \mathcal I \label{eq:flow_reg_all_branch_left}\\
&  \displaystyle z^i_{n,r(n)}\leq \sum_{f \in \sets F: x_{f}^i=1}b_{nf}  &\hspace{-5cm} \forall n \in \sets B, i \in \mathcal I \label{eq:flow_reg_all_branch_right}\\
&  \displaystyle z^i_{n,t_{k}} \leq  w^n_{k} &\hspace{-5cm} \forall n \in \sets B \cup \sets T,k \in \sets K, {i \in \mathcal I}\label{eq:flow_reg_all_sink}\\
%%%% DVs
&  \displaystyle \sum_{k \in \sets K}w^n_{k} = p_n  &\hspace{-5cm}  \forall n \in \sets B \cup \sets T \label{eq:flow_reg_all_leaf_prediction}\\
% & \displaystyle \sum_{k \in \sets K}z^i_{n,t_k} = p_n &\hspace{-5cm} \forall i \in \mathcal I, n \in \sets B \cup \sets T \label{eq:flow_reg_all_sink_2}\\
&  \displaystyle w^n_{k} \in \{0,1\}  &\hspace{-5cm}   \forall n \in \sets B \cup \sets T,k \in \sets K \\
&  \displaystyle b_{nf} \in \{0,1\}  &\hspace{-5cm}   \forall n \in \sets B,f \in \sets F \\
&  \displaystyle p_{n} \in \{0,1\}  &\hspace{-5cm}   \forall n \in \sets B \cup \sets T \\
&  \displaystyle z^i_{a(n),n}, z^i_{n,t_k}\in \{0,1\}  &\hspace{-5cm}  \forall n \in \sets B \cup \sets T,i \in \sets I, k \in \sets K.
\end{align}
\label{eq:flow_reg_all}
\end{subequations}
An interpretation of the problem constraints is as follows. Constraints~\eqref{eq:flow_reg_all_branch_or_predict}-\eqref{eq:flow_reg_all_leaf_prediction} exactly mirror constraints~\eqref{eq:flow_reg_branch_or_predict}-\eqref{eq:flow_reg_leaf_prediction} in problem~\eqref{eq:flow_reg}. There are three main differences in these constraints. Flow conservation constraints~\eqref{eq:flow_reg_all_conservation_internal} and~\eqref{eq:flow_reg_all_conservation_terminal} now allow flow to be directed to any one of the sink nodes~$t_k$ from any one of the nodes $n\in \sets B \cup \sets T$. Constraint{s}~\eqref{eq:flow_reg_all_source} {ensure} that the flow incoming into the source and associated with each datapoint equals~1. Constraint{s}~\eqref{eq:flow_reg_all_sink} stipulate that a datapoint is only allowed to be directed to the sink corresponding to the class predicted at the leaf where the datapoint landed. Constraints~\eqref{eq:flow_reg_all_conservation_internal}-\eqref{eq:flow_reg_all_source} and~\eqref{eq:flow_reg_all_sink} together ensure that all datapoints get routed to the sink associated with their predicted class. As a result of these changes, sink node~$t_k$ collects \emph{all} datapoints, whether correctly classified or not, with predicted class~$k \in \sets K$. The objective \eqref{eq:flow_reg_all_obj} is updated to reflect that a datapoint is correctly classified if and only if it flows to sink $t_{y^i}$.

At a feasible solution to problem~\eqref{eq:flow_reg_all}, the quantity $\sum_{i \in \sets I} z^i_{a(n),n}$ represents the total number of datapoints that land at node~$n$. The number of datapoints of class~$k \in \sets K$ that are correctly (resp.\ incorrectly) classified is expressible as~$\sum_{i \in \sets I : y^i = k} \sum_{n \in \sets B \cup \sets T} z_{n,t_k}^i$ (resp.\ $| \{ i\in \sets I : y^i=k \} | - \sum_{i \in \sets I : y^i = k} \sum_{n \in \sets B \cup \sets T} z_{n,t_k}^i$), while the total number of datapoints for which we predict class $k$ can be written as $\sum_{i \in \sets I}\sum_{n \in \sets B \cup \sets T}z^i_{n,t_k}$.

%\delpv{$\sum_{i \in \sets I : y^i = k} \sum_{n \in \sets B \cup \sets T}\sum_{k' \in \sets K: k'\neq k} z_{n,t_{k'}}^i$)}

Problem~\eqref{eq:flow_reg_all} allows us to learn meaningful decision trees for the case of imbalanced data by modifying the objective function of the problem and/or by augmenting this nominal formulation with constraints. We detail these variants in what follows.

\paragraph{Balanced Accuracy.} A common approach to handle imbalanced datasets is to optimize the so-called \emph{balanced accuracy}, which averages the accuracy across classes. Intuitively, in the case of two classes, it corresponds to the average of the true positive and true negative rates, see e.g., \citet{mower2005prep}. It can also be viewed as the average between sensitivity and specificity. Maximizing balanced accuracy can be achieved by replacing the objective function of problem~\eqref{eq:flow_reg_all} with
$$
\frac{1}{|\sets K|}\sum_{k \in \sets K} \frac{ 1 }{ | \{ i \in \sets I : y^i=k \} | } \displaystyle \sum_{i \in \sets I : y^i = k} \sum_{n \in \sets B \cup \sets T} z_{n,t_k}^i. 
$$
Each term in the summation above represents the proportion of datapoints from class~$k$ that are correctly classified. %It can thus also be viewed as the average \emph{recall} across classes. 
Having high balanced accuracy is important when dealing with imbalanced data and correctly predicting all classes is (equally) important.

\paragraph{Worst-Case Accuracy.} As a variant to optimizing the average accuracy across classes, we propose to maximize the worst-case (minimum) accuracy. This can be achieved by replacing the objective function of problem~\eqref{eq:flow_reg_all} with
$$
\min_{k \in \sets K} \; \; \frac{ 1 }{ | \{ i \in \sets I : y^i=k \} | } \displaystyle \sum_{i \in \sets I : y^i = k} \sum_{n \in \sets B \cup \sets T} z_{n,t_k}^i. 
$$
To the best of our knowledge, this idea has not been proposed in the literature. Compared to minimizing average accuracy, this objective attaches more importance to the class that is ``worse-off.'' Having high worst-case accuracy is important when dealing with imbalanced data and correctly predicting the class that is the hardest to predict is important.
This is the case for example when diagnosing cancer where positive cases are hard to identify and having large false negative rates (i.e., low true positive rate) could cost a patient's life.
{There are also related works on robust classification where the objective happens to be the minimization of worst-case accuracy over noisy features/labels, see~\citet{bertsimas2019robust}, or over adversarial examples, see~\citet{vos2021robust}.} 

In the remaining of this section we focus on the case of binary classification where $\sets K=\{0,1\}$ and we refer to $k=1$ (resp.\ $k=0$) as the positive (resp.\ negative) class. In this setting we can discuss metrics such as recall, precision, sensitivity, and specificity more conveniently. Naturally, these definitions can be generalized to cases with more classes.

\paragraph{Constraining Recall.} An important metric in the classification problem when dealing with imbalanced datasets is recall (also referred to as sensitivity), which is the fraction of all datapoints from the positive class that are correctly identified. Guaranteeing a certain level~$C \in [0,1]$ of recall can be achieved by augmenting problem~\eqref{eq:flow_reg_all} with the constraint
$$
 \frac{ 1 }{ | \{ i \in \sets I : y^i=1 \} | } \displaystyle \sum_{i \in \sets I : y^i = 1} \sum_{n \in \sets B \cup \sets T} z_{n,t_{1}}^i \geq C. 
$$
Decreasing the number of false negatives (datapoints from the positive class that are incorrectly predicted) increases recall. Thus, in cases where false negatives can have dramatic consequences and even cost lives, such as in cancer diagnosis or when predicting landslides, guaranteeing a certain level of recall can help mitigate such risks.

\paragraph{Constraining Precision.}Our method can conveniently be used to learn decision trees that have sufficiently high precision, which is defined as the portion of correctly classified datapoints from the positive class out of all datapoints with positive predicted class. This can be achieved by augmenting problem~\eqref{eq:flow_reg_all} with the constraint 
\begin{equation} \label{eq:precision_lower_bound}
    \displaystyle \sum_{i \in \sets I: y^i = 1}\sum_{n \in \sets B \cup \sets T}z^i_{n,t_1}  \geq C \sum_{i \in \sets I}\sum_{n \in \sets B \cup \sets T}z^i_{n,t_1},
\end{equation}
where~$C \in [0,1]$ is a hyper-parameter that represents the minimum acceptable precision that can be tuned via cross-validation. Decreasing the number of false positives (datapoints from the negative class that are incorrectly classified) increases precision. Thus, constraining precision is useful in settings where having a low number of false negatives is not as important and having a low number of false positives, such as when making product recommendations.

\paragraph{Balancing Sensitivity and Specificity.} \citet{gunluk2018optimal} address the issue of imbalanced data by  maximizing sensitivity (true positive rate) in the objective, while guaranteeing a certain level of specificity (true negative rate), or the converse, instead of optimizing the total accuracy. In the case of binary classification where $\sets K = \{0,1\}$, we can constrain specificity from below by augmenting problem~\eqref{eq:flow_reg_all} with the  constraint
$$\frac{1}{|\{i \in \sets I: y^i=0\}|}\displaystyle \sum_{i \in \sets I: y^i = 0}\sum_{n \in \sets B \cup \sets T}z^i_{n,t_0} \geq C, $$
and maximize sensitivity by replacing its objective function with
$$\frac{1}{|\{i \in \sets I: y^i=1\}|}\displaystyle \sum_{i \in \sets I: y^i = 1}\sum_{n \in \sets B \cup \sets T}z^i_{n,t_1}. $$
This method is useful in settings such as infectious disease testing wherein we want to maximize the chance of correctly predicting someone to be infectious while making sure to identify non-infectious individuals with a high enough confidence. 
% {There are other works in the literature to address treatment of imbalanced data such as the work of~\citet{lin2020generalized}, where they present techniques that produce optimal decision trees over a variety of objectives such as F-score and area under the receiver operating characteristic curve (AUROC).} 

% Rudin et al provide a general framework for decision tree optimization that addresses treatment of imbalanced data and fully optimizing over continuous variables. They present techniques that produce optimal decision trees over a variety of objectives such as F-score and AUC.

\subsection{Learning Fair Decision Trees}
\label{sec:fairness}

In recent years, machine learning algorithms are increasingly being used to assist decision-making in socially sensitive, high-stakes, domains. For example, they are used to help decide who to give access to credit, benefits, and public services, see e.g.,~\citet{Byrnes2016artificialintollerance} and \citet{azizi2018designing}, to help guide policing, see e.g.,~\citet{Rudin2013policing}, or to assist with screening decisions for jobs/college admissions, see e.g.,~\citet{Miller2015algorithmhire}. Yet, decision-making systems based on standard machine learning algorithms may result in discriminative decisions as they may treat {or impact} individuals unequally based on {certain characteristics, often referred to as protected or sensitive, including but not limited to age, disability, ethnicity, gender, marital status, national origin, race, religion, and sexual orientation,} see e.g.,~\citet{Dwork:2012:FTA:2090236.2090255},~\citet{Machine_Bias}, and~\citet{barocas2016big}.

Over the last decade, so-called ``in-process'' fair machine learning algorithms have been proposed as a way to mitigate bias in standard ML. These methods incorporate a fairness notion, such as statistical parity\revision{~\citep{Dwork:2012:FTA:2090236.2090255} or equalized odds~\citep{hardt2016equality}} in the training step, either penalizing or constraining discrimination. Notably, several in-process approaches have been proposed for learning fair decision trees. The vast majority of these methods are based on heuristics.  For example, \citet{kamiran2010discrimination} and~\citet{zhang2019faht} propose to augment the splitting criterion of CART and of the Hoeffding Tree algorithm, respectively, with a regularizer to promote statistical parity and accuracy. \citet{grari2019fair} propose a fair gradient boosting algorithm to design fair decision trees that satisfy either equalized odds or statistical parity. \citet{ranzato2021fair} introduce a genetic algorithm for training decision trees that maximize both accuracy and  robustness to adversarial perturbations and accounting for individual fairness. In contrast with the aforementioned works, \citet{aghaei2019learning} propose a{n} \emph{MIO based} formulation involving a fairness regularizer in the objective aimed at mitigating disparate treatment and disparate impact, see~\citet{barocas2016big}. While this approach has proved effective relative to heuristics, it is based on a weak formulation and is therefore slow to converge, see Section~\ref{sec:Introduction}.

In this section, we discuss how our \emph{stronger formulation}~\eqref{eq:flow_reg_all} can be augmented with fairness constraints to learn \emph{optimal and fair decision trees} satisfying some of the most common fairness requirements in the literature. We refer the interested reader to the book by~\citet{barocas-hardt-narayanan} and to the survey papers of \citet{corbett2018measure},~\citet{mehrabi2019survey} and \citet{caton2020fairness} and to the references there-in for in-depth reviews of the literature on the topic of fair ML including analyses on the relative merits of various fairness metrics. 

Throughout this section, and for ease of exposition, we consider the case of binary classification where $\sets K=\{0,1\}$. We let $k=1$ correspond to the positive outcome. For example, in a hiring problem where we want to decide whether to interview someone or not, being invited for an interview is regarded as the positive outcome. We let \revision{$p^i \in \Lambda_{\rm p}$} denote the value of the protected feature\revision{(s)} of datapoint~$i$, where\revision{~$\Lambda_{\rm p}$} denotes the set of all possible levels of the protected feature\revision{(s)}. Depending on whether branching on the protected feature is allowed or not, $p^i$ can either be included or excluded as an element of the feature vector~$\bm x^i$, see e.g.,~\citet{gajane2017formalizing} and~\citet{chen2019fairness}.

\paragraph{Statistical Parity. }
A classifier satisfies \emph{statistical parity} if the probability of predicting the positive outcome is similar across all the protected groups, see~\citet{Dwork:2012:FTA:2090236.2090255}. For example in the hiring problem mentioned above, it may be appropriate to impose that the probability of receiving an interview should be similar across genders if people from different genders are equally likely to be meritorious. Augmenting model~\eqref{eq:flow_reg_all} with the following constraint{s} ensures that the decision tree learned by our MIO formulation satisfies statistical parity up to a bias $\delta$
\begin{equation*}\label{eq:statistical_parity}
     \displaystyle \left|\frac{\displaystyle \sum_{n \in \sets B \cup \sets T}\sum_{i \in  \sets I: p^i=p} z^i_{n,t_1}}{|\{i \in \sets I : p^i=p\}|} - 
     \frac{\displaystyle \sum_{n \in \sets B \cup \sets T}\sum_{i \in  \sets I: p^i=p'} z^i_{n,t_1}}{|\{i \in \sets I : p^i=p'\}|}\right| \leq \delta \quad \forall p,p' \in \Lambda_{\rm p}: p\neq p',
\end{equation*}
where the first and second term correspond to empirical estimates of the conditional probabilities of predicting the positive outcome given $p$ and $p'$, respectively.

\paragraph{Conditional Statistical Parity. }
A classifier satisfies \emph{conditional statistical parity} across protected groups if the probability of predicting the positive outcome is similar between all groups conditional on some feature{, e.g., a legitimate feature that can justify differences across protected groups}, see~\citet{corbett2017algorithmic}. We let \revision{$l^i \in \Lambda_{\rm l}$} denote the value of the feature\revision{(s)} of datapoint $i$ that can legitimize differences,~\revision{where~$\Lambda_{\rm l}$ denotes the set of all possible levels of the legitimate feature(s).} For example, in the problem of matching people experiencing homelessness to scarce housing resources, it is natural to require that the probability of receiving a resource among all individuals with the same vulnerability (a risk score) should be similar across genders, races, or other protected attribute.
By adding the following constraint{s} to model~\eqref{eq:flow_reg_all}, we can ensure our learned trees satisfy conditional statistical parity up to a bias $\delta$ given any value $l \in \Lambda_{\rm l}$
\begin{equation*}\label{eq:cond_statistical_parity}
     \displaystyle \left|\frac{\displaystyle \sum_{n \in \sets B \cup \sets T}\sum_{i \in  \sets I} \mathbb I(p^i = p ~{\land}~ l^i=l) z^i_{n,t_1}}{|\{i \in \sets I : p^i=p ~{\land}~ l^i=l\}|} - 
     \frac{\displaystyle \sum_{n \in \sets B \cup \sets T}\sum_{i \in  \sets I}\mathbb I(p^i = p' ~{\land}~ l^i=l) z^i_{n,t_1}}{|\{i \in \sets I : p^i=p' ~{\land}~ l^i=l \}|}\right| \leq \delta \quad \forall p,p' \in \Lambda_{\rm p}: p\neq p',\; l \in \Lambda_{\rm l},
\end{equation*}
where the first and second term  correspond to empirical estimates  of the conditional probabilities of predicting positive outcome given $p$ and $p'$, respectively, conditional on $l$.

\paragraph{Predictive Equality. }
A classifier satisfies \emph{predictive equality} if it results in the same false positive rates across protected groups, see~\citet{chouldechova2017fair}. For example, this fairness notion may be useful when  using machine learning to predict if a convicted person will recidivate so as to decide if it is appropriate to release them on bail, see~\citet{Machine_Bias}. Indeed, predictive equality in this context requires that, among defendants who would not have gone on to recidivate if released, detention rates should be similar across all races. Adding the following constraint{s} to model~\eqref{eq:flow_reg_all} ensures that the learned decision trees satisfy predictive equality, up to a constant $\delta$ 
\begin{equation*}
 \left | \frac{ \displaystyle \sum_{ i \in  \sets I}\sum_{n \in  \sets B \cup \sets T}\mathbb I(p^i = p ~{\land}~ y^i=0)z^i_{n,t_1}}{|\{i \in \sets I:p^i = p ~{\land}~ y^i=0 \}|} - \frac{ \displaystyle  \sum_{ i \in  \sets I} \sum_{n \in  \sets B \cup \sets T}\mathbb I(p^i = p' ~{\land}~ y^i=0)z^i_{n,t_1}}{|\{i \in \sets I:p^i = p' ~{\land}~ y^i=0 \}|}  \right | \leq \delta \quad \forall p,p' \in \Lambda_{\rm p}: p \neq p',
\end{equation*}
where the first and second terms inside the absolute value are the estimated false positive rate given~$p$ and~$p'$, respectively.

\paragraph{Equalized Odds.}
A classifier satisfies \emph{equalized odds} if the predicted outcome and protected feature are independent conditional on the outcome, see~\citet{hardt2016equality}. In other words, equalized odds requires same true positive rates and same false positive rates across protected groups. For example, in the college admissions process,  equalized odds requires that no matter the applicant's gender, if they are qualified (or unqualified), they should get admitted at equal rates.
By adding the following constraints to model~\eqref{eq:flow_reg_all} we ensure the decision trees learned by our MIO formulation satisfy equalized odds up to a constant $\delta$
\begin{equation*}\label{eq:equalized_odds}
     \displaystyle \left|\frac{\displaystyle \sum_{n \in \sets B \cup \sets T}\sum_{i \in  \sets I} \mathbb I(p^i = p ~{\land}~ y^i=k) z^i_{n,t_1}}{|\{i \in \sets I : p^i=p ~{\land}~ y^i=k\}|} - 
     \frac{\displaystyle \sum_{n \in \sets B \cup \sets T}\sum_{i \in  \sets I}\mathbb I(p^i = p' ~{\land}~ y^i=k) z^i_{n,t_1}}{|\{i \in \sets I : p^i=p' ~{\land}~ y^i=k \}|}\right| \leq \delta \quad \forall k \in \sets K, p,p' \in \Lambda_{\rm p}: p\neq p',
\end{equation*}
where the first and second terms correspond to empirical estimates of conditional probabilities of predicting the positive outcome given $(p,y=k)$ and $(p',y=k)$, respectively. If we relax the above constraints to only hold for $k=1$, we achieve \emph{equal opportunity} up to a constant~$\delta$, see~\citet{hardt2016equality}.

{All the introduced nonlinear constraints can be linearized using standard MIO techniques. We leave this task to the reader.} Our approach can also impose fairness by means of regularization and can also model more sophisticated fairness metrics such as the ones of~\citet{aghaei2019learning}. We leave these extensions to the reader.

\revision{
\subsection{Solution Approach for Generalizations}
\label{sec:solution_approach}

We now briefly discuss how to solve formulation (7) (for learning imbalanced decision trees), and formulation (11) (which tracks all datapoints through the flow graph), and their variants introduced in Sections~\ref{sec:regularization} through~\ref{sec:fairness}.

The nominal formulations~\eqref{eq:flow_reg} and~\eqref{eq:flow_reg_all} and all their variants augmented with constraints that \emph{do not} couple datapoints with one another (e.g., sparsity constraints~\eqref{eq:branching_limit} or interpretability constraints~\eqref{eq:flow_regularization_max_num_features}) can be effectively solved using Benders’ decomposition. This can be achieved by adapting Algorithm~\ref{alg:cut} from Section~\ref{sec:Benders}.
On the other hand, when these formulations are augmented with constraints that \emph{do} couple datapoints with one another (like the fairness constraints from Section~\ref{sec:fairness}), Benders' decomposition is no longer applicable. In such cases, these formulations need to be solved directly as monolithic MIO problems. 

The Benders' decomposition approach for solving formulation~\eqref{eq:flow_reg} and the corresponding variants is provided in Electronic Companion~\ref{appendix_sec:Benders}. The derivation of the Benders' decomposition for formulation~\eqref{eq:flow_reg_all} and the corresponding variants  is left for the reader.
}

%%%%%%%%%%%%%%%%%%%%%%%%%%%%%%%%%%%%%%%%%%%%%%%%%%
%%%%%%%%%%%%%%%%%%%%%%%%%%%%%%%%%%%%%%%%%%%%%%%%%%
\section{Experiments}
\label{sec:Experiments}
%%%%%%%%%%%%%%%%%%%%%%%%%%%%%%%%%%%%%%%%%%%%%%%%%%
%%%%%%%%%%%%%%%%%%%%%%%%%%%%%%%%%%%%%%%%%%%%%%%%%%
\revision{In the following section, we discuss the datasets we use in our experiments, the approaches that we compare to, the experimental setup, and our findings. More extensive numerical results are included in the electronic companion.}
\subsection{Benchmark Approaches and Datasets}
\revision{In our numerical experiments, we have two sets of experiments, one on datasets with only categorical features and one on datasets with a mixture of categorical and real-valued features. We now describe both sets of data and the approaches we compare to in each case.}
\paragraph{Datasets with only Categorical Features.} In \revision{the first part of }our experiments we use {all} twelve publicly available datasets with only categorical features from the UCI data repository~\citep{Dua:2019} as detailed in Table~\ref{tab:datasets_classification_categorical}.
We compare the flow-based formulation (\flow) given in problem~\eqref{eq:flow_reg} and its Benders' decomposition (\benders) described in Electronic Companion~\ref{appendix_sec:Benders}
to the \revision{univariate splits} formulations proposed by~\citet{bertsimas2017optimal} (\texttt{OCT}) and~\citet{verwer2019learning} (\texttt{BinOCT}). We also compare to the heuristic algorithm based on local search for solving \texttt{OCT} proposed by~\citet{bertsimas2019machine} (\texttt{LST}).
As the code used for \texttt{OCT} is not publicly available, we implemented the corresponding formulation (adapted to the case of binary data). The details of this implementation are given in Electronic Companion~\ref{appendix_sec:OCT}. 
We also used the \revision{\texttt{Python}} implementation of \texttt{LST} which is available for academic use. In \texttt{LST}, we set the complexity parameter `\texttt{cp}' to 0 which is analogous to setting $\lambda=0$.
In order to make the optimality gap comparable across approaches, we made some adjustments to the objective function of \flow{} and \benders{}, by subtracting number of total datapoints from the objective, such that, for all approaches, the objective value reflects the number of misclassified datapoints.
The results for worst-case accuracy objective discussed in Section~\ref{sec:imbalanced_datasets} can be found in Electronic Companion~\ref{appendix_sec:flow_worst}. \revision{ We also numerically analyze the strength of all formulations by looking at their LO relaxation, see Electronic Companion~\ref{appendix_sec:LO_relaxation}. Analysis of some implementation variants of \benders{}, see Remark~\ref{remark:benders_implementation}, can be found in Electronic Companion~\ref{appendix_sec:benders_variants}}

\begin{table}[]
\OneAndAHalfSpacedXII
\caption{Benchmark datasets with only categorical features, along with their number of rows ($|\sets I|$), number of features ($|\sets F|$), and number of classes ($|\sets K|$).}
\small{
\label{tab:datasets_classification_categorical}
\vskip 0.15in
\begin{center}
\begin{tabular}{lccc}
\hline
Dataset         & $|\sets I|$ & $|\sets F|$ & $|\sets K|$ \\ \hline
soybean-small   & 47          & 45          & 4           \\
monk3           & 122         & 15          & 2           \\
monk1           & 124         & 15          & 2           \\
hayes-roth      & 132         & 15          & 3           \\
monk2           & 169         & 15          & 2           \\
house-votes-84  & 232         & 16          & 2           \\
spect           & 267         & 22          & 2           \\
breast-cancer   & 277         & 38          & 2           \\
balance-scale   & 625         & 20          & 3           \\
tic-tac-toe     & 958         & 27          & 2           \\
car-evaluation & 1728        & 20          & 4           \\
kr-vs-kp        & 3196        & 38          & 2           \\ \hline
\end{tabular}%
\end{center}
}
\vskip -0.1in
\end{table}

\revision{\paragraph{Datasets with Mixed Features.} 
In the second part of our experiments, we use 28 publicly available datasets with both categorical and real-valued features from the UCI data repository as detailed in  Table~\ref{tab:datasets_classification_non_categorial}. 
In this part, we compare~\benders{} to~\texttt{OCT}. We did not include~\texttt{BinOCT} as it cannot handle real-valued features. We also did not include~\flow{} as it is outperformed by~\benders{}. To run \benders{} on these datasets, we first discretize the real-valued features into $5$ and $10$ buckets (quantiles) and then one-hot encode the discretized columns. We refer to the version of \benders{} which we run on the discretized data with 5 (resp.\ 10) buckets as \bendersFiveBuckets{} (resp.\ \bendersTenBuckets{}). As \texttt{OCT} can handle real-valued features without special encoding, we use the original format of the datasets for \texttt{OCT}. For both \texttt{OCT} and \benders{} the categorical features are one-hot encoded. 
\begin{table}[]
\OneAndAHalfSpacedXII
\caption{\revision{Benchmark datasets with mixed features, along with their number of rows ($|\sets I|$), number of features ($|\sets F|$), and number of classes ($|\sets K|$).}}
\small{
\label{tab:datasets_classification_non_categorial}
\vskip 0.15in
\begin{center}
\color{black}\begin{tabular}{lccccc}
\hline
Dataset& $|\sets I|$ & \begin{tabular}[c]{@{}c@{}} $|\sets F|$ for \\\texttt{OCT}\end{tabular} & \begin{tabular}[c]{@{}c@{}} $|\sets F|$ for \\\bendersFiveBuckets{}\end{tabular} & \begin{tabular}[c]{@{}c@{}} $|\sets F|$ for \\\bendersTenBuckets{}\end{tabular}& $|\sets K|$ \\ \hline
echocardiogram&61&8&32&61&2\\
hepatitis&80&19&43&68&2\\
fertility&100&20&28&28&2\\
iris&150&4&20&38&3\\
wine&178&13&65&130&3\\
planning-relax&182&12&60&120&2\\
breast-cancer-prognostic&194&33&164&321&2\\
parkinsons&195&22&110&218&2\\
connectionist-bench-sonar&208&60&300&600&2\\
seeds&210&7&35&70&3\\
cylinder-bands&277&257&314&370&2\\
heart-cleveland&297&22&44&68&5\\
ionosphere&351&33&157&298&2\\
thoracic-surgery&470&27&39&54&2\\
climate&540&18&90&180&2\\
breast-cancer-diagnostic&569&30&150&300&2\\
indian-liver-patient&579&10&45&88&2\\
credit-approval&653&42&64&89&2\\
blood-transfusion&748&4&20&36&2\\
diabetes&768&8&39&75&2\\
qsar-biodegradation&1055&41&139&234&2\\
banknote-authentication&1372&4&20&40&2\\
ozone-level-detection-one&1848&72&358&714&2\\
image-segmentation&2310&18&82&164&7\\
seismic-bumps&2584&19&48&73&2\\
thyroid-disease-ann-thyroid&3772&21&45&74&3\\
spambase&4601&57&108&190&2\\
wall-following-robot-2&5456&24&116&228&4\\
 \hline
\end{tabular}%
\end{center}
}
\vskip -0.1in
\end{table}

}

\subsection{Experimental Setup}
\paragraph{} For each dataset, we create~5 random splits of the data each consisting of a training set (50\%), a calibration set (25\%) used to calibrate the hyperparameters, and a test set (25\%). For each split, for each depth $d \in \{2,3,4,5\}$, and for each choice of regularization parameter $\lambda \in \{0,0.1,0.2,\dots 0.9\}$, we train a decision tree. For any given depth we calibrate $\lambda$ on the  calibration set. Having the best $\lambda$ for any given split and depth, we train a decision tree on  the union of the training and calibration sets and report the out-of-sample performance on the test set. 

All approaches are implemented in \texttt{Python} programming language and solved using Gurobi 8.1, see~\citet{gurobi2015gurobi}.
All problems are solved on a single core of SL250s Xeon CPUs and 4GB of memory with a {60-minute} time limit. Our implementation of \flow{} and \benders{} can be found online at \revision{{\url{https://github.com/D3M-Research-Group/StrongTree}}} along with instructions and is freely distributed for academic and non-profit use. \revision{In what follows, we discuss the computational efficiency and statistical performance of the different MIO approaches.} %the CPU brand is HPE

\subsection{Results on Categorical Datasets}
\label{sec:exp_categorical}

\begin{figure}[t!]
%\vskip 0.2in
\begin{center}
\includegraphics[width = 0.75 \textwidth]{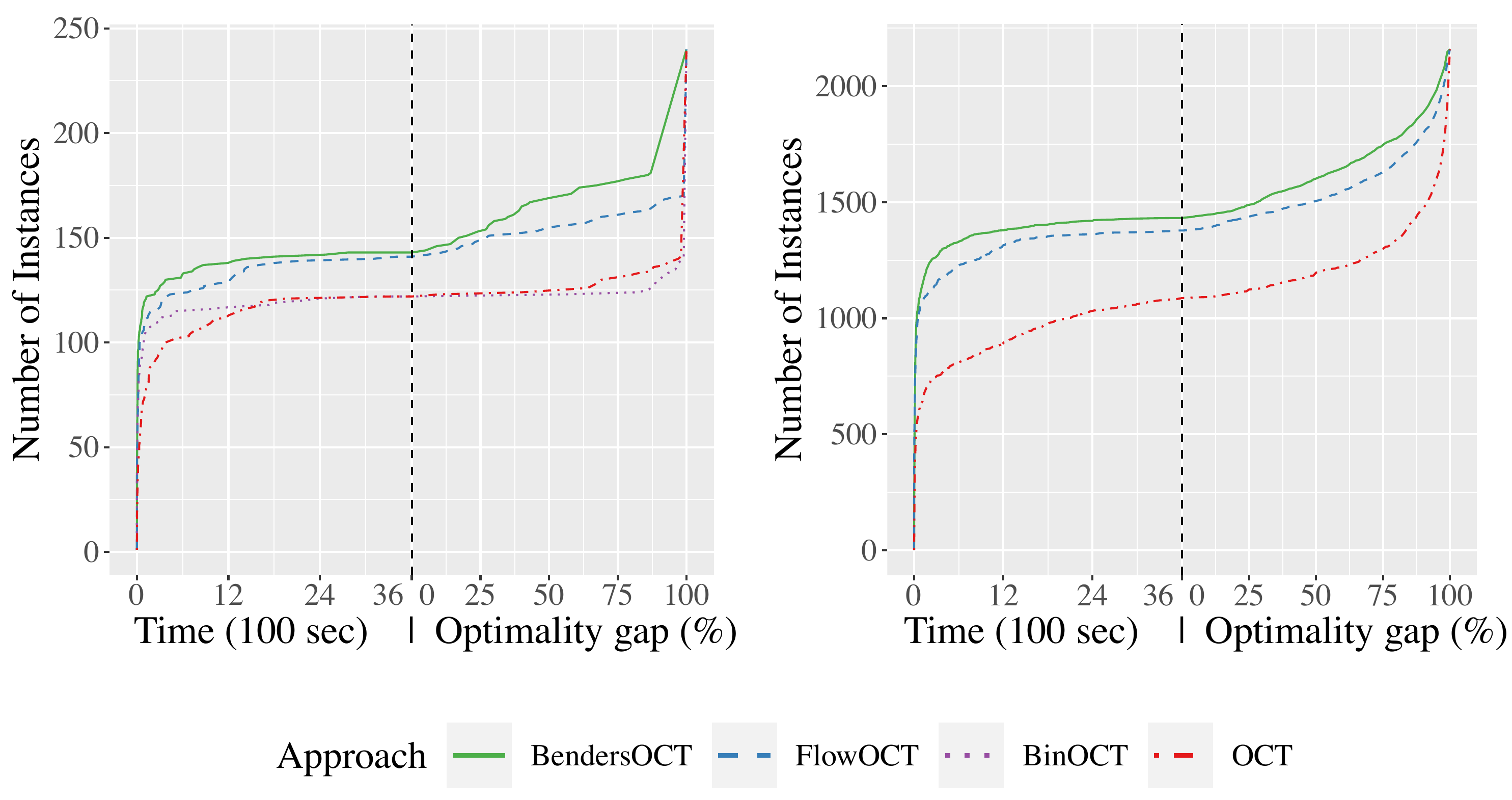}
\caption{ {The left (resp.\ right) figure shows for balanced (resp.\ imbalanced) decision trees the number of instances solved to optimality by each approach within a given time on the time axis, and the number of instances with optimality gap no larger than each given value at the time limit on the optimality gap axis.}} 
\label{fig:performace-gap}
\end{center}
\end{figure}

\paragraph{In-sample (Optimization) Performance.} Figure~\ref{fig:performace-gap} summarizes the in-sample performance of all methods. Detailed results are provided \revision{in Electronic Companion \ref{appendix_sec:ext_results}.}
From the time axis of Figure~\ref{fig:performace-gap} (left), we observe that for the case of balanced decision trees, \texttt{BinOCT} and \texttt{OCT} are able to solve 122 instances (out of 240) within the time limit, but \benders{} solves the same  number of instances in only \revision{125 seconds, resulting in a $\floor*{\frac{3600}{125}}=29\times$ speedup. Similarly, from the time axis of Figure~\ref{fig:performace-gap} (right), it can be seen that in the case of imbalanced decision trees, \texttt{OCT} is able to solve 1087 instances (out of 2400) within the time limit, while \benders{} requires only 71 seconds to do so, resulting in a $51\times$ speedup.~\texttt{BinOCT}'s implementation does not allow for imbalanced decision trees, so it is excluded from Figure~\ref{fig:performace-gap} (right).}
From the optimality gap axis of Figure~\ref{fig:performace-gap} (left), we observe that \benders{} and \flow{} both achieve better optimality gaps than either of \texttt{BinOCT} or \texttt{OCT}. Similarly, from the optimality gap axis of Figure~\ref{fig:performace-gap} (right), we observe the smaller optimality gap of \benders{} and \flow{} compared to \texttt{OCT}, when we optimize over imbalanced decision trees. 

\revision{Our experiments also demonstrate the limitations of our approaches. We observe that \benders{} successfully solves almost all the MIO instances up to the dataset ``spect'' (consisting of 267 datapoints and 22 features) within the 1-hour time limit, regardless of the chosen depth. However, when dealing with larger datasets like ``kr-vs-kp'' (consisting of 3196 datapoints and 38 features), we are unable to find the optimal solution for depths greater than 3. For instance, when considering a depth of 5 for the ``kr-vs-kp'' dataset, we observe an average optimality gap of~93\%.~\revision{It is worth noting that our approach achieves an out-of-sample accuracy of 89\% in this instance, surpassing the performance of both \texttt{BinOCT} (87\%) and \texttt{OCT} (66\%).} }

\paragraph{Out-of-sample Performance.}  \revision{Table~\ref{tab:out_of_sample_summary} summarizes the out-of-sample performance of all methods. Detailed results are reported in Electronic Companion~\ref{appendix_sec:ext_results}. From the table we observe that the better optimization performance translates to superior out-of-sample properties as well: out of 48 instances (average accuracy across 5 samples for each dataset and depth given the calibrated $\lambda$), \texttt{OCT} achieves the best out-of-sample accuracy in 7 instances (excluding ties), \texttt{BinOCT} in 8, while the new formulation \benders{} (resp.\ \flow{}) achieves the best accuracy in 9 (resp.\ 8) instances. \benders{} (resp.\ \flow) improves out-of-sample accuracy with respect to \texttt{BinOCT} and \texttt{OCT} by up to $8\%$ (resp.\ $7\%$) and $36\%$ (resp.\ $21\%$), respectively.

\begin{table}[]
\OneAndAHalfSpacedXII
\caption{\revision{The summary of the out-of-sample performance of all methods on categorical datasets}}
\small{
\begin{center}
\color{black}\begin{tabular}{l||c|c|c}
\hline
Approach & 
\begin{tabular}[c]{@{}c@{}} Best accuracy \\(out of 48, excluding ties)\end{tabular}& 
avg. accuracy & max accuracy improvement\\ \hline
\texttt{OCT}           & 7     & $0.76\pm 0.14$   &   - \\
\texttt{BinOCT}        & 8     & $0.79\pm 0.13$   &   -\\
\flow                  & 8     & $0.79\pm 0.13$   & 7\% (21\%) w.r.p. to \texttt{OCT} (\texttt{BinOCT})  \\ 
\benders               & 9     & $0.80\pm 0.14$   &    8\% (36\%) w.r.p. to \texttt{OCT} (\texttt{BinOCT}) 
\\\hline
\end{tabular}
\label{tab:out_of_sample_summary}
\end{center}}
\end{table}
}

\paragraph{Comparing with \texttt{LST}.}

We compare \benders{} (with $\lambda=0$) to the state-of-the-art approach~\texttt{LST} --which is based on local search-- on 240 MIO instances (which consist of all 12 datasets, 5 splits per dataset, and 4 different depths).
On the one hand, \texttt{LST} is much faster, requiring only seconds to find a local optimum (does not guarantee optimality). \revision{The average solving time among all 240 instances for \texttt{LST} (resp.\ \benders{}) is 0.52 (resp.\ 1539) seconds.} On the other hand, \benders{} can solve 143 of the instances to provable optimality: out of those, \texttt{LST} is able to find an optimal solution (without certificate) in 117, and produces a suboptimal~\revision{(by $1\%$ in average in-sample accuracy)} solution in the remaining 26. \revision{The average solving time among the 117 instances that both approaches solve to optimality, for \texttt{LST} (resp.\ \benders{}) is 0.26 (resp.\ 80) seconds.} Overall, out of the 240 instances (including those not solved to optimality), \benders{} produces a better solution in \revision{67} \revision{instances (by $1\%$ in average in-sample accuracy)} while \texttt{LST} outperforms \benders{} in \revision{37} \revision{instances (by $2\%$ in average in-sample accuracy)}, with the remaining \revision{136} being tied between the two methods.
Thus, we conclude that \texttt{LST} is a method able to deliver high-quality solutions very fast; however, if enough computational resources are available, the exact \benders{} is able to deliver better solutions overall. \revision{Table~\ref{tab:LST_results_summary} reports a summary of these results. Detailed results are provided in Electronic Companion~\ref{appendix_sec:ext_results}.} 

% Overall, out of the 240 instances (including those not solved to optimality), \benders{} produces a better solution in \revision{67} \revision{instances (by up to $4\%$ in in-sample accuracy)} while \texttt{LST} outperforms \benders{} in \revision{37} \revision{instances (by up to $12\%$ in in-sample accuracy)}, with the remaining \revision{136} being tied between the two methods.

\begin{table}[]
\caption{\revision{The summary of the comparison of the in-sample results of \texttt{LST} vs \benders{} on categorical datasets.}}
\OneAndAHalfSpacedXII
\small{
\begin{center}
\color{black}\begin{tabular}{c||c|c}
\hline
Metric & \texttt{LST} & \benders \\ \hline
Avg. solving time  & 0.52 s & 1539 s   \\ 
Avg. in-sample accuracy & 89\% & 89\% \\ 
Instances solved to optimality & 117 (without certificate) & 143 \\ 
Avg. solving time (117 optimal instances) & 0.26 s& 80 s\\ 
Better solutions (excluding ties) & 37  & 67  \\ \hline
\end{tabular}
\label{tab:LST_results_summary}
\end{center}
}
\end{table} %Max in-sample accuracy improvement & 12\% & 4\% \\

\revision{

\subsection{Results on Mixed-Feature Datasets}
\label{sec:exp_non_categorical}

\paragraph{In-sample Performance.}

\revision{First, we point that \texttt{OCT} results in numerical issues performance in over half of the instances tested. In our experiments, we found out that while the formulations described \cite{bertsimas2017optimal} are correct \emph{if solved on real arithmetic}, they may fail to produce optimal solutions with solvers working with numerical tolerances. In particular, in our experiments with Gurobi, we found out that in many cases the solver terminates with a solution it claims as optimal (often in seconds, with a solution that allegedly correctly classifies all points), but upon further inspection the tree described by the decision variables $(\bm{b},\bm{w})$ misclassifies most of the points. A detailed discussion on this matter, along with an example, can be found in Electronic Companion~\ref{appendix_sec:oct_numerical_issue}. Thus, we do not report computational times of OCT (since several instances that are solved very fast are in fact considerably suboptimal), but we do report out-of-sample performance corresponding to the solution found by the solver.}

Figure~\ref{fig:performace-gap-mixed-feature} summarizes the in-sample performance of \bendersFiveBuckets{} and \bendersTenBuckets. Detailed results can be found in Electronic Companion~\ref{appendix_sec:ext_results}. 
% \delag{\texttt{OCT} is not included in this analysis as it suffers from numerical instability because of the ``little-$m$'' constraints which result in a discrepancy between the optimal objective value and the actual in-sample accuracy. A detailed discussion on this matter can be found in Electronic Companion~\ref{appendix_sec:oct_numerical_issue}.}
From the figure, we observe that adding more buckets to the discretization process increases the computational time. This outcome is expected since adding more features leads to a linear growth in the size of the MIO formulation.
\revision{When comparing average in-sample accuracy, using 10 buckets does not offer an advantage over 5 buckets; in fact, it slightly underperforms. For balanced (resp.\ imbalanced) decision trees, utilizing 5 buckets improves in-sample accuracy from 0.8872 to 0.8914 (resp.\ from 0.8594 to 0.8614), while reducing computational time by 7\% (resp.\ 9\%).} \revision{Note that using 10 buckets in large instances may hamper solvers, and thus using just five buckets results in better feasible solutions found within the time limit.}

\begin{figure}[t!]
%\vskip 0.2in
\begin{center}
\includegraphics[width = 0.75 \textwidth]{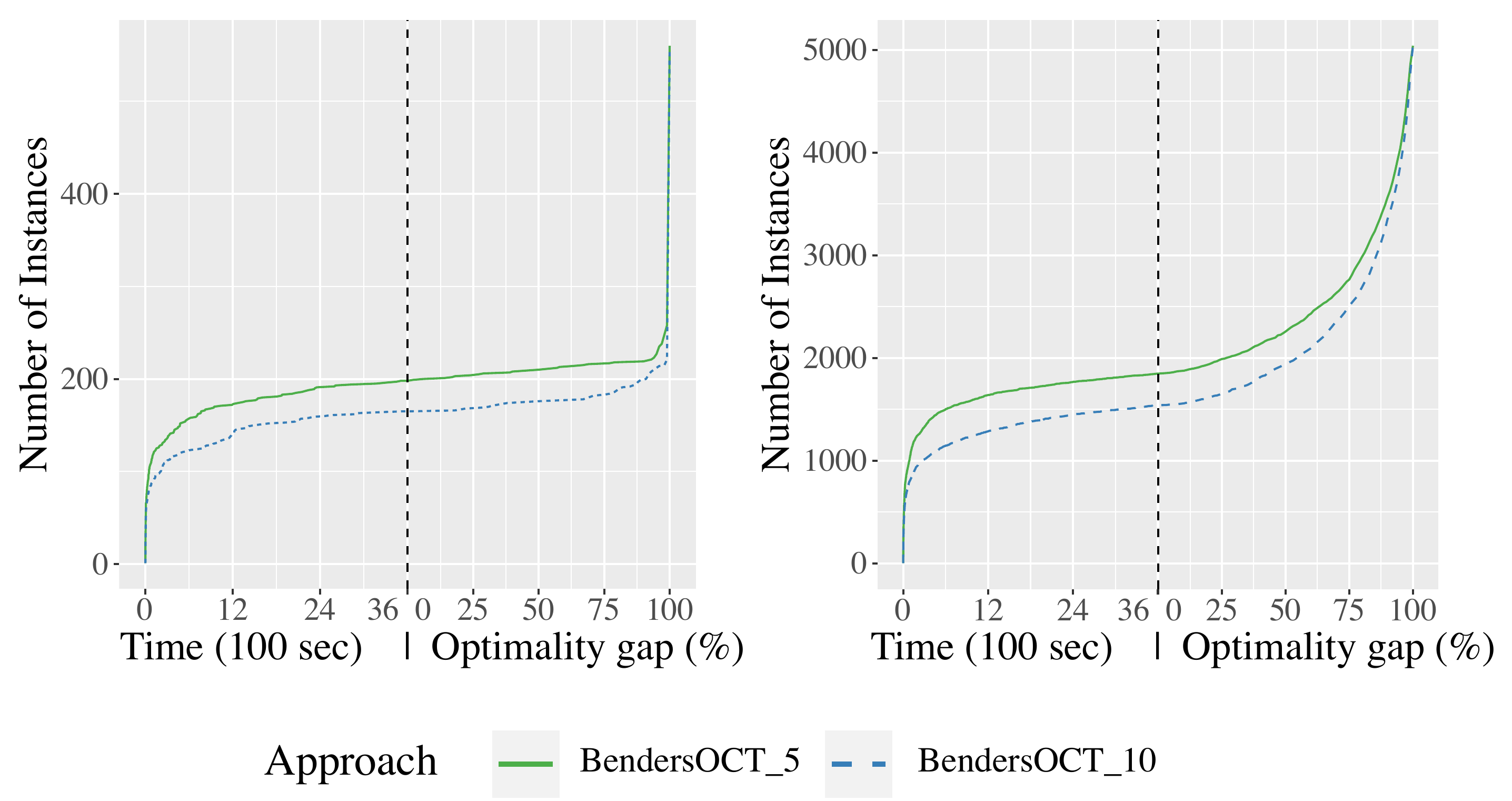}
\caption{ {The left (resp.\ right) figure shows for balanced (resp.\ imbalanced) decision trees the number of instances solved to optimality by each approach on the time axis, and the number of instances with optimality gap no larger than each given value at the time limit on the optimality gap axis. \texttt{OCT} is not included in this figure because of its numerical instabilities due to having ``little-$m$'' constraints which caused discrepancy between the optimal objective value and the actual in-sample accuracy. Refer to Electronic Companion~\ref{appendix_sec:oct_numerical_issue} for further information.}} 
\label{fig:performace-gap-mixed-feature}
\end{center}
\end{figure}

\paragraph{Out-of-sample Performance.}

Table~\ref{tab:mixed_features_out_of_sample_summary} summarizes the out-of-sample results on the mixed-features datasets. Detailed results are reported in Electronic Companion~\ref{appendix_sec:ext_results}. We see that in terms of out-of-sample accuracy, out of 112 instances (average accuracy across 5 samples for each dataset and depth given the calibrated $\lambda$) \texttt{OCT} is the best method in 25 instances (excluding ties), while \bendersFiveBuckets{} and \bendersTenBuckets{} are better in 71. Furthermore, \bendersFiveBuckets{} (\bendersTenBuckets) improves out-of-sample accuracy with respect to \texttt{OCT} by up to $50\%$ ($51\%$). So despite the fact that we are losing some information by discretizing the features, \benders{} can still output higher quality solutions compared to \texttt{OCT}. \revision{Moreover, the gains achieved by \bendersTenBuckets{} over \bendersFiveBuckets{} are on average small, suggesting that a coarse discretization is sufficient in practice.} One could justify these findings by interpreting the discretization as some form of regularization which helps avoiding overfitting and yields better out-of-sample performance.

\begin{table}[]
\OneAndAHalfSpacedXII
\caption{\revision{The summary of the out-of-sample performance of various approaches on mixed-feature datasets given the calibrated $\lambda$.}}
\small{
\begin{center}
\color{black}\begin{tabular}{l||c|c|c}
\hline
Approach & 
\begin{tabular}[c]{@{}c@{}} Best accuracy \\(out of 112, excluding ties)\end{tabular}& 
avg. accuracy & max accuracy improvement\\ \hline
\texttt{OCT}           & 25     & $0.75\pm 0.18$   &   -    \\
\bendersFiveBuckets{}  & 37     & $0.81\pm 0.12$   &   50\% w.r.p. to \texttt{OCT}    \\
\bendersTenBuckets{}   & 34     & $0.81\pm 0.13$   &    51\% w.r.p. to \texttt{OCT}  \\ \hline
\end{tabular}
\label{tab:mixed_features_out_of_sample_summary}
\end{center}}
\end{table}

% \begin{tabular}[c]{@{}c@{}} 6\% w.r.p. to \bendersFiveBuckets{} \\
% 8\% w.r.p. to \bendersTenBuckets{} \end{tabular}
}
\section{Conclusion}
\label{sec:conclusion}

We proposed a new MIO formulation for classification trees \revision{with univariate splits} with a stronger LO relaxation than {state-of-the-art} MIO based approaches. We also provided a tailored Benders' decomposition method to speed-up the computations. Our experiments reveal better computational performance than {state-of-the-art} methods including \revision{$29$ (resp.\ 51)} times speedup when we optimize over balanced (resp.\ imbalanced) trees. These also translated to improved {out-of-sample} performance up to \revision{$36\%$}. We showcase the modeling power of our framework to deal with imbalanced datasets and to design interpretable and fair decision trees. {
Our model can also act as a building block for more sophisticated predictive and prescriptive tasks. For example, variants of our method can be used to learn optimal prescriptive trees from observational data, see~\citet{jo2021learning}, or optimal robust classification trees, see~\citet{justin2021}.}

% \newpage
\phantom{dsf}
% \newpage

% Acknowledgments here
\ACKNOWLEDGMENT{P.\ Vayanos and S.\ Aghaei gratefully acknowledge support from the Hilton C.\ Foundation, the Homeless Policy Research Institute, the Home for Good foundation under the ``C.E.S.\ Triage Tool Research \& Refinement'' grant{. P.\ Vayanos is funded in part by the National Science Foundation, under CAREER grant 2046230. She is grateful for this support.} A.\ G\'omez is funded in part by the National Science Foundation under grants 1930582 and 2006762. \revision{We would like to express our sincere thanks to the anonymous reviewers for their valuable and constructive feedback, which greatly improved the quality of this paper.}}

\bibliographystyle{informs2014} % outcomment this and next line in Case 1
\bibliography{bib.bib} % if more than one, comma separated

\newpage

%% Here starts the e-companion (EC)
%%%%%%%%%%%%%%%%%%%%%%%%%%%%%%%%%%%%%%%%%%%%%%%%%%%%%%%%%%
\ECSwitch

\ECHead{E-Companion}

%%%%%%%%%%%%%%%%%%%%%%%%%%%%%%%%%%%%%%%%%%%%%%%%%%
%%%%%%%%%%%%%%%%%%%%%%%%%%%%%%%%%%%%%%%%%%%%%%%%%%
\section{Proof of Theorem~\ref{theo:facet}}
\label{appendix_sec:polyhedral}
%%%%%%%%%%%%%%%%%%%%%%%%%%%%%%%%%%%%%%%%%%%%%%%%%%
%%%%%%%%%%%%%%%%%%%%%%%%%%%%%%%%%%%%%%%%%%%%%%%%%%

% \proof{Proof.}
The proof proceeds in three steps. We fix $i \in \sets I$. We derive the specific structure of the cuts associated with datapoint~$i$ generated by our procedure. We then provide $| {\mathcal B} \times \mathcal F| + |\mathcal L \times \mathcal K| + |\mathcal I|$ affinely independent points that lie in $\textrm{conv}(\sets H_{\leq})$ and at each of which the cut generated holds with equality. Since the choice of $i \in \sets I$ is arbitrary and since the cuts generated by our procedure are valid (by construction), this will conclude the proof.

Given a set {$\sets M$ and a point $m\in \sets M$, we use $\sets M\setminus m$} as a shorthand for {$\sets M\setminus\{m\}$}. Finally, {with a slight abuse of notation,} we let $\bm e_{ij}$ be a vector (whose dimensions will be clear from the context) with a $1$ in coordinate $(i,j)$ and $0$ elsewhere. 

Fix $i \in \sets I$ and let $(\bar{\bm b},\bar{\bm w}, \bar{\bm g}) \in \mathcal{H}_\leq$ be integral. Given $j\in \sets I$ (possibly different from $i$), let $l(j)\in \sets L$ be the leaf node of the tree defined by $(\bar{\bm b},\bar{\bm w})$ that datapoint~$j$ is assigned to. Given $n\in \sets B$, let $f(n)\in \sets F$ be the feature selected for branching at node~$n$ under $(\bar{\bm b},\bar{\bm w})$, i.e., $\bar b_{nf(n)}=1$.

We now derive the structure of the cuts~\eqref{eq:master2_benders_cut} generated by Algorithm~\ref{alg:cut} when $(\bar{\bm b},\bar{\bm w},\bar{\bm g})$ is input. A minimum cut is returned by Algorithm~\ref{alg:cut} if and only if~$s$ and~$t$ belong to different connected components in the graph $\sets G^i(\bar{\bm b},\bar{\bm w})$. Under this assumption, the connected component $\sets S$ constructed in Algorithm~\ref{alg:cut} forms a path from~$s$ to~$n_d = l(i) \in \sets L$, i.e., $\sets S=\{s,n_1,n_2,\ldots,n_d\}$. The cut-set $\sets C(\sets S)$ then corresponds to the arcs adjacent to nodes in~$\sets S$ that do not belong to the path formed by~$\sets S$. Therefore, the cut~\eqref{eq:master2_benders_cut} returned by Algorithm~\ref{alg:cut} reads
\begin{equation}
    g_i \; \leq \; w^{l(i)}_{y^i} +  \sum_{n \in \sets S} \sum_{ \begin{smallmatrix} f \in \sets F : \\  x_f^i \neq x_{f(n)}^i \end{smallmatrix} } b_{nf}.
    \label{eq:facet}
\end{equation}

Next, we give $|\sets B\times \sets F|+|\sets L\times \sets K|+|\sets I|$ affinely independent points in $\sets H_{\leq}$ for which~\eqref{eq:facet} holds with equality. Given a vector $\bm b\in \{0,1\}^{|\sets B| \cdot |\sets F|}$, we let $\bm b_{\sets S}$ (resp.\ $\bm b_{\sets B\setminus \sets S}$) collect those elements of~$\bm b$ whose first index is $n\in \sets S$ (resp.\ $n \notin \sets S$). 

 We now describe the points, which are also summarized in Table~\ref{tab:affine}, and argue that the points belong to $\sets H_{\leq}$ and that inequality \eqref{eq:facet} is active. All the $|\sets B\times \sets F|+|\sets L\times \sets K|+|\sets I|$ points  are affinely independent, since each differs from all the previously introduced points in at least one (new) coordinate. 

\begin{table*}[h!]
\begin{center}
\OneAndAHalfSpacedXII
\small{
	\caption{Companion table for the proof of Theorem~\ref{theo:facet}: list of affinely independent points that lie on the cut generated by inputting $i \in \sets I$ and $(\bar{\bm b},\bar{ \bm w},\bar{ \bm g})$ in Algorithm~\ref{alg:cut}.}
	\label{tab:affine}
	\begin{tabular}{c|l|r c c c c }
		\hline
		\multirow{2}{*}{\#} & \multirow{2}{*}{condition} \hfill dim$=$ & &$|\sets S|\cdot |\sets F|$ & $|\sets B\setminus \sets S|\cdot |\sets F|$&$|\sets L|\cdot |\sets K|$& $|\sets I|$\\
		& \hfill sol$=$ & & $\bm b_{\sets S}$ & $\bm b_{\sets B\setminus \sets S}$ & $\bm w$&$\bm g$\\
		\hline
		&&&&&&\\
		\textit{1} & ``baseline'' point  && $\bar{\bm b}_\sets S$ & 0 & 0 & 0 \\
		&&&&&&\\
		\textit{2} & $n\in \sets L,k\in \sets K\setminus y^i$&& $\bar{\bm b}_{\sets S}$ & 0 &  $\bm{e}_{nk}$ & 0 \\
		\textit{3} & $n\in \sets L\setminus l(i)$&  & $\bar{\bm b}_{\sets S}$ & 0 &  $\bm{e}_{ny^i}$  & 0\\
		\textit{4} & $n=l(i)$ &  & $\bar{\bm b}_{\sets S}$ & 0 & $\bm{e}_{l(i) y^i}$ & $\bm{e}_{i}$ \\
		&&&&&&\\
		\textit{5} & $n\in \sets B\setminus \sets S,f\in \sets F$&& $\bar{\bm b}_{\sets S}$ & $\bm{e}_{nf}$ & 0 & 0 \\
		&&&&&&\\
		\textit{6} & $n\in \sets S$& & $\bar{\bm b}_{\sets S}-\bm{e}_{nf(n)}$ & 0 & 0 & 0 \\
		\textit{7} & $n\in \sets S,f\in \sets F:f\neq f(n),x_f^i=x_{f(n)}^i$&  & $\bar{\bm b}_{\sets S}-\bm{e}_{nf(n)}+\bm{e}_{nf}$ & 0 & 0 & 0\\
		\textit{8} & $n\in \sets S,f\in \sets F:f\neq f(n),x_f^i\neq x_{f(n)}^i$&  & $\bar{\bm b}_{\sets S}-\bm{e}_{nf(n)}+\bm{e}_{nf}$ & 0 & $\displaystyle\sum_{n\in \sets L: n\neq l(i)}\bm{e}_{ny^i}$ & $\bm{e}_i$\\
		&&&&&&\\
		\textit{9} & $j\in \sets I\setminus i: y^j\neq y^i$ && $\bar{\bm b}_{\sets S}$ & $\bar{\bm b}_{\sets B\setminus \sets S}$ & $\bm{e}_{l(j)y^j}$ & $\bm{e}_j$ \\
		\textit{10} & $j\in \sets I\setminus i: y^j=y^i$, $l(j)\neq l(i)$& & $\bar{\bm b}_{\sets S}$ & $\bar{\bm b}_{\sets B\setminus \sets S}$ & $\bm{e}_{l(j)y^j}$ & $\bm{e}_j$\\
		\textit{11} & $j\in \sets I\setminus i: y^j=y^i$, $l(j)= l(i)$& & $\bar{\bm b}_{\sets S}$ & $\bar{\bm b}_{\sets B\setminus \sets S}$ & $\bm{e}_{l(i)y^i}$ & $\bm{e}_i+\bm{e}_j$\\
		\hline
	\end{tabular}
	}
	\end{center}
\end{table*}

\begin{description}
	\item[1] \hspace{0.25cm} One point that is a ``baseline'' point; all other points are variants of it. It is given by $\bm b_{\sets S}=\bar{\bm b}_{\sets S}$, $\bm b_{\sets B\setminus \sets S} =0$, $\bm w=0$ and $\bm g=0$ and corresponds to selecting the features to branch on according to~$\bar{\bm b}$ for nodes in $\sets S$ and setting all remaining variables to $0$. The baseline point belongs to $\sets H_\leq$ and constraint~\eqref{eq:facet} is active at this point.
	\item[2-4] $|\sets L|\times |\sets K|$ points obtained from the baseline point by varying the $w$ coordinates and adjusting $\bm g$ as necessary to ensure~\eqref{eq:facet} remains active: \textit{2:}~$|\sets L|\times (|\sets K|-1)$ points, each associated with a leaf $n \in \sets L$ and class $k \in \sets K : k\neq y^i$, where the label of leaf $n$ is changed to $k$. \textit{3:}~$|\sets L|-1$ points, each associated with a leaf $n \in \sets L : n \neq l(i)$, where the class label of~$n$ is changed to~$y^i$. \textit{4:}~One point where the class label of leaf $l(i)$ is set to~$y^i$, allowing for correct classification of datapoint~$i$; in this case, the value of the right-hand side (rhs) of \eqref{eq:facet} is 1, and we set $g^i=1$ to ensure the cut~\eqref{eq:facet} remains active.
	\item[5] \hspace{0.25cm} $|\sets B\setminus \sets S|\times |\sets F|$ points obtained from the baseline point by varying the~$\bm b_{\sets B \backslash \sets S}$ coordinates, that is branching decisions made at nodes outside of the path $\mathcal{S}$. Each point is associated with a node $n\in \sets B \backslash \sets S$ and feature $f \in \sets F$ and is obtained by changing the decision to branch on feature $f$ and node $n$ to 1. As those branching decisions do not impact the routing of datapoint~$i$, the value of the rhs of inequality~\eqref{eq:facet} remains unchanged and the inequality stays active.
	\item[6-8] $| \sets S |\times |\sets F|$ points, obtained from the baseline point by varying the $\bm{b}_\sets S$ coordinates (that is, the branching decisions in the path $\mathcal{S}$ used by datapoint $i$) and adjusting~$\bm w$ and~$\bm g$ as necessary to guarantee feasibility of the resulting point and to ensure that~\eqref{eq:facet} stays active.  \textit{6:} $| \sets S |$ points, each associated with a node $n \in \sets S$ obtained by not branching on feature $f(n)$ at node $n$ (nor on any other feature), resulting in a ``dead-end'' node. The value of the rhs of~\eqref{eq:facet} is unchanged in this case and the inequality remains active. \textit{7-8:} $| \sets S |\times (|\sets F|-1)$ points, each associated with a node $n \in \sets S$ and feature $f \neq f(n)$. 
	\textit{7:}~If the branching decision $f(n)$ at node $n$ is replaced with a branching decision that results in the same path for datapoint~$i$, i.e., if $x_f^i=x_{f(n)}^i$, it is possible to swap those decisions without affecting the value of the rhs in inequality~\eqref{eq:facet}. \textit{8:}~If a feature that causes~$i$ to change paths is chosen for branching, i.e., if $x_f^i \neq x_{f(n)}^i$, then the value of the rhs of~\eqref{eq:facet} is increased by 1, and we set $g^i=1$ to ensure the inequality remains active; to guarantee feasibility of the resulting point, we label each leaf node except for $l(i)$ with the class $y^i$, which does not affect inequality~\eqref{eq:facet}. 
	\item[9-11] $|{\sets I}|-1$ points, one for each $j\in \sets I \setminus \{ i \}$, where point $j$ is correctly classified. We let~$\bm{b}=\bar{\bm b} $ (that is, all branching decisions coincide with $\bar{\bm b}$, both for nodes in path~$\mathcal{S}$ and elsewhere), and adjusting~$\bm w$ and~$\bm g$ as necessary. \textit{9:}~If datapoint~$j$ has a different class than datapoint~$i$ ($y^j \neq y^i$), we label the leaf node ~$j$ is routed to with the class of~$j$, i.e., $w_{l(j)y^j}=1$. The value of the rhs of~\eqref{eq:facet} is unaffected and the inequality remains active.
	\textit{10:}~If datapoint~$j$ has the same class as datapoint~$i$ but is routed to a different leaf than $i$, an argument paralleling that in ~\textit{9} can be made. \textit{11:}~If datapoint~$j$ has the same class as datapoint~$i$ and is routed to the same leaf $l(i)$, we label $l(i)$ with the class of $y^i=y^j$ and set $g^j=1$; the value of the rhs of~\eqref{eq:facet} increases by~$1$. Thus, we set also correctly classify datapoint $i$ by setting $g^i=1$ to ensure that \eqref{eq:facet} is active.
\end{description}
 This concludes the proof.%\halmos
% \endproof

%%%%%%%%%%%%%%%%%%%%%%%%%%%%%%%%%%%%%%%%%%%%%%%%%%
%%%%%%%%%%%%%%%%%%%%%%%%%%%%%%%%%%%%%%%%%%%%%%%%%%
\section{OCT}
\label{appendix_sec:OCT}

%%%%%%%%%%%%%%%%%%%%%%%%%%%%%%%%%%%%%%%%%%%%%%%%%%
%%%%%%%%%%%%%%%%%%%%%%%%%%%%%%%%%%%%%%%%%%%%%%%%%%
In this section, we provide a simplified version of the formulation of \citet{bertsimas2017optimal} \revision{(formulation (24) in their paper)} specialized to the case of binary data.

\begin{figure}[ht]
\vskip 0.2in
\begin{center}
\centerline{\includegraphics[width=0.5\columnwidth]{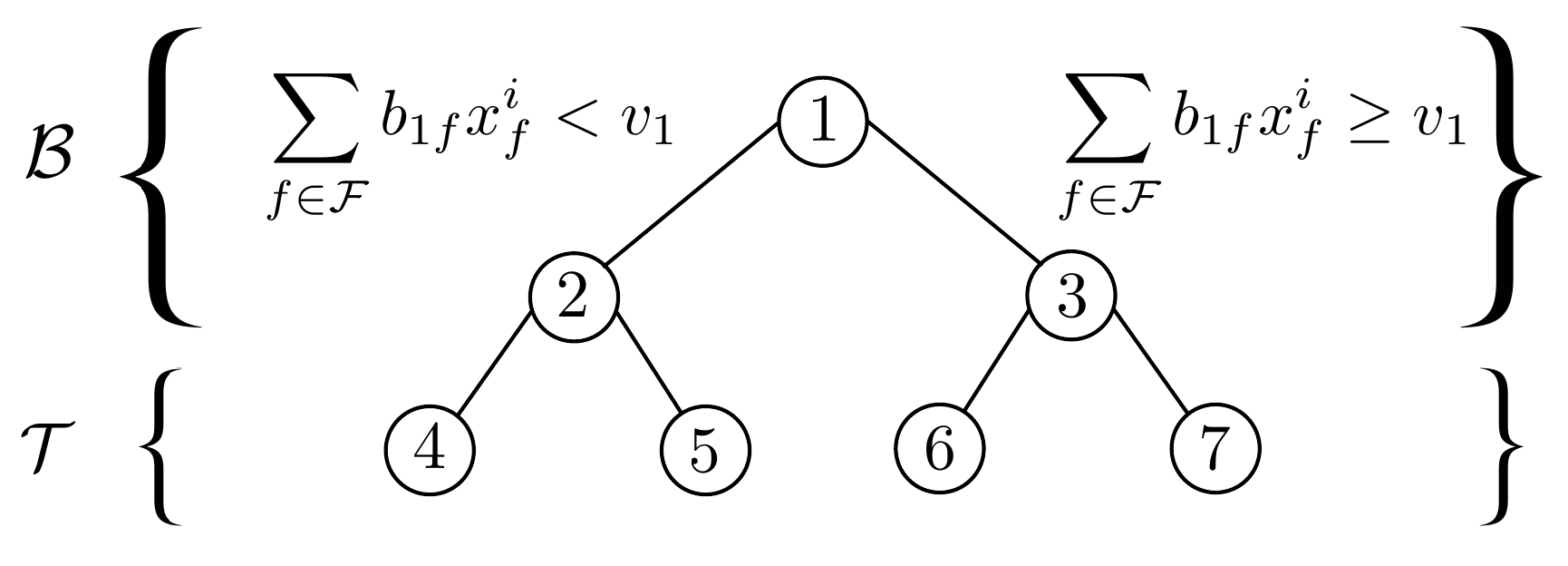}}
\caption{A classification tree of depth 2}
\label{fig:sample_tree_B}
\end{center}
\vskip -0.2in
\end{figure}

We start by introducing the notation that is used in the formulation. Let $\sets B$ and $\sets T$ denote the sets of all branching and terminal nodes in the tree structure. For each node $n \in \sets B \cup \sets T \backslash \{1\}$, $a(n)$ refers to the \revision{\textit{parent}} of node $n$. We let $\sets {PL}(n)$ (resp.\ $\sets {PR}(n)$) denote the set of ancestors of $n$ whose left (resp.\ right) branch has been followed on the path from the root node to $n$. In particular, $\sets P(n) = \sets{PL}(n) \cup \sets{PR}(n)$.

Let $b_{nf}$ be a binary decision variable where $b_{nf}=1$ if and only if feature $f$ is used for branching at node $n$.
For each  datapoint $i$ at node $n \in \sets B$, a test $\sum_{f \in \sets F}b_{nf}x^i_f < v_n$ is performed where $v_n \in \mathbb{R}$ is a decision variable representing the cut-off value of the test. If datapoint $i$ passes the test it follows the left branch; otherwise, it follows the right one. Let $p_n=1$ if and only if node $n$ applies a split, that is, if it is not a leaf.
To track each datapoint $i$ through the tree, the decision variable $ \zeta_{a(n),n}^i$ is introduced, where $\zeta_{a(n),n}^i = 1$ if and only if datapoint $i$ is routed to node $n$.

Let $Q_{nk}$ to be the number of datapoints of class $k$ assigned to leaf node $n$ and $Q_n$ to be the total number of datapoints in leaf node $n \in \sets T$. We denote by $w^n_{k}$  the prediction at leaf node $n$, where $w^n_{k}=1$ if and only if the predicted label at node $n$ is $k \in \sets K$.
Finally, we let $L_n$ denote the number of missclassified datapoints at node $n$. With this notation, the formulation of~\citet{bertsimas2017optimal} is expressible as
\begin{subequations}
\begin{align}
\text{maximize} \;\;&  \displaystyle (1-\lambda)\left(|\sets I|-\sum_{n \in \sets T}L_n\right) - \lambda \sum_{n \in \sets B}p_n \label{eq:OCT_a}\\
%%%%%%%%%%%%%%%%%%%%%%%%%%%%%%%%%The flow conservation constraints
\text{subject to} \; \; & \displaystyle L_n \geq Q_n - Q_{nk} - |\sets I|(1-w^n_k) & \vspace{-5cm} \forall k \in \sets K, n \in \sets T \label{eq:OCT_b}\\
& \displaystyle L_n \leq Q_n - Q_{nk} + |\sets I|w^n_k & \vspace{-5cm}  \forall k \in \sets K, n \in \sets T \label{eq:OCT_c}\\
&  \displaystyle Q_{nk}= \sum_{\begin{smallmatrix} i\in \sets I:\\ y^i=k \end{smallmatrix}} \zeta_{a(n),n}^i & \vspace{-5cm} \forall k \in \sets K, n \in \sets T \label{eq:OCT_d}\\
&  Q_n=\sum_{i \in \sets I} \zeta_{a(n),n}^i & \vspace{-5cm}  \forall n \in \sets T \label{eq:OCT_e}\\
&  l_n= \sum_{k \in \sets K}w^n_k & \vspace{-5cm} \forall n \in \sets T \label{eq:OCT_f}\\
&   \zeta_{a(n),n}^i \leq l_n & \vspace{-5cm} \forall n \in \sets T \label{eq:OCT_g}\\
&  \sum_{n \in \sets T} \zeta_{a(n),n}^i = 1 & \vspace{-5cm} \forall i \in \sets I \label{eq:OCT_h}\\
&    \sum_{f \in \sets F}b_{mf}x^i_f \geq v_m + \zeta_{a(n),n}^i -1 & \vspace{-5cm} \forall i \in \sets I, n \in \sets T, m \in \sets {PR}(n)  \label{eq:OCT_i}\\
&   \sum_{f \in \sets F}b_{mf}x^i_f \leq v_m - 2 \zeta_{a(n),n}^i +1 & \vspace{-5cm} \forall i \in \sets I, n \in \sets T, m \in \sets {PL}(n)  \label{eq:OCT_j}\\
&  \sum_{f \in \sets F}b_{nf} = p_n & \vspace{-5cm} \forall n \in \sets B \label{eq:OCT_k}\\
&   0 \leq v_n \leq p_n & \vspace{-5cm} \forall n \in \sets B \label{eq:OCT_l}\\
&   p_n\leq p_{a(n)} & \vspace{-5cm}  \forall n \in \sets B\backslash\{1\} \label{eq:OCT_m}\\
&   z^i_n, l_n \in \{0,1\}  & \vspace{-5cm} \forall i \in \sets I, n \in \sets T \label{eq:OCT_n}\\
&   b_{nf}, p_n \in \{0,1\} & \vspace{-5cm} \forall f \in \sets F, n \in \sets B \label{eq:OCT_o},
\end{align}
\label{eq:OCT}
\end{subequations}
where $\lambda \in [0,1]$ is a regularization term. The objective~\eqref{eq:OCT_a} maximizes the total number of correctly classified datapoints $|\sets I|-\sum_{n \in \sets T}L_n$ while minimizing the number of splits $\sum_{n \in \sets B}p_n$.
Constraints~\eqref{eq:OCT_b} and~\eqref{eq:OCT_c} {define} the number of missclassified datapoints at each node $n$. Constraints~\eqref{eq:OCT_d} and~\eqref{eq:OCT_e} give the definitions of $Q_{nk}$ and $Q_n$, respectively.
Constraints~\eqref{eq:OCT_f}-\eqref{eq:OCT_g}, enforce that if a terminal node $n$ does not have an assigned class label, no datapoint should land in that node. Constraint~\eqref{eq:OCT_h} makes sure that each datapoint~$i$ is assigned to exactly one of the terminal nodes. 
Constraint~\eqref{eq:OCT_i} implies that if datapoint $i$ is assigned to node $n$, it should take the right branch for all ancestors of $n$ belonging to $\sets {PR}(n)$.  Similarly, constraint~\eqref{eq:OCT_j} implies that if datapoint $i$ is assigned to node $n$, it should take the left branch for all ancestors of $n$ belonging to $\sets {PL}(n)$.
Constraint~\eqref{eq:OCT_k} enforces that if the tree branches at node $n$, it should branch on exactly one of the features $f \in \sets F$. 
Constraint~\eqref{eq:OCT_l} implies that if the tree does not branch at a node, all datapoints going through this node would take the right branch. Finally, constraint~\eqref{eq:OCT_m} makes sure that if the tree does not branch at node $n$ it cannot branch on any of the descendants of the node. We note that \eqref{eq:OCT} is slightly different from the original formulation of \citet{bertsimas2017optimal}. Indeed, the objective function \eqref{eq:OCT_a} maximizes correctly classified points instead of minimizing the number of missclassified datapoints--the two are clearly equivalent since the later can be obtained from the former by subtracting it from the number of datapoints $|\sets I|)$. Moreover, we have omitted a constraint similar to \eqref{eq:flow_regularization_min_datapoints_per_leaf}, as we do not use it in our computations in Section~\ref{sec:Experiments}. Unlike the original formulation in~\citet{bertsimas2017optimal}, the ``big $M$'' and ``little $m$'' constants in \revision{constraints~\eqref{eq:OCT_j}} are not directly visible since we have assumed all the features to be binary. 

\revision{
\section{OCT's Numerical Issues}
\label{appendix_sec:oct_numerical_issue}
In this section we show, by means of an example, that, for the case of real-valued features, the ``little-$m$'' constraints in the formulation of~\citet{bertsimas2017optimal} can cause numerical instabilities.

For the case of datasets with real-valued features, constraints~\eqref{eq:OCT_i} and~\eqref{eq:OCT_j} read as follows:
\begin{subequations}
\begin{align}
&\sum_{f \in \sets F}b_{mf}x^i_f \geq v_m - (1-\zeta_{a(n),n}^i) & \vspace{-5cm} \forall i \in \sets I, n \in \sets T, m \in \sets {PR}(n)  \label{eq:OCT_go_right}\\
&\sum_{f \in \sets F}b_{mf}x^i_f +\epsilon_{\min}\leq v_m + (1+\epsilon_{\max}) (1-\zeta_{a(n),n}^i) & \vspace{-5cm} \forall i \in \sets I, n \in \sets T, m \in \sets {PL}(n),  \label{eq:OCT_go_left}
\end{align}
\end{subequations}
where $\epsilon_{\min}$ (resp.\ $\epsilon_{\max}$) is defined as $\min_{f \in \sets F}(\epsilon_f)$ (resp.\ $\max_{f \in \sets F}(\epsilon_f)$), where
$$\epsilon_f := \{x_f^{(i+1)} - x_f^{(i)} \; | \; x_f^{(i+1)} \neq x_f^{(i)}, \; i = 1, \dots, n-1\} \quad \forall f \in \sets F,$$
and $x^{(i+1)}_f$ is the $i$th largest value taken by feature $f$ in the data. Note that $\epsilon_{\min}$ (resp.\ $\epsilon_{\max}$) represents the largest (resp.\ smallest) possible value that does not impact the feasibility of any valid solution to the problem.
Consider the ``ionosphere'' dataset, see Table~\ref{tab:in_sample_non_categorical_data_no_reg_part_2}. For this dataset, $\epsilon_{\min} = 4.99e-6$ and $\epsilon_{\max} = 1$. \texttt{OCT} outputs the decision tree shown in Figure~\ref{fig:OCT_issue}. In this instance, for node $m=3$ and datapoint $i=175$, constraints~\eqref{eq:OCT_go_right} and~\eqref{eq:OCT_go_left} read:
\begin{align*}
    1 &\geq 1 - (1-\zeta_{3,7}^i) \\
    1 + \epsilon_{\min}&\leq  1 + 2(1-\zeta_{3,6}^i).
\end{align*}
On paper, $\zeta_{3,6}^i=1$ is infeasible. In Gurobi on the other hand, $\zeta_{3,6}^i=1$ behaves as feasible, taking on value 0.9999950003800046. And since in this case, $\zeta_{3,6}^i=1$ results in a correct classification for datapoint $i=175$, Gurobi chooses this assignment. However, in reality datapoint $i=175$ should get routed to leaf node 7, i.e., $\zeta_{3,7}^{175}=1$, where it is misclassified. This numerical issue, which is caused by the small value of $\epsilon_{\min}$ creates a situation wherein there is a discrepancy between the optimization problem and the actual training accuracy. Since $\epsilon_{\min}$ is already the largest possible value, this issue cannot be resolved. In our numerical experiments, we obtained an optimal objective value of 12, which corresponds to the number of misclassified datapoints, resulting in an in-sample accuracy of~93\%. However, upon evaluating the output tree, we found that the actual in-sample accuracy is~92\%. In the case of balanced decision trees, out of the 560 MIO instances (28 datasets $\times$  5 samples $\times$ 4 depths) that we solved, we encountered numerical issues with \texttt{OCT} in~54\% of the instances. These issues led to a discrepancy in the in-sample accuracy of up to~92\%, see Tables~\ref{tab:in_sample_non_categorical_data_no_reg_part_1}-\ref{tab:in_sample_non_categorical_data_no_reg_part_3} for detailed results.
\begin{figure}[ht]
\vskip 0.2in
\begin{center}
\centerline{\includegraphics[width=0.5\columnwidth]{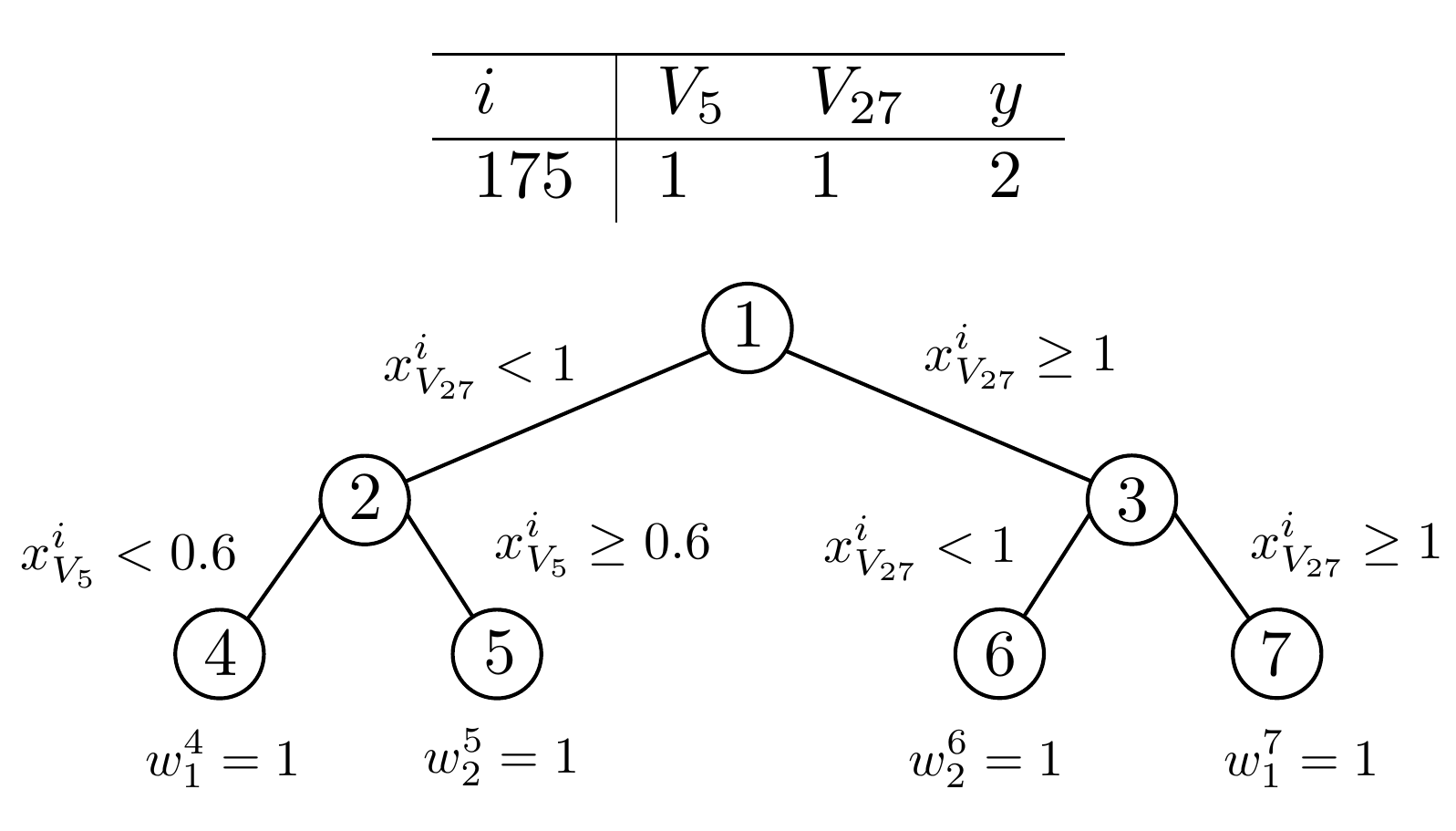}}
\caption{\revision{Example of an instance where \texttt{OCT} exhibits numerical issues. According to the solution of the optimization problem, datapoint 175 should get routed to leaf node 7 where it gets misclassified. However, in practice, we observe that this datapoint is assigned to leaf node 6 and mistakenly reported as being correctly classified. This causes a discrepancy between the optimization problem objective and the actual accuracy of the tree returned by the optimization}.}
\label{fig:OCT_issue}
\end{center}
\vskip -0.2in
\end{figure}
}
%%%%%%%%%%%%%%%%%%%%%%%%%%%%%%%%%%%%%%%%%%%%%%%%%%
%%%%%%%%%%%%%%%%%%%%%%%%%%%%%%%%%%%%%%%%%%%%%%%%%%
\section{Comparison with OCT {(Proof of Theorem~\ref{theo:OCT})}}
\label{appendix_sec:oct_comparison}
%%%%%%%%%%%%%%%%%%%%%%%%%%%%%%%%%%%%%%%%%%%%%%%%%%
%%%%%%%%%%%%%%%%%%%%%%%%%%%%%%%%%%%%%%%%%%%%%%%%%%

In this section, we demonstrate that formulation~\eqref{eq:flow} has a stronger LO relaxation than formulation~\eqref{eq:OCT}. In formulation~\eqref{eq:OCT}, $v_n$ can be fixed to $p_n$ for all nodes (in the case of binary data). Moreover, regularization variables $p_n$ can be fixed to $1$ for balanced trees. Using the identity $\sum_{f \in \sets F:x_f^i=1}b_{mf}=1-\sum_{f \in \sets F:x_f^i=0}b_{mf}$ and noting that $l_n$ can be fixed to $1$ in the formulation, we obtain the simplified OCT formulation 
\begin{subequations}\label{eq:OCT2}
\begin{align}
\text{maximize} \;\;&  \displaystyle \left(|\sets I|-\sum_{n \in \sets T}L_n\right)\label{eq:OCT2_obj} \\
%%%%%%%%%%%%%%%%%%%%%%%%%%%%%%%%%The flow conservation constraints
\text{subject to}\;\;&  \displaystyle L_n \geq Q_n - Q_{nk} - |\sets I|(1-w^n_k) \quad&\hspace{-2cm}  \forall k \in \sets K, n \in \sets T \hfill\label{eq:OCT2_first}\\
& \displaystyle L_n \leq Q_n - Q_{nk} + |\sets I|w^n_k \quad &\hspace{-2cm} \forall k \in \sets K, n \in \sets T \hfill \label{eq:OCT2_ub}\\
&  \displaystyle Q_{nk}= \sum_{\begin{smallmatrix} i\in \sets I:\\ y^i=k \end{smallmatrix}} \zeta_{a(n),n}^i \quad &\hspace{-2cm}\forall k \in \sets K, n \in \sets T \hfill\label{eq:OCT2_P1}\\
&  Q_n=\sum_{i \in \sets I} \zeta_{a(n),n}^i \quad &\hspace{-2cm}\forall n \in \sets T \hfill\label{eq:OCT2_P2}\\
&   \sum_{k \in \sets K}w^n_k=1 \quad &\hspace{-2cm}\forall n \in \sets T \hfill\label{eq:OCT2_label}\\
&  \sum_{n \in \sets T} \zeta_{a(n),n}^i = 1 \quad  &\hspace{-2cm} \forall i \in \sets I \hfill\label{eq:OCT2_assign}\\
&    \sum_{f \in \sets F:x_f^i=1}b_{mf} \geq  \zeta_{a(n),n}^i &\hspace{-2cm}\forall i \in \sets I, n \in \sets T, m \in \sets {PR}(n) \label{eq:OCT2_right} \hfill\\
&  \sum_{f \in \sets F:x_f^i=0}b_{mf} \geq  2\zeta_{a(n),n}^i-1 &\hspace{-2cm}\forall i \in \sets I, n \in \sets T, m \in \sets {PL}(n)  \hfill\label{eq:OCT2_bound}\\
&  \sum_{f \in \sets F}b_{nf} = 1  \quad  &\hspace{-2cm}\forall n \in \sets B \hfill\label{eq:OCT2_branch}\\
&   \zeta_{a(n),n}^i \in \{0,1\}  \quad&\hspace{-2cm} \forall i \in \sets I, n \in \sets T \hfill\\
&   b_{nf}\in \{0,1\} \quad & \hspace{-10cm}\forall f \in \sets F, n \in \sets B.\label{eq:OCT2_last}
\end{align}
\end{subequations}

\subsection{Strengthening}\label{sec:strength}

We now show how formulation \eqref{eq:OCT2} can be strengthened resulting in formulation \eqref{eq:flow}. We note that the validity of the steps below is guaranteed by correctness of formulation~\eqref{eq:flow}. Thus we do not explicitly discuss the validity in the arguments. 

\paragraph{Bound tightening for \eqref{eq:OCT2_bound}. } Adding the quantity $1-\zeta_{a(n),n}^i\geq 0$ to the right{-}hand side of \eqref{eq:OCT2_bound}, we obtain the stronger constraints
\begin{equation}\label{eq:OCT2_left}
    \sum_{f \in \sets F:x_f^i=0}b_{mf} \; \geq \;  \zeta_{a(n),n}^i \quad \forall i \in \sets I,\; n \in \sets T,\; m \in \sets {PL}(n).
\end{equation}

\paragraph{Improved branching constraints.} 
Constraints \eqref{eq:OCT2_right} can be strengthened to 
\begin{equation}\label{eq:OCT2_right_new}
     \sum_{f \in \sets F:x_f^i=1}b_{mf} \; \geq \; \sum_{n\in \sets T: m\in \sets{ PR}(n)}\zeta_{a(n),n}^i \quad \forall i \in \sets I,\;  m \in \sets B.
\end{equation}
Observe that constraints \eqref{eq:OCT2_right_new}, in addition to being stronger than \eqref{eq:OCT2_right}, also reduce the number of constraints required to represent the LO relaxation. Similarly, constraint \eqref{eq:OCT2_left} can be further improved to 
\begin{equation}\label{eq:OCT2_left_new}
     \sum_{f \in \sets F:x_f^i=0}b_{mf} \; \geq \;  \sum_{n\in \sets T: m\in \sets{ PL}(n)}\zeta_{a(n),n}^i \quad \forall i \in \sets I, \; m \in \sets B.
\end{equation}

\paragraph{Improved missclassification formulation. } 
For all $i\in \sets I$ and $n\in \sets T$, define additional variables $z_{a(n),n}^i \in \{0,1\}$ such that $z_{a(n),n}^i\leq \zeta_{a(n),n}w_{n,y^i}$. Note that $z_{a(n),n}^i=1$ implies that datapoint $i$ is routed to terminal node $n$ ($\zeta_{a(n),n}^i=1$) and the class of $i$ is assigned to $n$ ($w_{ny^i}=1$). Hence $z_{a(n),n}^i=1$ only if datapoint~$i$ is correctly classified at terminal node~$n$. Upper bounds of $z_{a(n),n}^i=1$ can be imposed via the linear constraints %
\begin{equation}\label{eq:linearization}
  z_{a(n),n}^i \; \leq \; \zeta_{a(n),n}^i,\; z_{a(n),n}^i \; \leq \; w^n_{y^i} \quad \forall n\in\sets T,\; i\in \sets I. 
\end{equation}
In addition, since $L_n$ corresponds to the number of missclassified points at terminal node $n\in \sets T$ and the total number of missclassified points is $\sum_{n\in \sets T}L_n$, we find that constraints 
\begin{equation}\label{eq:missclass}
    L_n \; \geq \; \sum_{i\in \sets I}(\zeta_{a(n),n}^i -z_{a(n),n}^i)
\end{equation}
are valid.
Note that constraints \eqref{eq:missclass} and \eqref{eq:OCT2_assign} imply that 
\begin{equation}\label{eq:totalMissclass}
    \sum_{n\in \sets T}L_n \; \geq \;  |\sets I| -\sum_{i\in \sets I}\sum_{n\in \sets T}z_{a(n),n}^i.
\end{equation}

\subsection{Simplification}\label{sec:simple}
As discussed in the preceding sections, the linear optimization relaxation of the formulation obtained in Section~\ref{sec:strength}, given by constraints \eqref{eq:OCT2_first}-\eqref{eq:OCT2_assign}, \eqref{eq:OCT2_branch}-\eqref{eq:OCT2_last}, \eqref{eq:OCT2_right_new}, \eqref{eq:OCT2_left_new}, \eqref{eq:linearization} and \eqref{eq:missclass}, is  stronger than the relaxation of OCT, as either constraints were tightened or additional constraints were added. We now show how the resulting formulation can be simplified without loss of relaxation quality to obtain problem~\eqref{eq:flow}.  

\paragraph{Upper bound on missclassification. } Variable $L_n$ has a negative objective coefficient and only appears in constraints \eqref{eq:OCT2_first}, \eqref{eq:OCT2_ub}, and \eqref{eq:missclass}, it will always be set to a lower bound. Therefore, constraint \eqref{eq:OCT2_ub} which imposes an upper bound on $L_n$ is redundant and can be eliminated without affecting the relaxation of the problem.

\paragraph{Lower bound on missclassification. } Substituting variables according to \eqref{eq:OCT2_P1} and \eqref{eq:OCT2_P2}, we find that for a given $k\in \sets K$ and $n\in \sets T$, \eqref{eq:OCT2_first} is equivalent to 
\begin{align}
    & \quad L_n \; \geq \; \sum_{i \in \sets I} \zeta_{a(n),n}^i  - \sum_{\begin{smallmatrix} i\in \sets I:\\ y^i=k \end{smallmatrix}} \zeta_{a(n),n}^i - |\sets I|(1-w^n_k)\nonumber\\
    \Leftrightarrow\;& \quad L_n \; \geq \; \sum_{\substack{i \in \sets I\\y_i=k}}(w^n_k-1)  +\sum_{\substack{i \in \sets I\\y^i\neq k}}(\zeta_{a(n),n}^i-1+w^n_k)\label{eq:Ln_ub_new}.
\end{align}
Observe that $w^n_k-1\leq 0 \leq \zeta_{a(n),n}^i-z_{a(n),n}^i$. Moreover, we also have that for any $i\in \sets I$ and $k\in \sets K\setminus\{y^i\}$, 
\begin{equation}\label{eq:ineqs}
z_{a(n),n}^i \; \leq \;  w^n_{y^i} \; \leq \; 1-w^n_k,
\end{equation} where the first inequality follows from \eqref{eq:linearization} and the second inequality follows from \eqref{eq:OCT2_label}. Therefore, from \eqref{eq:ineqs} we conclude that $\zeta_{a(n),n}^i-1+w^n_k\leq\zeta_{a(n),n}^i-z_{a(n),n}^i$ and inequalities \eqref{eq:missclass} dominate inequalities \eqref{eq:Ln_ub_new} and thus \eqref{eq:OCT2_first}. Since inequalities \eqref{eq:OCT2_P1} and \eqref{eq:OCT2_P2} only appear in inequalities \eqref{eq:OCT2_first} and \eqref{eq:OCT2_ub}, which where shown to be redundant, they can be dropped as well. Finally, as inequalities \eqref{eq:missclass} define the unique lower bounds of $L_n$ in the simplified formulation, inequalities \eqref{eq:totalMissclass} can be converted to an equality and the objective~\eqref{eq:OCT2_obj} can be updated accordingly. After all the changes outlined so far, the formulation \eqref{eq:OCT2} reduces to 
\begin{subequations}\label{eq:OCT2.0}
\begin{align}
\text{maximize} \;\;&  \displaystyle \sum_{i\in \sets I}\sum_{n\in \sets T}z_{n,a(n)}^i\\
%%%%%%%%%%%%%%%%%%%%%%%%%%%%%%%%%The flow conservation constraints
\text{subject to}\;\;& \sum_{k \in \sets K}w^n_k=1 \quad &\hspace{-2cm}\forall n \in \sets T \hfill\label{eq:OCT2.0_label}\\
&  \sum_{n \in \sets T} \zeta_{a(n),n}^i = 1 \quad  &\hspace{-2cm} \forall i \in \sets I \hfill\label{eq:OCT2.0_assign}\\
&\sum_{f \in \sets F:x_f^i=1}b_{mf} \geq  \sum_{n\in \sets T: m\in \sets{ PR}(n)}\zeta_{a(n),n}^i &\hspace{-2cm}\forall i \in \sets I, m \in \sets B\label{eq:OCT2.0_right}\\
&\sum_{f \in \sets F:x_f^i=0}b_{mf} \geq  \sum_{n\in \sets T: m\in \sets{ PL}(n)}\zeta_{a(n),n}^i &\hspace{-2cm}\forall i \in \sets I, m \in \sets B\label{eq:OCT2.0_left}\\
&  \sum_{f \in \sets F}b_{nf} = 1  \quad  &\hspace{-2cm}\forall n \in \sets B \hfill\label{eq:OCT2.0_branch}\\
& z_{a(n),n}^i\leq  \zeta_{a(n),n}^i\quad &\hspace{-2cm}\forall n\in\sets T, i\in \sets I\\
&z_{a(n),n}^i\leq w_{ny^i}\quad &\hspace{-2cm}\forall n\in\sets T, i\in \sets I\\
&   \zeta_{a(n),n}^i \in \{0,1\}  \quad&\hspace{-2cm} \forall i \in \sets I, n \in \sets T \hfill\\
&   b_{nf}\in \{0,1\} \quad & \hspace{-10cm}\forall f \in \sets F, n \in \sets B.\label{eq:OCT2.0_last}
\end{align}
\end{subequations}

\subsection{Projection}
We now project out the $\bm \zeta$ variables, obtaining a more compact formulation with the same LO relaxation. Specifically, consider the formulation 
\begin{subequations}\label{eq:Projection}
\begin{align}
\text{maximize} \;\;& \displaystyle \sum_{i\in \sets I}\sum_{n\in \sets T}z_{n,a(n)}^i\\
%%%%%%%%%%%%%%%%%%%%%%%%%%%%%%%%%The flow conservation constraints
\text{subject to }\;\;& \sum_{k \in \sets K}w^n_k=1 \quad &\hspace{-2cm}\forall n \in \sets T \hfill\\
&\sum_{f \in \sets F:x_f^i=1}b_{mf} \geq  \sum_{n\in \sets T: m\in \sets{ PR}(n)}z_{a(n),n}^i &\hspace{-2cm}\forall i \in \sets I, m \in \sets B\label{eq:Projection_right}\\
&\sum_{f \in \sets F:x_f^i=0}b_{mf} \geq  \sum_{n\in \sets T: m\in \sets{ PL}(n)}z_{a(n),n}^i &\hspace{-2cm}\forall i \in \sets I, m \in \sets B\label{eq:Projection_left}\\
&  \sum_{f \in \sets F}b_{nf} = 1  \quad  &\hspace{-2cm}\forall n \in \sets B \hfill\\
&z_{a(n),n}^i\leq w^n_{y^i}\quad &\hspace{-2cm}\forall n\in\sets T, i\in \sets I\\
&   z_{a(n),n}^i \in \{0,1\}  \quad&\hspace{-2cm} \forall i \in \sets I, n \in \sets T \hfill\\
&   b_{nf}\in \{0,1\} \quad & \hspace{-10cm}\forall f \in \sets F, n \in \sets B..
\end{align}
\end{subequations}

\begin{proposition}
Formulations \eqref{eq:OCT2.0} and \eqref{eq:Projection} are equivalent, i.e., their LO relaxations have the same optimal objective value.%~\hfill~\halmos
\end{proposition}
\proof{Proof.}
Let $\nu_1$ and $\nu_2$ be the optimal objective values of the LO relaxations of \eqref{eq:OCT2.0} and \eqref{eq:Projection}, respectively. Note that \eqref{eq:Projection} is a relaxation of \eqref{eq:OCT2.0}, obtained by dropping constraint \eqref{eq:OCT2.0_assign} and replacing $\bm \zeta$ with a lower bound in constraints \eqref{eq:OCT2.0_right} and \eqref{eq:OCT2.0_left}. Therefore, it follows that $\nu_2\geq \nu_1$. We now show that $\nu_2\leq \nu_1$.

Let $(\bm b^*,\bm w^*,\bm z^*)$ be an optimal solution of \eqref{eq:Projection} and let $i\in \sets I$. We show how to construct a feasible solution of \eqref{eq:OCT2.0} with same objective value, thus implying that $\nu_2\leq \nu_1$. For any given $i\in \sets I$, by summing constraints \eqref{eq:Projection_right} and \eqref{eq:Projection_left} for the root node $m=1$, we find that 
\begin{align}\label{eq:z_ub}1=\sum_{f \in \sets F:x_f^i=1}b_{1f}^* +\sum_{f \in \sets F:x_f^i=0}b_{1f}^*
\geq  \sum_{n\in \sets T}(z_{a(n),n}^{i})^*.\end{align}
Now let $\bm \zeta=\bm z^*$. If the inequality in \eqref{eq:z_ub} is active, then $(\bm b^*,\bm w^*,\bm z^*,\bm \zeta)$ satisfies all constraints in \eqref{eq:OCT2.0} and the proof is complete. Otherwise, it follows that either \eqref{eq:Projection_right} or \eqref{eq:Projection_left} is not active at node $m=1$, and without loss of generality assume \eqref{eq:Projection_right} is  not active. Summing up inequalities \eqref{eq:Projection_right} and \eqref{eq:Projection_left} for node $m=r(1)$, we find that 
\begin{align}\label{eq:z_ub2}1=\sum_{f \in \sets F}b_{r(1)f}^*
>  \sum_{n\in \sets T:r(1)\in \sets{PR}(n)\cup \sets{PL}(n)}(z_{a(n),n}^{i})^*,\end{align}
where the strict inequality holds since the right{-}hand side of \eqref{eq:z_ub2} is no greater than the right{-}hand side of \eqref{eq:z_ub}. By applying this process recursively, we obtain a path from node 1 to a terminal node $h\in \sets T$ such that all inequalities \eqref{eq:Projection_right} and \eqref{eq:Projection_left} associated with nodes in this path are inactive. The value $\zeta_{a(h),h}^i$ can be then increased by the minimum slack in the constraints, and the overall process can be repeated until inequality \eqref{eq:OCT2.0_assign} is tight.
\endproof

\subsection{Substitution}
Finally, to recover formulation~\eqref{eq:flow}, substitute, for all $m\in \sets T$, variables 
\begin{align*}z_{m,r(m)}^i&:=\sum_{n\in \sets T: m\in \sets{ PR}(n)}z_{a(n),n}^i,\text{ and}\\ 
z_{m,\ell(m)}^i&:=\sum_{n\in \sets T: m\in \sets{ PL}(n)}z_{a(n),n}^i.
\end{align*}
Similarly, for all $n\in \sets T$ introduce variables $z_{n,t}:=z_{a(n),n}$.
Constraints \eqref{eq:Projection_right} and \eqref{eq:Projection_left} reduce to $\sum_{f \in \sets F:x_f^i=1}b_{mf}\geq z_{m,r(m)}^i$ and $\sum_{f \in \sets F:x_f^i=0}b_{mf}\geq z_{m,\ell(m)}^i$. Finally, since 
\begin{align*}
    z_{a(m),m}&=\sum_{n\in \sets T:m\in \sets{PR}(n)\cup \sets{PL}(n)}z_{a(n),n}\\
    &=\sum_{n\in \sets T:m\in \sets{PR}(n)}z_{a(n),n}+\sum_{n\in \sets T:m\in \sets{PL}(n)}z_{a(n),n}\\
    &=z_{m,r(m)}+z_{m,\ell(m)},
\end{align*}
we recover the flow conservation constraints. In formulation~\eqref{eq:flow}, we do not use the notion of terminal nodes $\sets T$ and use $\sets L$ instead. However, in formulation~\eqref{eq:flow}, the set of leaf nodes $\sets L$ coincides with the set of terminal nodes $\sets T$. Therefore, we correctly recover formulation~\eqref{eq:flow}. 
%%%%%%%%%%%%%%%%%%%%%%%%%%%%%%%%%%%%%%%%%%%%%%%%%%
%%%%%%%%%%%%%%%%%%%%%%%%%%%%%%%%%%%%%%%%%%%%%%%%%%
\section{Benders' Decomposition for Regularized Problems}
\label{appendix_sec:Benders}
%%%%%%%%%%%%%%%%%%%%%%%%%%%%%%%%%%%%%%%%%%%%%%%%%%
%%%%%%%%%%%%%%%%%%%%%%%%%%%%%%%%%%%%%%%%%%%%%%%%%%
In this section, we describe our proposed Benders' decomposition approach adapted to formulation~\eqref{eq:flow_reg}, which can be written equivalently as:
\begin{subequations}
\begin{align}
\text{maximize}\;\;&   \displaystyle (1-\lambda)\sum_{i \in \sets I}g^i(\bm b,\bm w,\bm p) -\lambda \sum_{n\in \sets B}\sum_{f\in \sets F}b_{nf} \label{eq:bendersOCT_master_1_obj}\\
\text{subject to} \; \; & \displaystyle \sum_{f \in \sets F}b_{nf} + p_n + \sum_{m \in \sets P(n)}p_m = 1   &\hspace{-5cm}  \forall n \in \sets B \label{eq:bendersOCT_master_1_branch_or_predict}\\
&  p_n+\sum_{m \in \sets P(n)}p_{m} =1   &  \forall n \in \sets T \label{eq:bendersOCT_master_1_terminal_leaf}\\ 
&  \displaystyle \sum_{k \in \sets K}w^n_{k} = p_n  &\hspace{-5cm}  \forall n \in \sets B \cup \sets T \label{eq:bendersOCT_master_1_leaf_prediction}\\
%%%%%%%%%%%%%%%%%%%%%%%%%%%%%%%%%binary constraints
&  \displaystyle b_{nf} \in \{0,1\}  &\hspace{-5cm} \forall n \in \sets B,f \in \sets F \\
&  \displaystyle p_{n} \in \{0,1\}  &\hspace{-5cm}   \forall n \in \sets B \cup \sets T \\
&  \displaystyle w^n_{k} \in \{0,1\}  &\hspace{-5cm}   \forall n \in  \sets B \cup \sets T,k \in \sets K,
\end{align}
\label{eq:bendersOCT_master_1}
\end{subequations}
where, for any fixed $i\in \sets I$, $\bm w$ and $\bm b$, $g^i(\bm b,\bm w,\bm p)$ is defined as the optimal objective value of the problem 
\begin{subequations}
\begin{align}
\text{maximize}\;\;&  \displaystyle \sum_{i \in \mathcal I} \sum_{n \in \sets B \cup \sets T } z^i_{n,t} \label{eq:bendersOCT_slave_obj}\\
%%%%%%%%%%%%%%%%%%%%%%%%%%%%%%%%%binary constraints
%%%%%%%%%%%%%%%%%%%%%%%%%%%%%%%%%The flow conservation constraints
\text{subject to} \; \;& \displaystyle z^i_{a(n),n} =  z^i_{n,\ell(n)} + z^i_{n,r(n)} + z^i_{n,t}  &\hspace{-5cm}  \forall n \in \sets B \label{eq:bendersOCT_slave_conservation}\\
&  \displaystyle z^i_{a(n),n} = z^i_{n,t} &\hspace{-5cm}   \forall n \in \sets T \label{eq:bendersOCT_slave_conservation_leaf}\\
%%%%%%%%%%%%%%%%%%%%%%%%%%%%%%%%%capacity constraints
& \displaystyle z^i_{s,1} \leq 1 \label{eq:bendersOCT_slave_source}\\
&  \displaystyle z^i_{n,\ell(n)}\leq \sum_{f \in \sets F: x_{f}^i=0}b_{nf} &\hspace{-5cm} \forall n \in \sets B\label{eq:bendersOCT_slave_branch_left}\\
&  \displaystyle z^i_{n,r(n)}\leq \sum_{f \in \sets F: x_{f}^i=1}b_{nf}  &\hspace{-5cm} \forall n \in \sets B \label{eq:bendersOCT_slave_branch_right}\\
&  \displaystyle z^i_{n,t} \leq  w^n_{y^i} &\hspace{-5cm} \forall  n \in \sets B \cup \sets T \label{eq:bendersOCT_slave_sink}\\
%%%% DVs
&  \displaystyle z^i_{a(n),n}, z^i_{n,t}\in \{0,1\}  &\hspace{-5cm}  \forall n \in \sets B \cup \sets T.
\end{align}
\label{eq:bendersOCT_slave}
\end{subequations}

Problem \eqref{eq:bendersOCT_slave} is a maximum flow problem on the flow graph $\sets G$, see Definition~\ref{def:terminology_imbalanced}, whose arc capacities are determined by $({\bm b},{\bm w})$ and datapoint $i\in \sets I$, as formalized next.
\begin{definition}[Capacitated flow graph of imbalanced trees] 
\label{def:imbalanced_flow_graph}
Given the flow graph $\sets G=(\sets V,\sets A)$, vectors~$({\bm b},{\bm w})$, and datapoint $i \in \sets I$, define arc capacities $c^i({\bm b},{\bm w})$ as follows. Let $c^i_{s,1}({\bm b},{\bm w}):=1$, $c^i_{n,\ell(n)}({\bm b},{\bm w}):= \sum_{f \in \sets F: x_{f}^i=0}b_{nf}$, $c^i_{n,r(n)}({\bm b},{\bm w}):=\sum_{f \in \sets F: x_{f}^i=1}b_{nf}$ for all $n \in \sets B$, and $c^i_{n,t}({\bm b},{\bm w}):= w^n_{y^i}$ for $n\in \sets B \cup \sets T$. Define the \emph{capacitated flow graph} $\sets G^i({\bm b},{\bm w})$ as the flow graph $\sets G$ augmented with capacities $c^i({\bm b},{\bm w})$.
\end{definition}
Similar to the derivation of problem~\eqref{eq:master2}, we can reformulate problem~\eqref{eq:bendersOCT_master_1} as
\begin{subequations}
\begin{align}
\displaystyle \mathop \text{maximize}_{g,b,w} \;\;&   \displaystyle (1-\lambda)\sum_{i \in \sets I}g^i -\lambda \sum_{n\in \sets B}\sum_{f\in \sets F}b_{nf} \label{eq:bendersOCT_master_2_obj}\\
\text{subject to}\;\;& 
g^i \leq  \sum_{(n_1,n_2)\in \sets C(\sets S)}c_{n_1,n_2}^i({\bm b},{\bm w}) \quad\quad\qquad&\hspace{-5cm}  \forall i\in \sets I, \sets S\subseteq \sets V\setminus \{t\}: s\in \sets S \label{eq:bendersOCT_master_2_benders_cut}\\
& \displaystyle \sum_{f \in \sets F}b_{nf} + p_n + \sum_{m \in \sets P(n)}p_m = 1   &\hspace{-5cm}  \forall n \in \sets B \label{eq:bendersOCT_master_2_branch_or_predict}\\
&  p_n+\sum_{m \in \sets P(n)}p_{m} =1   &  \forall n \in \sets T \label{eq:bendersOCT_master_2_terminal_leaf}\\ 
&  \displaystyle \sum_{k \in \sets K}w^n_{k} = p_n  &\hspace{-5cm}  \forall n \in \sets B \cup \sets T \label{eq:bendersOCT_master_2_leaf_prediction}\\
%%%%%%%%%%%%%%%%%%%%%%%%%%%%%%%%%binary constraints
&  \displaystyle b_{nf} \in \{0,1\}  &\hspace{-5cm} \forall n \in \sets B,f \in \sets F \\
&  \displaystyle p_{n} \in \{0,1\}  &\hspace{-5cm}   \forall n \in \sets B \cup \sets T \\
&  \displaystyle w^n_{k} \in \{0,1\}  &\hspace{-5cm}   \forall n \in  \sets B \cup \sets T,k \in \sets K\\
&  \displaystyle g^i \leq 1   & \forall i \in \sets I \label{eq:bendersOCT_master_2_g_upperbound}.
\end{align}
\label{eq:bendersOCT_master_2}
\end{subequations}

\begin{figure}[ht]
\vskip 0.2in
\begin{center}
\centerline{\includegraphics[width=0.5\textwidth]{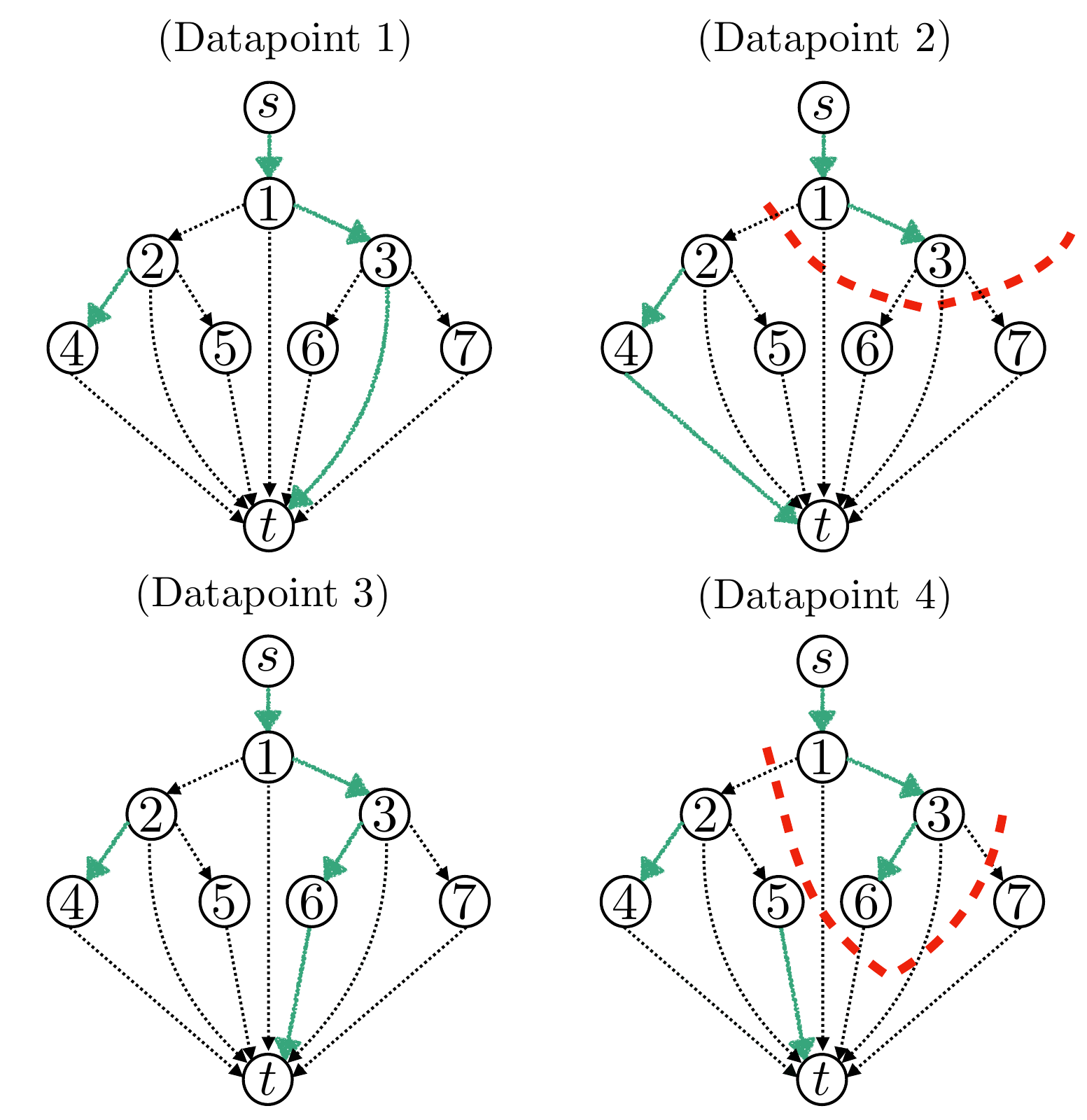}}
\caption{Illustration of Algorithm~\ref{alg:cut_reg} on four datapoints, two of which  are correctly classified (datapoints 1 and 3) and two of which are incorrectly classified (datapoints 2 and 4). Unbroken (green) arcs~$(n,n')$ have capacity~$c_{n,n'}^i({\bm b},{\bm w})=1$ (and others, capacity 0). In the case of datapoints 1 and 3 which are correctly classified, since there exists a path from source to sink, Algorithm~\ref{alg:cut_reg} terminates on line~\ref{line:-1_reg} and returns~$-1$. In the case of datapoints 2 and 4 which are incorrectly classified, Algorithm~\ref{alg:cut_reg} returns set~$\sets S = \{s,1,3\}$ set~$\sets S = \{s,1,3,6\}$ respectively on line~\ref{line:return_reg}. The associated minimum cut for datapoint 2 consists of arcs $(1,2)$, $(1,t)$, $(3,6)$, $(3,t)$ and $(3,7)$ and is represented by the thick (red) dashed line. Similarly the associated minimum cut for datapoint 4 consists of arcs $(1,2)$, $(1,t)$, $(6,t)$, $(3,t)$ and $(3,7)$.}
\label{fig:subproblem_tree_reg}
\end{center}
\vskip -0.2in
\end{figure}

Algorithm~\ref{alg:cut_reg} is a modified version of Algorithm~\ref{alg:cut} tailored to the flow graph introduced in Definition~\ref{def:imbalanced_flow_graph}.
Figure~\ref{fig:subproblem_tree_reg} illustrates Algorithm~\ref{alg:cut_reg}. The difference between Algorithms~\ref{alg:cut_reg} and~\ref{alg:cut} is highlighted with \underline{underline}.

\begin{algorithm}[h]
	\OneAndAHalfSpacedXI
		\caption{Separation procedure for constraints \eqref{eq:bendersOCT_master_2_benders_cut}}
		\label{alg:cut_reg}
		 \textbf{Input:} $(\bm b,\bm w,\bm{\underline{p}},\bm g)\in \{0,1\}^{\sets B\cdot\sets F}\cdot \{0,1\}^{\sets T\cdot \sets K}\cdot \underline{\{0,1\}^{\sets B\cup \sets T}}\cdot \R^{\sets I} \text{  satisfying~\eqref{eq:bendersOCT_master_2_branch_or_predict}-\eqref{eq:bendersOCT_master_2_g_upperbound}; } \newline
		 \hspace*{\algorithmicindent} i~\in~\sets I:  \text{ datapoint used to generate the cut.}$ \newline
		 \textbf{Output:} $-1$ if all constraints \eqref{eq:bendersOCT_master_2_benders_cut} corresponding to $i$ are satisfied;  \newline 
		 \hspace*{\algorithmicindent} source set $\sets S$ of min-cut otherwise.
	    \begin{algorithmic}[1]
 		\State \textbf{if }$g^i=0$ \textbf{ then return } $-1$\label{line:simple_reg}
 		\State \textbf{Initialize} $n\leftarrow 1$ \hfill \Comment{Current node $=$ root}
 		\State \textbf{Initialize} $\sets S\leftarrow \{s\}$ \hfill \Comment{$\sets S$ is in the source set of the cut}
 		\While{\underline{$p_n=0$}}\label{line:ini_reg}
 		\State $\sets S\leftarrow \sets S\cup\{n\}$
 		\If{$c_{n,\ell(n)}^i({\bm b},{\bm w})=1$}  \label{lin:sub-start_reg}
 		\State $n\leftarrow \ell(n)$ \hfill \Comment{Datapoint $i$ is routed left}
 		\ElsIf{$c_{n,r(n)}^i({\bm b},{\bm w})=1$} \label{lin:sub-start2_reg}
 		\State $n\leftarrow r(n)$ \hfill \Comment{Datapoint $i$ is routed right} \label{lin:sub-end_reg}
 		\EndIf
 		\EndWhile \label{line:end_reg}\Comment{\underline{At this point, $n$ is a leaf node of the tree}}
 		\State $\sets S\leftarrow \sets S\cup\{n\}$ \label{line:terminal_start_reg}
 		\If{$g^i > c_{n,t}^i({\bm b},{\bm w})$} \Comment{Minimum cut $\sets S$ with capacity 0 found}\label{line:separation_reg}
 		\State \textbf{return }$\sets S \label{line:return_reg}$
 		\Else \Comment{Minimum cut $\sets S$ has capacity 1, constraints \eqref{eq:bendersOCT_master_2_benders_cut} are satisfied}\label{line:satisfaction_reg}
 		\State \textbf{return} $-1 \label{line:-1_reg}$
 		\EndIf\label{line:terminal_end_reg}
		%\State 
	\end{algorithmic}
\end{algorithm}

\begin{proposition}\label{prop:algorithm_reg}
 Given $i\in \sets I$ and {$(\bm b,\bm w,\bm p,\bm g)$} satisfying~\eqref{eq:bendersOCT_master_2_branch_or_predict}-\eqref{eq:bendersOCT_master_2_g_upperbound}, Algorithm~\ref{alg:cut_reg} either finds a violated inequality \eqref{eq:bendersOCT_master_2_benders_cut} or proves that all such inequalities are satisfied.
\end{proposition}
\proof{Proof.}
Note that the right{-}hand side of \eqref{eq:bendersOCT_master_2_benders_cut}, which corresponds to the capacity of a cut in the graph, is nonnegative. Therefore, if $g^i=0$ (line~\ref{line:simple_reg}), all inequalities are automatically satisfied. Since $({\bm b},{\bm w})$ is integer, all arc capacities in formulation~\eqref{eq:bendersOCT_slave} are either~0 or~1. Moreover, since $g^i\leq 1$, we find that either the value of a minimum cut is~$0$ and there exists a violated inequality, or the value of a minimum cut is at least~$1$ and there is no violated inequality. Finally, there exists a 0-capacity cut if and only if $s$ and~$t$ belong to different connected components in the graph $\sets G^i({\bm b},{\bm w})$.

The connected component $s$ belongs to, can be found using depth-first search.
For any fixed $n \in \sets B$, constraints~\eqref{eq:bendersOCT_slave_conservation}-\eqref{eq:bendersOCT_slave_conservation_leaf} and the definition of $c^i({\bm b},{\bm w})$ imply that only one of the arcs $(n,\ell(n))$, $(n,r(n))$ and $(n,t)$ has capacity 1.
If arc $(n,\ell(n))$ has capacity 1 (line~\ref{lin:sub-start}), then $\ell(n)$ can be added to the component connected to~$s$ (set $\sets S$); the case where arc $(n,r(n))$ has capacity 1 (line~\ref{lin:sub-start2}) is handled analogously. This process continues until a leaf node is reached, i.e., $p_n=1$ (line~\ref{line:terminal_start}).
If the capacity of the arc to the sink is 1 (line~\ref{line:satisfaction_reg}), then an $(s,t)$ is found and no cut with capacity $0$ exists. Otherwise (line~\ref{line:separation_reg}), $\sets S$ is the connected component of $s$ and $t\not \in \sets S$, thus $\sets S$ is the source of a minimum cut with capacity $0$.\halmos
\endproof

%%%%%%%%%%%%%%%%%%%%%%%%%%%%%%%%%%%%%%%%%%%%%%%%%%
%%%%%%%%%%%%%%%%%%%%%%%%%%%%%%%%%%%%%%%%%%%%%%%%%%
\section{Extended Results}
\label{appendix_sec:ext_results}
%%%%%%%%%%%%%%%%%%%%%%%%%%%%%%%%%%%%%%%%%%%%%%%%%%
%%%%%%%%%%%%%%%%%%%%%%%%%%%%%%%%%%%%%%%%%%%%%%%%%%

\revision{\paragraph{Categorical Datasets.} The detail of the experiments presented in Section~\ref{sec:exp_categorical} is provided in Tables~\ref{tab:in_sample_no_reg} and~\ref{tab:in_sample_reg} (for in-sample results),~\ref{tab:out_of_sample} (for out-of-sample results) and~\ref{tab:in_sample_lst_comparison} (for comparison with \texttt{LST}).}

\revision{\paragraph{Mixed-Feature Datasets.} The detail of the experiments presented in Section~\ref{sec:exp_non_categorical} is provided in Tables~\ref{tab:in_sample_non_categorical_data_no_reg_part_1}-~\ref{tab:in_sample_non_categorical_data_with_reg_part_2} (for in-sample results), Tables~\ref{tab:out_of_sample_non_categorical_data_part_1} and~\ref{tab:out_of_sample_non_categorical_data_part_2} (for out-of-sample results)}

% \begin{sidewaystable*}[t]
\begin{landscape}
\begin{table}
\OneAndAHalfSpacedXII
\caption{{
In-sample results including the average and standard deviation of training accuracy, optimality gap, and solving time across 5 samples for the case of $\lambda = 0$ \revision{on categorical datasets}.
The best performance achieved in a given dataset and depth is reported in \textbf{bold}.}
}
\small{
\label{tab:in_sample_no_reg}
\setlength{\tabcolsep}{4pt}
% \resizebox{1.3\textheight}{!}{%
\scalebox{0.79}{
\begin{tabular}{lc||ccc|ccc|ccc|>{\color{black}}c>{\color{black}}c>{\color{black}}c}
\hline
Dataset &
  Depth &
  \multicolumn{3}{c|}{\texttt{OCT}} &
  \multicolumn{3}{c|}{\texttt{BinOCT}} &
  \multicolumn{3}{c|}{\flow} &
  \multicolumn{3}{c}{\benders}\\
 & & Train-acc & Gap & Time & Train-acc & Gap & Time & Train-acc & Gap & Time & Train-acc &
  Gap &Time\\
  \hline
  soybean-small&2&\textbf{1.00$\pm$0.00}&\textbf{0.00$\pm$0.00}&   2$\pm$   1&\textbf{1.00$\pm$0.00}&\textbf{0.00$\pm$0.00}&\textbf{   0$\pm$   0}&\textbf{1.00$\pm$0.00}&\textbf{0.00$\pm$0.00}&\textbf{   0$\pm$   0}&\textbf{1.00$\pm$0.00}&\textbf{0.00$\pm$0.00}&   1$\pm$   0\\
  soybean-small&3&\textbf{1.00$\pm$0.00}&\textbf{0.00$\pm$0.00}&   2$\pm$   0&\textbf{1.00$\pm$0.00}&\textbf{0.00$\pm$0.00}&\textbf{   0$\pm$   0}&\textbf{1.00$\pm$0.00}&\textbf{0.00$\pm$0.00}&   1$\pm$   0&\textbf{1.00$\pm$0.00}&\textbf{0.00$\pm$0.00}&   1$\pm$   0\\
  soybean-small&4&\textbf{1.00$\pm$0.00}&\textbf{0.00$\pm$0.00}&   7$\pm$   1&\textbf{1.00$\pm$0.00}&\textbf{0.00$\pm$0.00}&\textbf{   1$\pm$   0}&\textbf{1.00$\pm$0.00}&\textbf{0.00$\pm$0.00}&\textbf{   1$\pm$   0}&\textbf{1.00$\pm$0.00}&\textbf{0.00$\pm$0.00}&   2$\pm$   0\\
  soybean-small&5&\textbf{1.00$\pm$0.00}&\textbf{0.00$\pm$0.00}&  16$\pm$   2&\textbf{1.00$\pm$0.00}&\textbf{0.00$\pm$0.00}&\textbf{   1$\pm$   0}&\textbf{1.00$\pm$0.00}&\textbf{0.00$\pm$0.00}&   2$\pm$   0&\textbf{1.00$\pm$0.00}&\textbf{0.00$\pm$0.00}&   3$\pm$   2\\
  monk3&2&\textbf{0.94$\pm$0.01}&\textbf{0.00$\pm$0.00}&   2$\pm$   0&\textbf{0.94$\pm$0.01}&\textbf{0.00$\pm$0.00}&\textbf{   0$\pm$   0}&\textbf{0.94$\pm$0.01}&\textbf{0.00$\pm$0.00}&   1$\pm$   0&\textbf{0.94$\pm$0.01}&\textbf{0.00$\pm$0.00}&   1$\pm$   0\\
  monk3&3&\textbf{0.98$\pm$0.01}&\textbf{0.00$\pm$0.00}& 409$\pm$ 359&\textbf{0.98$\pm$0.01}&\textbf{0.00$\pm$0.00}&  32$\pm$  25&\textbf{0.98$\pm$0.01}&\textbf{0.00$\pm$0.00}&  39$\pm$  18&\textbf{0.98$\pm$0.01}&\textbf{0.00$\pm$0.00}&\textbf{  15$\pm$  12}\\
  monk3&4&\textbf{1.00$\pm$0.00}&\textbf{0.00$\pm$0.00}& 177$\pm$ 123&\textbf{1.00$\pm$0.00}&\textbf{0.00$\pm$0.00}& 652$\pm$ 876&\textbf{1.00$\pm$0.00}&\textbf{0.00$\pm$0.00}&\textbf{  18$\pm$  14}&\textbf{1.00$\pm$0.00}&\textbf{0.00$\pm$0.00}&  58$\pm$ 121\\
  monk3&5&\textbf{1.00$\pm$0.00}&\textbf{0.00$\pm$0.00}& 108$\pm$  71&\textbf{1.00$\pm$0.00}&\textbf{0.00$\pm$0.00}& 156$\pm$ 157&\textbf{1.00$\pm$0.00}&\textbf{0.00$\pm$0.00}&  15$\pm$  13&\textbf{1.00$\pm$0.00}&\textbf{0.00$\pm$0.00}&\textbf{  11$\pm$   2}\\
  monk1&2&\textbf{0.86$\pm$0.03}&\textbf{0.00$\pm$0.00}&   3$\pm$   1&\textbf{0.86$\pm$0.03}&\textbf{0.00$\pm$0.00}&   1$\pm$   1&\textbf{0.86$\pm$0.03}&\textbf{0.00$\pm$0.00}&\textbf{   1$\pm$   0}&\textbf{0.86$\pm$0.03}&\textbf{0.00$\pm$0.00}&\textbf{   1$\pm$   0}\\
  monk1&3&\textbf{0.95$\pm$0.02}&\textbf{0.00$\pm$0.00}& 687$\pm$ 783&\textbf{0.95$\pm$0.02}&\textbf{0.00$\pm$0.00}&  45$\pm$  29&\textbf{0.95$\pm$0.02}&\textbf{0.00$\pm$0.00}&  29$\pm$   7&\textbf{0.95$\pm$0.02}&\textbf{0.00$\pm$0.00}&\textbf{   5$\pm$   2}\\
  monk1&4&\textbf{1.00$\pm$0.00}&\textbf{0.00$\pm$0.00}&  89$\pm$  91&\textbf{1.00$\pm$0.00}&\textbf{0.00$\pm$0.00}&  62$\pm$ 127&\textbf{1.00$\pm$0.00}&\textbf{0.00$\pm$0.00}&   8$\pm$  10&\textbf{1.00$\pm$0.00}&\textbf{0.00$\pm$0.00}&\textbf{   4$\pm$   2}\\
  monk1&5&\textbf{1.00$\pm$0.00}&\textbf{0.00$\pm$0.00}&  91$\pm$  44&\textbf{1.00$\pm$0.00}&\textbf{0.00$\pm$0.00}&\textbf{   5$\pm$   2}&\textbf{1.00$\pm$0.00}&\textbf{0.00$\pm$0.00}&  13$\pm$   5&\textbf{1.00$\pm$0.00}&\textbf{0.00$\pm$0.00}&   7$\pm$   2\\
  hayes-roth&2&\textbf{0.66$\pm$0.02}&\textbf{0.00$\pm$0.00}&  11$\pm$   2&\textbf{0.66$\pm$0.02}&\textbf{0.00$\pm$0.00}&   5$\pm$   1&\textbf{0.66$\pm$0.02}&\textbf{0.00$\pm$0.00}&\textbf{   1$\pm$   0}&\textbf{0.66$\pm$0.02}&\textbf{0.00$\pm$0.00}&\textbf{   1$\pm$   0}\\
  hayes-roth&3&\textbf{0.81$\pm$0.04}&0.13$\pm$0.29&1426$\pm$1326&\textbf{0.81$\pm$0.04}&0.75$\pm$0.42&3496$\pm$ 233&\textbf{0.81$\pm$0.04}&\textbf{0.00$\pm$0.00}&  28$\pm$   7&\textbf{0.81$\pm$0.04}&\textbf{0.00$\pm$0.00}&\textbf{  13$\pm$   5}\\
  hayes-roth&4&0.87$\pm$0.03&1.00$\pm$0.00&3600$\pm$   0&0.85$\pm$0.03&1.00$\pm$0.00&3600$\pm$   0&\textbf{0.89$\pm$0.02}&\textbf{0.07$\pm$0.10}&\textbf{3104$\pm$ 751}&\textbf{0.89$\pm$0.02}&0.17$\pm$0.11&3434$\pm$ 372\\
  hayes-roth&5&0.89$\pm$0.02&\textbf{1.00$\pm$0.00}&\textbf{3600$\pm$   0}&0.90$\pm$0.03&\textbf{1.00$\pm$0.00}&\textbf{3600$\pm$   0}&\textbf{0.92$\pm$0.02}&\textbf{1.00$\pm$0.00}&\textbf{3600$\pm$   0}&\textbf{0.92$\pm$0.02}&\textbf{1.00$\pm$0.00}&\textbf{3600$\pm$   0}\\
  monk2&2&\textbf{0.71$\pm$0.02}&\textbf{0.00$\pm$0.00}&  28$\pm$  10&\textbf{0.71$\pm$0.02}&\textbf{0.00$\pm$0.00}&   6$\pm$   0&\textbf{0.71$\pm$0.02}&\textbf{0.00$\pm$0.00}&\textbf{   5$\pm$   2}&\textbf{0.71$\pm$0.02}&\textbf{0.00$\pm$0.00}&  11$\pm$  18\\
  monk2&3&0.80$\pm$0.02&0.82$\pm$0.11&3600$\pm$   0&0.80$\pm$0.03&1.00$\pm$0.01&3600$\pm$   0&\textbf{0.81$\pm$0.02}&0.01$\pm$0.02&1533$\pm$1288&\textbf{0.81$\pm$0.02}&\textbf{0.00$\pm$0.00}&\textbf{ 591$\pm$ 456}\\
  monk2&4&0.87$\pm$0.03&1.00$\pm$0.00&\textbf{3600$\pm$   0}&0.88$\pm$0.03&1.00$\pm$0.00&\textbf{3600$\pm$   0}&\textbf{0.90$\pm$0.03}&0.75$\pm$0.24&\textbf{3600$\pm$   0}&0.90$\pm$0.02&\textbf{0.66$\pm$0.19}&\textbf{3600$\pm$   0}\\
  monk2&5&0.91$\pm$0.02&\textbf{1.00$\pm$0.00}&\textbf{3600$\pm$   0}&0.93$\pm$0.01&\textbf{1.00$\pm$0.00}&\textbf{3600$\pm$   0}&\textbf{0.95$\pm$0.03}&\textbf{1.00$\pm$0.00}&\textbf{3600$\pm$   0}&0.94$\pm$0.01&\textbf{1.00$\pm$0.00}&\textbf{3600$\pm$   0}\\
  house-votes-84&2&\textbf{0.97$\pm$0.00}&\textbf{0.00$\pm$0.00}&   6$\pm$   2&\textbf{0.97$\pm$0.00}&\textbf{0.00$\pm$0.00}&\textbf{   1$\pm$   0}&\textbf{0.97$\pm$0.00}&\textbf{0.00$\pm$0.00}&   2$\pm$   0&\textbf{0.97$\pm$0.00}&\textbf{0.00$\pm$0.00}&   2$\pm$   0\\
  house-votes-84&3&\textbf{0.99$\pm$0.00}&\textbf{0.00$\pm$0.00}& 341$\pm$ 229&\textbf{0.99$\pm$0.00}&\textbf{0.00$\pm$0.00}& 181$\pm$ 182&\textbf{0.99$\pm$0.00}&\textbf{0.00$\pm$0.00}& 184$\pm$ 120&\textbf{0.99$\pm$0.00}&\textbf{0.00$\pm$0.00}&\textbf{  29$\pm$  18}\\
  house-votes-84&4&\textbf{1.00$\pm$0.00}&\textbf{0.00$\pm$0.00}&  52$\pm$  26&\textbf{1.00$\pm$0.00}&0.20$\pm$0.45&1107$\pm$1585&\textbf{1.00$\pm$0.00}&\textbf{0.00$\pm$0.00}& 204$\pm$ 301&\textbf{1.00$\pm$0.00}&\textbf{0.00$\pm$0.00}&\textbf{  17$\pm$  15}\\
  house-votes-84&5&\textbf{1.00$\pm$0.00}&\textbf{0.00$\pm$0.00}& 156$\pm$ 168&\textbf{1.00$\pm$0.00}&\textbf{0.00$\pm$0.00}& 454$\pm$ 603&\textbf{1.00$\pm$0.00}&\textbf{0.00$\pm$0.00}&  34$\pm$  34&\textbf{1.00$\pm$0.00}&\textbf{0.00$\pm$0.00}&\textbf{  29$\pm$  31}\\
  spect&2&\textbf{0.81$\pm$0.01}&\textbf{0.00$\pm$0.00}&  35$\pm$  17&\textbf{0.81$\pm$0.01}&\textbf{0.00$\pm$0.00}&   8$\pm$   4&\textbf{0.81$\pm$0.01}&\textbf{0.00$\pm$0.00}&   8$\pm$   3&\textbf{0.81$\pm$0.01}&\textbf{0.00$\pm$0.00}&\textbf{   6$\pm$   1}\\
  spect&3&\textbf{0.86$\pm$0.02}&\textbf{0.00$\pm$0.00}&1495$\pm$1045&\textbf{0.86$\pm$0.02}&0.51$\pm$0.47&2985$\pm$ 874&\textbf{0.86$\pm$0.02}&0.02$\pm$0.05&1027$\pm$1453&\textbf{0.86$\pm$0.02}&\textbf{0.00$\pm$0.00}&\textbf{ 335$\pm$ 613}\\
  spect&4&0.88$\pm$0.07&0.78$\pm$0.10&\textbf{3600$\pm$   0}&0.90$\pm$0.01&0.96$\pm$0.04&\textbf{3600$\pm$   0}&0.91$\pm$0.01&0.28$\pm$0.10&\textbf{3600$\pm$   0}&\textbf{0.92$\pm$0.01}&\textbf{0.15$\pm$0.10}&\textbf{3600$\pm$   0}\\
  spect&5&0.91$\pm$0.02&0.84$\pm$0.17&3600$\pm$   0&0.93$\pm$0.01&0.88$\pm$0.19&3600$\pm$   0&\textbf{0.95$\pm$0.01}&0.40$\pm$0.21&3600$\pm$   0&\textbf{0.95$\pm$0.01}&\textbf{0.30$\pm$0.22}&\textbf{3238$\pm$ 810}\\
  breast-cancer&2&\textbf{0.80$\pm$0.02}&\textbf{0.00$\pm$0.00}& 108$\pm$  62&\textbf{0.80$\pm$0.02}&\textbf{0.00$\pm$0.00}&\textbf{  12$\pm$   2}&\textbf{0.80$\pm$0.02}&\textbf{0.00$\pm$0.00}&  21$\pm$   3&\textbf{0.80$\pm$0.02}&\textbf{0.00$\pm$0.00}&  21$\pm$  27\\
  breast-cancer&3&\textbf{0.85$\pm$0.02}&1.00$\pm$0.00&3600$\pm$   0&0.84$\pm$0.02&1.00$\pm$0.00&3600$\pm$   0&\textbf{0.85$\pm$0.02}&0.72$\pm$0.10&3600$\pm$   0&\textbf{0.85$\pm$0.02}&\textbf{0.38$\pm$0.23}&\textbf{3376$\pm$ 501}\\
  breast-cancer&4&0.87$\pm$0.02&\textbf{1.00$\pm$0.00}&\textbf{3600$\pm$   0}&0.87$\pm$0.02&\textbf{1.00$\pm$0.00}&\textbf{3600$\pm$   0}&\textbf{0.89$\pm$0.02}&\textbf{1.00$\pm$0.00}&\textbf{3600$\pm$   0}&\textbf{0.89$\pm$0.02}&\textbf{1.00$\pm$0.00}&\textbf{3600$\pm$   0}\\
  breast-cancer&5&0.89$\pm$0.04&\textbf{1.00$\pm$0.00}&\textbf{3600$\pm$   0}&0.91$\pm$0.02&\textbf{1.00$\pm$0.00}&\textbf{3600$\pm$   0}&0.90$\pm$0.03&\textbf{1.00$\pm$0.00}&\textbf{3600$\pm$   0}&\textbf{0.92$\pm$0.03}&\textbf{1.00$\pm$0.00}&\textbf{3600$\pm$   0}\\
  balance-scale&2&\textbf{0.70$\pm$0.01}&\textbf{0.00$\pm$0.00}& 203$\pm$ 101&\textbf{0.70$\pm$0.01}&\textbf{0.00$\pm$0.00}&\textbf{   7$\pm$   1}&\textbf{0.70$\pm$0.01}&\textbf{0.00$\pm$0.00}&  11$\pm$   2&\textbf{0.70$\pm$0.01}&\textbf{0.00$\pm$0.00}&   8$\pm$   3\\
  balance-scale&3&0.75$\pm$0.02&1.00$\pm$0.00&3600$\pm$   0&0.76$\pm$0.01&0.95$\pm$0.05&3600$\pm$   0&\textbf{0.77$\pm$0.01}&\textbf{0.00$\pm$0.00}&1229$\pm$ 204&\textbf{0.77$\pm$0.01}&\textbf{0.00$\pm$0.00}&\textbf{ 439$\pm$ 230}\\
  balance-scale&4&0.74$\pm$0.03&1.00$\pm$0.00&\textbf{3600$\pm$   0}&0.79$\pm$0.02&1.00$\pm$0.00&\textbf{3600$\pm$   0}&0.79$\pm$0.01&1.00$\pm$0.01&\textbf{3600$\pm$   0}&\textbf{0.80$\pm$0.01}&\textbf{0.41$\pm$0.08}&\textbf{3600$\pm$   0}\\
  balance-scale&5&0.76$\pm$0.05&\textbf{1.00$\pm$0.00}&\textbf{3600$\pm$   0}&0.82$\pm$0.01&\textbf{1.00$\pm$0.00}&\textbf{3600$\pm$   0}&0.80$\pm$0.01&\textbf{1.00$\pm$0.00}&\textbf{3600$\pm$   0}&\textbf{0.83$\pm$0.01}&\textbf{1.00$\pm$0.00}&\textbf{3600$\pm$   0}\\
  tic-tac-toe&2&\textbf{0.73$\pm$0.01}&0.10$\pm$0.17&2340$\pm$1154&\textbf{0.73$\pm$0.01}&\textbf{0.00$\pm$0.00}&\textbf{ 114$\pm$  49}&\textbf{0.73$\pm$0.01}&\textbf{0.00$\pm$0.00}& 312$\pm$  24&\textbf{0.73$\pm$0.01}&\textbf{0.00$\pm$0.00}& 255$\pm$ 207\\
  tic-tac-toe&3&0.75$\pm$0.02&1.00$\pm$0.00&\textbf{3600$\pm$   0}&0.78$\pm$0.01&1.00$\pm$0.00&\textbf{3600$\pm$   0}&\textbf{0.79$\pm$0.01}&1.00$\pm$0.00&\textbf{3600$\pm$   0}&\textbf{0.79$\pm$0.01}&\textbf{0.72$\pm$0.13}&\textbf{3600$\pm$   0}\\
  tic-tac-toe&4&0.76$\pm$0.05&\textbf{1.00$\pm$0.00}&\textbf{3600$\pm$   0}&0.84$\pm$0.01&\textbf{1.00$\pm$0.00}&\textbf{3600$\pm$   0}&0.83$\pm$0.01&\textbf{1.00$\pm$0.00}&\textbf{3600$\pm$   0}&\textbf{0.85$\pm$0.02}&\textbf{1.00$\pm$0.00}&\textbf{3600$\pm$   0}\\
  tic-tac-toe&5&0.70$\pm$0.01&\textbf{1.00$\pm$0.00}&\textbf{3600$\pm$   0}&\textbf{0.89$\pm$0.02}&\textbf{1.00$\pm$0.00}&\textbf{3600$\pm$   0}&0.82$\pm$0.01&\textbf{1.00$\pm$0.00}&\textbf{3600$\pm$   0}&0.88$\pm$0.02&\textbf{1.00$\pm$0.00}&\textbf{3600$\pm$   0}\\
  car-evaluation&2&\textbf{0.78$\pm$0.01}&\textbf{0.00$\pm$0.00}&1168$\pm$ 317&\textbf{0.78$\pm$0.01}&\textbf{0.00$\pm$0.00}&\textbf{  28$\pm$   6}&\textbf{0.78$\pm$0.01}&\textbf{0.00$\pm$0.00}&  59$\pm$  27&\textbf{0.78$\pm$0.01}&\textbf{0.00$\pm$0.00}&  63$\pm$  22\\
  car-evaluation&3&0.77$\pm$0.04&1.00$\pm$0.00&\textbf{3600$\pm$   0}&\textbf{0.82$\pm$0.01}&1.00$\pm$0.00&\textbf{3600$\pm$   0}&\textbf{0.82$\pm$0.01}&0.87$\pm$0.06&\textbf{3600$\pm$   0}&\textbf{0.82$\pm$0.01}&\textbf{0.23$\pm$0.10}&\textbf{3600$\pm$   0}\\
  car-evaluation&4&0.78$\pm$0.02&\textbf{1.00$\pm$0.00}&\textbf{3600$\pm$   0}&0.83$\pm$0.01&\textbf{1.00$\pm$0.00}&\textbf{3600$\pm$   0}&0.81$\pm$0.02&\textbf{1.00$\pm$0.00}&\textbf{3600$\pm$   0}&\textbf{0.84$\pm$0.01}&\textbf{1.00$\pm$0.00}&\textbf{3600$\pm$   0}\\
  car-evaluation&5&0.73$\pm$0.03&\textbf{1.00$\pm$0.00}&\textbf{3600$\pm$   0}&0.86$\pm$0.01&\textbf{1.00$\pm$0.00}&\textbf{3600$\pm$   0}&0.81$\pm$0.03&\textbf{1.00$\pm$0.00}&\textbf{3600$\pm$   0}&\textbf{0.86$\pm$0.02}&\textbf{1.00$\pm$0.00}&\textbf{3600$\pm$   0}\\
  kr-vs-kp&2&0.83$\pm$0.05&0.98$\pm$0.02&3600$\pm$   0&\textbf{0.87$\pm$0.01}&\textbf{0.00$\pm$0.00}&\textbf{ 102$\pm$  41}&\textbf{0.87$\pm$0.01}&\textbf{0.00$\pm$0.00}&1399$\pm$ 430&\textbf{0.87$\pm$0.01}&\textbf{0.00$\pm$0.00}& 710$\pm$ 369\\
  kr-vs-kp&3&0.72$\pm$0.04&1.00$\pm$0.00&\textbf{3600$\pm$   0}&0.92$\pm$0.01&1.00$\pm$0.00&\textbf{3600$\pm$   0}&0.77$\pm$0.07&1.00$\pm$0.00&\textbf{3600$\pm$   0}&\textbf{0.92$\pm$0.03}&\textbf{0.93$\pm$0.15}&\textbf{3600$\pm$   0}\\
  kr-vs-kp&4&0.70$\pm$0.06&\textbf{1.00$\pm$0.00}&\textbf{3600$\pm$   0}&\textbf{0.94$\pm$0.02}&\textbf{1.00$\pm$0.00}&\textbf{3600$\pm$   0}&0.76$\pm$0.07&\textbf{1.00$\pm$0.00}&\textbf{3600$\pm$   0}&0.91$\pm$0.05&\textbf{1.00$\pm$0.00}&\textbf{3600$\pm$   0}\\
  kr-vs-kp&5&0.65$\pm$0.06&\textbf{1.00$\pm$0.00}&\textbf{3600$\pm$   0}&\textbf{0.94$\pm$0.01}&\textbf{1.00$\pm$0.00}&\textbf{3600$\pm$   0}&0.74$\pm$0.06&\textbf{1.00$\pm$0.00}&\textbf{3600$\pm$   0}&0.93$\pm$0.03&\textbf{1.00$\pm$0.00}&\textbf{3600$\pm$   0}\\ \hline

\end{tabular}%
}
}
% \end{sidewaystable*}
\end{table}
\end{landscape}

% \begin{sidewaystable*}[t]
% \begin{landscape}
\begin{table}[ht]
\center{
\OneAndAHalfSpacedXII
\caption{  
{In-sample results including the average and standard deviation of optimality gap and solving time across 45 instances (5 samples $\times$ 9 value of lambdas) for the case of $\lambda > 0$ \revision{on categorical datasets}.
The best performance achieved in a given dataset and depth is reported in \textbf{bold}.}}
\small{
\label{tab:in_sample_reg}
\setlength{\tabcolsep}{4pt}
% \resizebox{0.7\textheight}{!}{%
\scalebox{0.79}{
\begin{tabular}{lc||cc|cc|>{\color{black}}c>{\color{black}}c}
\hline
Dataset &
  Depth &
  \multicolumn{2}{c|}{\texttt{OCT}} &
  \multicolumn{2}{c|}{\flow} &
  \multicolumn{2}{c}{\benders}\\
 &  & Gap & Time  & Gap & Time & Gap & Time \\
  \hline
  soybean-small&2&\textbf{0.00$\pm$0.00}&   3$\pm$   1&\textbf{0.00$\pm$0.00}&\textbf{   0$\pm$   0}&\textbf{0.00$\pm$0.00}&   1$\pm$   0\\
  soybean-small&3&\textbf{0.00$\pm$0.00}&   7$\pm$   6&\textbf{0.00$\pm$0.00}&\textbf{   1$\pm$   0}&\textbf{0.00$\pm$0.00}&   3$\pm$   1\\
  soybean-small&4&\textbf{0.00$\pm$0.00}&  22$\pm$  20&\textbf{0.00$\pm$0.00}&\textbf{   2$\pm$   1}&\textbf{0.00$\pm$0.00}&   6$\pm$   4\\
  soybean-small&5&\textbf{0.00$\pm$0.00}&  33$\pm$  22&\textbf{0.00$\pm$0.00}&\textbf{   6$\pm$   3}&\textbf{0.00$\pm$0.00}&  14$\pm$  10\\
  monk3&2&\textbf{0.00$\pm$0.00}&   3$\pm$   1&\textbf{0.00$\pm$0.00}&\textbf{   1$\pm$   0}&\textbf{0.00$\pm$0.00}&   2$\pm$   0\\
  monk3&3&\textbf{0.00$\pm$0.00}& 395$\pm$ 760&\textbf{0.00$\pm$0.00}&\textbf{  10$\pm$  12}&\textbf{0.00$\pm$0.00}&  11$\pm$   8\\
  monk3&4&0.07$\pm$0.14&1161$\pm$1456&\textbf{0.00$\pm$0.00}& 104$\pm$ 186&\textbf{0.00$\pm$0.00}&\textbf{  41$\pm$  53}\\
  monk3&5&0.14$\pm$0.22&1633$\pm$1534&\textbf{0.00$\pm$0.00}& 153$\pm$ 261&\textbf{0.00$\pm$0.00}&\textbf{  46$\pm$  44}\\
  monk1&2&\textbf{0.00$\pm$0.00}&   4$\pm$   2&\textbf{0.00$\pm$0.00}&\textbf{   1$\pm$   0}&\textbf{0.00$\pm$0.00}&   2$\pm$   0\\
  monk1&3&0.01$\pm$0.04&1048$\pm$1015&\textbf{0.00$\pm$0.00}&  13$\pm$   7&\textbf{0.00$\pm$0.00}&\textbf{   9$\pm$   4}\\
  monk1&4&0.01$\pm$0.04&1097$\pm$ 819&\textbf{0.00$\pm$0.00}&  22$\pm$  11&\textbf{0.00$\pm$0.00}&\textbf{  18$\pm$  28}\\
  monk1&5&0.11$\pm$0.15&2600$\pm$1303&\textbf{0.00$\pm$0.00}&  32$\pm$  16&\textbf{0.00$\pm$0.00}&\textbf{  24$\pm$  12}\\
  hayes-roth&2&\textbf{0.00$\pm$0.00}&   9$\pm$   3&\textbf{0.00$\pm$0.00}&\textbf{   1$\pm$   0}&\textbf{0.00$\pm$0.00}&   2$\pm$   1\\
  hayes-roth&3&0.17$\pm$0.28&2326$\pm$1257&\textbf{0.00$\pm$0.00}&  26$\pm$  14&\textbf{0.00$\pm$0.00}&\textbf{  19$\pm$   8}\\
  hayes-roth&4&0.67$\pm$0.30&3298$\pm$ 886&0.05$\pm$0.09&1850$\pm$1470&\textbf{0.00$\pm$0.00}&\textbf{ 954$\pm$ 799}\\
  hayes-roth&5&0.73$\pm$0.29&3263$\pm$ 966&0.39$\pm$0.32&2812$\pm$1383&\textbf{0.29$\pm$0.27}&\textbf{2612$\pm$1485}\\
  monk2&2&\textbf{0.00$\pm$0.00}&  15$\pm$  15&\textbf{0.00$\pm$0.00}&\textbf{   4$\pm$   2}&\textbf{0.00$\pm$0.00}&   5$\pm$   2\\
  monk2&3&0.55$\pm$0.33&3056$\pm$1228&0.00$\pm$0.02&1141$\pm$1127&\textbf{0.00$\pm$0.00}&\textbf{ 566$\pm$ 550}\\
  monk2&4&0.71$\pm$0.30&3205$\pm$1131&0.41$\pm$0.27&2892$\pm$1356&\textbf{0.36$\pm$0.25}&\textbf{2828$\pm$1463}\\
  monk2&5&0.76$\pm$0.30&3205$\pm$1129&0.55$\pm$0.33&2973$\pm$1292&\textbf{0.51$\pm$0.32}&\textbf{2850$\pm$1427}\\
  house-votes-84&2&\textbf{0.00$\pm$0.00}&   4$\pm$   2&\textbf{0.00$\pm$0.00}&\textbf{   2$\pm$   1}&\textbf{0.00$\pm$0.00}&\textbf{   2$\pm$   1}\\
  house-votes-84&3&\textbf{0.00$\pm$0.00}& 348$\pm$ 482&\textbf{0.00$\pm$0.00}&  42$\pm$  66&\textbf{0.00$\pm$0.00}&\textbf{  19$\pm$  26}\\
  house-votes-84&4&0.02$\pm$0.08& 808$\pm$1109&\textbf{0.00$\pm$0.00}& 118$\pm$ 309&\textbf{0.00$\pm$0.00}&\textbf{  37$\pm$  85}\\
  house-votes-84&5&0.10$\pm$0.21&1187$\pm$1418&\textbf{0.00$\pm$0.00}& 105$\pm$ 242&\textbf{0.00$\pm$0.00}&\textbf{  42$\pm$  58}\\
  spect&2&\textbf{0.00$\pm$0.00}&  20$\pm$  16&\textbf{0.00$\pm$0.00}&   5$\pm$   3&\textbf{0.00$\pm$0.00}&\textbf{   5$\pm$   2}\\
  spect&3&0.02$\pm$0.07&1231$\pm$1065&0.01$\pm$0.02& 409$\pm$ 952&\textbf{0.00$\pm$0.00}&\textbf{ 209$\pm$ 597}\\
  spect&4&0.55$\pm$0.29&3194$\pm$1136&0.07$\pm$0.09&1814$\pm$1661&\textbf{0.02$\pm$0.06}&\textbf{1040$\pm$1374}\\
  spect&5&0.70$\pm$0.29&3206$\pm$1126&0.16$\pm$0.16&2528$\pm$1554&\textbf{0.09$\pm$0.13}&\textbf{2035$\pm$1653}\\
  breast-cancer&2&\textbf{0.00$\pm$0.00}& 124$\pm$  92&\textbf{0.00$\pm$0.00}&  20$\pm$   7&\textbf{0.00$\pm$0.00}&\textbf{  11$\pm$   3}\\
  breast-cancer&3&0.75$\pm$0.27&3403$\pm$ 663&0.44$\pm$0.25&3082$\pm$1185&\textbf{0.35$\pm$0.21}&\textbf{2949$\pm$1274}\\
  breast-cancer&4&0.81$\pm$0.25&3584$\pm$ 106&0.67$\pm$0.31&3212$\pm$1110&\textbf{0.64$\pm$0.31}&\textbf{3206$\pm$1128}\\
  breast-cancer&5&0.84$\pm$0.23&3600$\pm$   0&0.70$\pm$0.30&3215$\pm$1100&\textbf{0.67$\pm$0.30}&\textbf{3209$\pm$1120}\\
  balance-scale&2&\textbf{0.00$\pm$0.00}& 152$\pm$  67&\textbf{0.00$\pm$0.00}&  10$\pm$   1&\textbf{0.00$\pm$0.00}&\textbf{   7$\pm$   4}\\
  balance-scale&3&0.93$\pm$0.08&3600$\pm$   0&\textbf{0.00$\pm$0.00}& 981$\pm$ 372&\textbf{0.00$\pm$0.00}&\textbf{ 418$\pm$ 235}\\
  balance-scale&4&0.95$\pm$0.06&3600$\pm$   0&0.81$\pm$0.17&3600$\pm$   0&\textbf{0.48$\pm$0.21}&\textbf{3342$\pm$ 741}\\
  balance-scale&5&0.97$\pm$0.04&\textbf{3600$\pm$   0}&0.88$\pm$0.13&\textbf{3600$\pm$   0}&\textbf{0.83$\pm$0.19}&\textbf{3600$\pm$   0}\\
  tic-tac-toe&2&0.03$\pm$0.12&2300$\pm$ 859&\textbf{0.00$\pm$0.00}& 352$\pm$ 104&\textbf{0.00$\pm$0.00}&\textbf{ 159$\pm$  25}\\
  tic-tac-toe&3&0.96$\pm$0.05&\textbf{3600$\pm$   0}&0.91$\pm$0.11&\textbf{3600$\pm$   0}&\textbf{0.79$\pm$0.14}&\textbf{3600$\pm$   0}\\
  tic-tac-toe&4&0.97$\pm$0.03&\textbf{3600$\pm$   0}&0.93$\pm$0.08&\textbf{3600$\pm$   0}&\textbf{0.91$\pm$0.10}&\textbf{3600$\pm$   0}\\
  tic-tac-toe&5&0.98$\pm$0.02&\textbf{3600$\pm$   0}&0.93$\pm$0.07&\textbf{3600$\pm$   0}&\textbf{0.91$\pm$0.09}&\textbf{3600$\pm$   0}\\
  car-evaluation&2&0.02$\pm$0.11&1451$\pm$ 730&\textbf{0.00$\pm$0.00}&  67$\pm$  26&\textbf{0.00$\pm$0.00}&\textbf{  67$\pm$  17}\\
  car-evaluation&3&0.99$\pm$0.02&3600$\pm$   0&0.73$\pm$0.10&3600$\pm$   0&\textbf{0.24$\pm$0.12}&\textbf{3449$\pm$ 464}\\
  car-evaluation&4&0.99$\pm$0.01&\textbf{3600$\pm$   0}&0.95$\pm$0.06&\textbf{3600$\pm$   0}&\textbf{0.91$\pm$0.10}&\textbf{3600$\pm$   0}\\
  car-evaluation&5&0.99$\pm$0.01&\textbf{3600$\pm$   0}&0.97$\pm$0.04&\textbf{3600$\pm$   0}&\textbf{0.93$\pm$0.08}&\textbf{3600$\pm$   0}\\
  kr-vs-kp&2&0.98$\pm$0.03&3600$\pm$   0&\textbf{0.00$\pm$0.00}& 921$\pm$ 291&\textbf{0.00$\pm$0.00}&\textbf{ 293$\pm$ 151}\\
  kr-vs-kp&3&0.99$\pm$0.01&\textbf{3600$\pm$   0}&0.96$\pm$0.05&\textbf{3600$\pm$   0}&\textbf{0.91$\pm$0.11}&\textbf{3600$\pm$   0}\\
  kr-vs-kp&4&1.00$\pm$0.01&\textbf{3600$\pm$   0}&0.97$\pm$0.03&\textbf{3600$\pm$   0}&\textbf{0.92$\pm$0.09}&\textbf{3600$\pm$   0}\\
  kr-vs-kp&5&1.00$\pm$0.00&\textbf{3600$\pm$   0}&0.97$\pm$0.04&\textbf{3600$\pm$   0}&\textbf{0.93$\pm$0.07}&\textbf{3600$\pm$   0}\\ \hline
\end{tabular}%
}
}
}
% \end{sidewaystable*}
\end{table}
% \end{landscape}

\begin{table}[!ht]
\OneAndAHalfSpacedXII
\begin{center}
\caption{{Average out-of-sample accuracy and standard deviation of accuracy across 5 samples \revision{given the calibrated $\lambda$ on categorical datasets}.  The highest accuracy achieved in a given dataset and depth is reported in \textbf{bold}.}}
\label{tab:out_of_sample}
\small{
\setlength{\tabcolsep}{3pt}
\begin{tabular}{l||c|c|c|c|>{\color{black}}c}
\hline
dataset&depth&\texttt{OCT}&\texttt{BinOCT}&\flow&\benders\\
\hline
soybean-small&2&\textbf{1.00$\pm$0.00}&0.98$\pm$0.04&\textbf{1.00$\pm$0.00}&\textbf{1.00$\pm$0.00}\\soybean-small&3&0.98$\pm$0.04&0.98$\pm$0.04&\textbf{1.00$\pm$0.00}&0.98$\pm$0.04\\soybean-small&4&0.98$\pm$0.04&0.98$\pm$0.04&\textbf{1.00$\pm$0.00}&0.98$\pm$0.04\\soybean-small&5&\textbf{0.98$\pm$0.04}&\textbf{0.98$\pm$0.04}&\textbf{0.98$\pm$0.04}&\textbf{0.98$\pm$0.04}\\monk3&2&\textbf{0.92$\pm$0.02}&\textbf{0.92$\pm$0.02}&\textbf{0.92$\pm$0.02}&\textbf{0.92$\pm$0.02}\\monk3&3&\textbf{0.92$\pm$0.02}&0.91$\pm$0.01&0.91$\pm$0.03&0.91$\pm$0.03\\monk3&4&0.92$\pm$0.02&0.84$\pm$0.08&\textbf{0.92$\pm$0.03}&0.92$\pm$0.02\\monk3&5&\textbf{0.92$\pm$0.03}&0.87$\pm$0.04&\textbf{0.92$\pm$0.03}&\textbf{0.92$\pm$0.03}\\monk1&2&0.71$\pm$0.08&\textbf{0.72$\pm$0.10}&0.71$\pm$0.08&0.71$\pm$0.08\\monk1&3&\textbf{0.83$\pm$0.14}&0.83$\pm$0.07&0.81$\pm$0.13&0.83$\pm$0.13\\monk1&4&\textbf{1.00$\pm$0.00}&0.99$\pm$0.01&\textbf{1.00$\pm$0.00}&\textbf{1.00$\pm$0.00}\\monk1&5&0.88$\pm$0.19&0.97$\pm$0.07&\textbf{1.00$\pm$0.00}&\textbf{1.00$\pm$0.00}\\hayes-roth&2&0.39$\pm$0.09&\textbf{0.45$\pm$0.06}&0.44$\pm$0.08&0.41$\pm$0.10\\hayes-roth&3&0.53$\pm$0.07&\textbf{0.56$\pm$0.07}&0.55$\pm$0.09&0.55$\pm$0.07\\hayes-roth&4&\textbf{0.72$\pm$0.09}&0.71$\pm$0.09&0.72$\pm$0.05&0.72$\pm$0.06\\hayes-roth&5&0.64$\pm$0.12&0.76$\pm$0.06&0.79$\pm$0.05&\textbf{0.81$\pm$0.02}\\monk2&2&\textbf{0.57$\pm$0.06}&0.50$\pm$0.05&\textbf{0.57$\pm$0.06}&\textbf{0.57$\pm$0.06}\\monk2&3&0.58$\pm$0.08&0.59$\pm$0.09&\textbf{0.66$\pm$0.06}&0.63$\pm$0.05\\monk2&4&\textbf{0.63$\pm$0.08}&0.60$\pm$0.04&0.62$\pm$0.06&0.60$\pm$0.06\\monk2&5&0.64$\pm$0.07&0.57$\pm$0.08&\textbf{0.65$\pm$0.05}&0.60$\pm$0.07\\house-votes-84&2&0.78$\pm$0.25&0.96$\pm$0.04&\textbf{0.97$\pm$0.02}&\textbf{0.97$\pm$0.02}\\house-votes-84&3&\textbf{0.97$\pm$0.02}&0.94$\pm$0.02&\textbf{0.97$\pm$0.02}&\textbf{0.97$\pm$0.02}\\house-votes-84&4&\textbf{0.98$\pm$0.02}&0.95$\pm$0.03&0.96$\pm$0.01&0.96$\pm$0.01\\house-votes-84&5&\textbf{0.97$\pm$0.02}&0.93$\pm$0.05&\textbf{0.97$\pm$0.02}&\textbf{0.97$\pm$0.02}\\spect&2&\textbf{0.76$\pm$0.05}&0.74$\pm$0.05&\textbf{0.76$\pm$0.05}&\textbf{0.76$\pm$0.05}\\spect&3&\textbf{0.76$\pm$0.05}&0.73$\pm$0.04&\textbf{0.76$\pm$0.05}&\textbf{0.76$\pm$0.05}\\spect&4&\textbf{0.76$\pm$0.05}&0.75$\pm$0.05&\textbf{0.76$\pm$0.05}&\textbf{0.76$\pm$0.05}\\spect&5&0.74$\pm$0.02&0.74$\pm$0.08&0.75$\pm$0.04&\textbf{0.76$\pm$0.05}\\breast-cancer&2&0.72$\pm$0.04&0.71$\pm$0.03&\textbf{0.73$\pm$0.05}&0.72$\pm$0.04\\breast-cancer&3&0.74$\pm$0.04&0.71$\pm$0.05&\textbf{0.75$\pm$0.02}&0.72$\pm$0.04\\breast-cancer&4&0.72$\pm$0.05&0.66$\pm$0.08&\textbf{0.73$\pm$0.04}&0.71$\pm$0.04\\breast-cancer&5&\textbf{0.74$\pm$0.01}&0.71$\pm$0.03&0.73$\pm$0.05&0.72$\pm$0.02\\balance-scale&2&\textbf{0.69$\pm$0.02}&0.68$\pm$0.03&\textbf{0.69$\pm$0.02}&\textbf{0.69$\pm$0.02}\\balance-scale&3&0.70$\pm$0.02&\textbf{0.72$\pm$0.03}&0.70$\pm$0.03&0.71$\pm$0.03\\balance-scale&4&0.67$\pm$0.04&\textbf{0.74$\pm$0.03}&0.73$\pm$0.03&0.72$\pm$0.03\\balance-scale&5&0.61$\pm$0.05&\textbf{0.76$\pm$0.02}&0.73$\pm$0.02&0.75$\pm$0.03\\tic-tac-toe&2&\textbf{0.67$\pm$0.02}&0.66$\pm$0.02&\textbf{0.67$\pm$0.02}&\textbf{0.67$\pm$0.02}\\tic-tac-toe&3&0.69$\pm$0.03&0.72$\pm$0.02&0.72$\pm$0.01&\textbf{0.72$\pm$0.03}\\tic-tac-toe&4&0.70$\pm$0.04&\textbf{0.78$\pm$0.03}&0.77$\pm$0.02&\textbf{0.78$\pm$0.03}\\tic-tac-toe&5&0.68$\pm$0.03&\textbf{0.80$\pm$0.04}&0.80$\pm$0.02&0.78$\pm$0.04\\car-evaluation&2&\textbf{0.77$\pm$0.01}&\textbf{0.77$\pm$0.01}&\textbf{0.77$\pm$0.01}&\textbf{0.77$\pm$0.01}\\car-evaluation&3&0.72$\pm$0.02&0.78$\pm$0.01&\textbf{0.79$\pm$0.01}&\textbf{0.79$\pm$0.01}\\car-evaluation&4&0.72$\pm$0.02&\textbf{0.81$\pm$0.01}&0.79$\pm$0.01&\textbf{0.81$\pm$0.01}\\car-evaluation&5&0.76$\pm$0.03&0.82$\pm$0.01&0.76$\pm$0.04&\textbf{0.84$\pm$0.01}\\kr-vs-kp&2&0.68$\pm$0.06&\textbf{0.87$\pm$0.01}&\textbf{0.87$\pm$0.01}&\textbf{0.87$\pm$0.01}\\kr-vs-kp&3&0.68$\pm$0.08&0.86$\pm$0.06&0.74$\pm$0.05&\textbf{0.92$\pm$0.02}\\kr-vs-kp&4&0.58$\pm$0.07&0.90$\pm$0.03&0.80$\pm$0.04&\textbf{0.94$\pm$0.00}\\kr-vs-kp&5&0.66$\pm$0.11&0.87$\pm$0.08&0.84$\pm$0.06&\textbf{0.89$\pm$0.08}\\
\hline
\end{tabular}
}
\end{center}
\end{table}

\revision{

\begin{table}[!ht]
\OneAndAHalfSpacedXII
\begin{center}
\caption{Average in-sample accuracy and standard deviation of accuracy across 5 samples for the case of $\lambda=0$ \revision{on categorical datasets}.  The highest accuracy achieved in a given dataset and depth is reported in \textbf{bold}.}
\label{tab:in_sample_lst_comparison}
\small{
\setlength{\tabcolsep}{3pt}
\begin{tabular}{l||c|c|c|c}
\hline
dataset&depth&\texttt{OCT}&\texttt{LST}&\benders\\
\hline
soybean-small&2&\textbf{1.00$\pm$0.00}&\textbf{1.00$\pm$0.00}&\textbf{1.00$\pm$0.00}\\
soybean-small&3&\textbf{1.00$\pm$0.00}&\textbf{1.00$\pm$0.00}&\textbf{1.00$\pm$0.00}\\
soybean-small&4&\textbf{1.00$\pm$0.00}&\textbf{1.00$\pm$0.00}&\textbf{1.00$\pm$0.00}\\
soybean-small&5&\textbf{1.00$\pm$0.00}&\textbf{1.00$\pm$0.00}&\textbf{1.00$\pm$0.00}\\
monk3&2&\textbf{0.94$\pm$0.01}&\textbf{0.94$\pm$0.01}&\textbf{0.94$\pm$0.01}\\
monk3&3&\textbf{0.98$\pm$0.01}&0.96$\pm$0.01&\textbf{0.98$\pm$0.01}\\
monk3&4&\textbf{1.00$\pm$0.00}&0.99$\pm$0.01&\textbf{1.00$\pm$0.00}\\
monk3&5&\textbf{1.00$\pm$0.00}&\textbf{1.00$\pm$0.00}&\textbf{1.00$\pm$0.00}\\
monk1&2&\textbf{0.86$\pm$0.03}&\textbf{0.86$\pm$0.03}&\textbf{0.86$\pm$0.03}\\
monk1&3&\textbf{0.95$\pm$0.02}&\textbf{0.95$\pm$0.02}&\textbf{0.95$\pm$0.02}\\
monk1&4&\textbf{1.00$\pm$0.00}&\textbf{1.00$\pm$0.00}&\textbf{1.00$\pm$0.00}\\
monk1&5&\textbf{1.00$\pm$0.00}&\textbf{1.00$\pm$0.00}&\textbf{1.00$\pm$0.00}\\
hayes-roth&2&\textbf{0.66$\pm$0.02}&\textbf{0.66$\pm$0.02}&\textbf{0.66$\pm$0.02}\\
hayes-roth&3&\textbf{0.81$\pm$0.04}&0.80$\pm$0.04&\textbf{0.81$\pm$0.04}\\
hayes-roth&4&0.87$\pm$0.03&0.87$\pm$0.03&\textbf{0.89$\pm$0.02}\\
hayes-roth&5&0.89$\pm$0.02&0.91$\pm$0.02&\textbf{0.92$\pm$0.02}\\
monk2&2&\textbf{0.71$\pm$0.02}&\textbf{0.71$\pm$0.02}&\textbf{0.71$\pm$0.02}\\
monk2&3&0.80$\pm$0.02&\textbf{0.81$\pm$0.02}&\textbf{0.81$\pm$0.02}\\
monk2&4&0.87$\pm$0.03&0.88$\pm$0.02&\textbf{0.90$\pm$0.02}\\
monk2&5&0.91$\pm$0.02&0.93$\pm$0.02&\textbf{0.94$\pm$0.01}\\
house-votes-84&2&\textbf{0.97$\pm$0.00}&\textbf{0.97$\pm$0.00}&\textbf{0.97$\pm$0.00}\\
house-votes-84&3&\textbf{0.99$\pm$0.00}&0.98$\pm$0.01&\textbf{0.99$\pm$0.00}\\
house-votes-84&4&\textbf{1.00$\pm$0.00}&\textbf{1.00$\pm$0.00}&\textbf{1.00$\pm$0.00}\\
house-votes-84&5&\textbf{1.00$\pm$0.00}&\textbf{1.00$\pm$0.00}&\textbf{1.00$\pm$0.00}\\
spect&2&\textbf{0.81$\pm$0.01}&\textbf{0.81$\pm$0.01}&\textbf{0.81$\pm$0.01}\\
spect&3&\textbf{0.86$\pm$0.02}&\textbf{0.86$\pm$0.02}&\textbf{0.86$\pm$0.02}\\
spect&4&0.88$\pm$0.07&0.91$\pm$0.01&\textbf{0.92$\pm$0.01}\\
spect&5&0.91$\pm$0.02&0.93$\pm$0.01&\textbf{0.95$\pm$0.01}\\
breast-cancer&2&\textbf{0.80$\pm$0.02}&\textbf{0.80$\pm$0.02}&\textbf{0.80$\pm$0.02}\\
breast-cancer&3&\textbf{0.85$\pm$0.02}&\textbf{0.85$\pm$0.02}&\textbf{0.85$\pm$0.02}\\
breast-cancer&4&0.87$\pm$0.02&\textbf{0.89$\pm$0.02}&\textbf{0.89$\pm$0.02}\\
breast-cancer&5&0.89$\pm$0.04&0.92$\pm$0.02&\textbf{0.92$\pm$0.03}\\
balance-scale&2&\textbf{0.70$\pm$0.01}&\textbf{0.70$\pm$0.01}&\textbf{0.70$\pm$0.01}\\
balance-scale&3&0.75$\pm$0.02&\textbf{0.77$\pm$0.01}&\textbf{0.77$\pm$0.01}\\
balance-scale&4&0.74$\pm$0.03&\textbf{0.80$\pm$0.01}&\textbf{0.80$\pm$0.01}\\
balance-scale&5&0.76$\pm$0.05&\textbf{0.84$\pm$0.01}&0.83$\pm$0.01\\
tic-tac-toe&2&\textbf{0.73$\pm$0.01}&\textbf{0.73$\pm$0.01}&\textbf{0.73$\pm$0.01}\\
tic-tac-toe&3&0.75$\pm$0.02&\textbf{0.79$\pm$0.01}&\textbf{0.79$\pm$0.01}\\
tic-tac-toe&4&0.76$\pm$0.05&\textbf{0.86$\pm$0.00}&0.85$\pm$0.02\\
tic-tac-toe&5&0.70$\pm$0.01&\textbf{0.93$\pm$0.01}&0.88$\pm$0.02\\
car-evaluation&2&\textbf{0.78$\pm$0.01}&\textbf{0.78$\pm$0.01}&\textbf{0.78$\pm$0.01}\\
car-evaluation&3&0.77$\pm$0.04&\textbf{0.82$\pm$0.01}&\textbf{0.82$\pm$0.01}\\
car-evaluation&4&0.78$\pm$0.02&0.84$\pm$0.00&\textbf{0.84$\pm$0.01}\\
car-evaluation&5&0.73$\pm$0.03&\textbf{0.88$\pm$0.00}&0.86$\pm$0.02\\
kr-vs-kp&2&0.83$\pm$0.05&\textbf{0.87$\pm$0.01}&\textbf{0.87$\pm$0.01}\\
kr-vs-kp&3&0.72$\pm$0.04&\textbf{0.94$\pm$0.01}&0.92$\pm$0.03\\
kr-vs-kp&4&0.70$\pm$0.06&\textbf{0.95$\pm$0.00}&0.91$\pm$0.05\\
kr-vs-kp&5&0.65$\pm$0.06&\textbf{0.97$\pm$0.01}&0.93$\pm$0.03\\
\hline
\end{tabular}
}
\end{center}
\end{table}

% \begin{sidewaystable*}[t]
\begin{landscape}
\begin{table}
\OneAndAHalfSpacedXII
\caption{\revision{Average in-sample accuracy and standard deviation of accuracy across 5 samples for the case of $\lambda = 0$ on mixed-feature datasets (part 1). {Due to the numerical issues of \texttt{OCT} (as discussed in Electronic Companion~\ref{appendix_sec:oct_numerical_issue}), we do not provide solving time and optimality gap for this approach as it tackles a different problem. Instead, we report the number of instances (out of five samples) in a given dataset and depth, where we observe a discrepancy of at least $0.001$ between the objective value of the optimization problem and the actual in-sample accuracy.}
The best performance achieved in a given dataset and depth is reported in \textbf{bold}.}}
\small{
\label{tab:in_sample_non_categorical_data_no_reg_part_1}
\setlength{\tabcolsep}{4pt}
% \resizebox{1.3\textheight}{!}{%
\scalebox{0.9}{
\color{black}\begin{tabular}{lc||cc|ccc|ccc}
\hline
Dataset &
  Depth &
  \multicolumn{2}{c|}{\texttt{OCT}} &
  \multicolumn{3}{c|}{\bendersFiveBuckets} &
  \multicolumn{3}{c}{\bendersTenBuckets} \\
 & & Train-acc & Numerical Issues  & Train-acc & Gap & Time & Train-acc & Gap & Time \\
  \hline
  echocardiogram&2&\textbf{1.00$\pm$0.00}&0&\textbf{1.00$\pm$0.00}&\textbf{0.00$\pm$0.00}&\textbf{   1$\pm$   0}&\textbf{1.00$\pm$0.00}&\textbf{0.00$\pm$0.00}&\textbf{   1$\pm$   0}\\
  echocardiogram&3&\textbf{1.00$\pm$0.00}&0&\textbf{1.00$\pm$0.00}&\textbf{0.00$\pm$0.00}&\textbf{   1$\pm$   0}&\textbf{1.00$\pm$0.00}&\textbf{0.00$\pm$0.00}&\textbf{   1$\pm$   0}\\
  echocardiogram&4&0.99$\pm$0.01&1&\textbf{1.00$\pm$0.00}&\textbf{0.00$\pm$0.00}&\textbf{   2$\pm$   0}&\textbf{1.00$\pm$0.00}&\textbf{0.00$\pm$0.00}&   2$\pm$   2\\
  echocardiogram&5&\textbf{1.00$\pm$0.00}&0&\textbf{1.00$\pm$0.00}&\textbf{0.00$\pm$0.00}&\textbf{   0$\pm$   0}&\textbf{1.00$\pm$0.00}&\textbf{0.00$\pm$0.00}&   3$\pm$   1\\
  hepatitis&2&\textbf{0.98$\pm$0.01}&1&0.96$\pm$0.03&\textbf{0.00$\pm$0.00}&\textbf{   3$\pm$   2}&0.96$\pm$0.03&\textbf{0.00$\pm$0.00}&   9$\pm$   7\\
  hepatitis&3&\textbf{1.00$\pm$0.00}&0&1.00$\pm$0.01&0.20$\pm$0.45& 722$\pm$1609&\textbf{1.00$\pm$0.00}&\textbf{0.00$\pm$0.00}&\textbf{ 224$\pm$ 495}\\
  hepatitis&4&1.00$\pm$0.01&1&\textbf{1.00$\pm$0.00}&\textbf{0.00$\pm$0.00}&   8$\pm$   7&\textbf{1.00$\pm$0.00}&\textbf{0.00$\pm$0.00}&\textbf{   6$\pm$   3}\\
  hepatitis&5&0.99$\pm$0.01&2&\textbf{1.00$\pm$0.00}&\textbf{0.00$\pm$0.00}&\textbf{   5$\pm$   2}&\textbf{1.00$\pm$0.00}&\textbf{0.00$\pm$0.00}&   8$\pm$   3\\
  fertility&2&\textbf{0.91$\pm$0.01}&0&0.90$\pm$0.01&\textbf{0.00$\pm$0.00}&\textbf{   4$\pm$   1}&0.90$\pm$0.01&\textbf{0.00$\pm$0.00}&\textbf{   4$\pm$   1}\\
  fertility&3&\textbf{0.97$\pm$0.02}&0&0.96$\pm$0.01&\textbf{0.23$\pm$0.32}&\textbf{2504$\pm$1193}&0.96$\pm$0.01&\textbf{0.23$\pm$0.32}&2692$\pm$1062\\
  fertility&4&\textbf{0.98$\pm$0.02}&1&\textbf{0.98$\pm$0.02}&\textbf{0.60$\pm$0.55}&2220$\pm$1893&\textbf{0.98$\pm$0.02}&\textbf{0.60$\pm$0.55}&\textbf{2213$\pm$1901}\\
  fertility&5&0.97$\pm$0.05&1&\textbf{0.99$\pm$0.01}&\textbf{0.60$\pm$0.55}&\textbf{2171$\pm$1956}&\textbf{0.99$\pm$0.01}&\textbf{0.60$\pm$0.55}&2172$\pm$1956\\
  iris&2&0.96$\pm$0.04&2&0.95$\pm$0.03&\textbf{0.00$\pm$0.00}&\textbf{   1$\pm$   0}&\textbf{0.96$\pm$0.02}&\textbf{0.00$\pm$0.00}&   2$\pm$   0\\
  iris&3&\textbf{0.99$\pm$0.01}&2&0.98$\pm$0.02&\textbf{0.00$\pm$0.00}&\textbf{  31$\pm$  31}&\textbf{0.99$\pm$0.01}&\textbf{0.00$\pm$0.00}& 267$\pm$ 427\\
  iris&4&0.99$\pm$0.01&2&0.99$\pm$0.01&0.40$\pm$0.55&1443$\pm$1969&\textbf{1.00$\pm$0.01}&\textbf{0.20$\pm$0.45}&\textbf{1216$\pm$1706}\\
  iris&5&0.98$\pm$0.02&3&0.99$\pm$0.01&0.40$\pm$0.55&1444$\pm$1969&\textbf{1.00$\pm$0.00}&\textbf{0.00$\pm$0.00}&\textbf{  25$\pm$  31}\\
  wine&2&\textbf{0.96$\pm$0.01}&3&0.95$\pm$0.02&\textbf{0.00$\pm$0.00}&\textbf{   8$\pm$   3}&0.95$\pm$0.01&\textbf{0.00$\pm$0.00}&  31$\pm$   9\\
  wine&3&0.99$\pm$0.01&2&1.00$\pm$0.01&0.10$\pm$0.22&\textbf{ 736$\pm$1601}&\textbf{1.00$\pm$0.00}&\textbf{0.00$\pm$0.00}& 783$\pm$ 857\\
  wine&4&0.99$\pm$0.01&3&\textbf{1.00$\pm$0.00}&\textbf{0.00$\pm$0.00}& 175$\pm$ 186&\textbf{1.00$\pm$0.00}&\textbf{0.00$\pm$0.00}&\textbf{  42$\pm$  13}\\
  wine&5&0.99$\pm$0.01&3&\textbf{1.00$\pm$0.00}&\textbf{0.00$\pm$0.00}&\textbf{  41$\pm$  11}&\textbf{1.00$\pm$0.00}&\textbf{0.00$\pm$0.00}&  91$\pm$  65\\
  planning-relax&2&0.68$\pm$0.27&2&0.79$\pm$0.03&\textbf{0.00$\pm$0.00}&\textbf{  83$\pm$  27}&\textbf{0.81$\pm$0.03}&\textbf{0.00$\pm$0.00}& 861$\pm$ 373\\
  planning-relax&3&0.72$\pm$0.19&4&\textbf{0.86$\pm$0.03}&\textbf{0.99$\pm$0.02}&\textbf{3600$\pm$   0}&\textbf{0.86$\pm$0.03}&1.00$\pm$0.00&\textbf{3600$\pm$   0}\\
  planning-relax&4&0.67$\pm$0.20&5&\textbf{0.91$\pm$0.03}&\textbf{1.00$\pm$0.00}&\textbf{3600$\pm$   0}&\textbf{0.91$\pm$0.03}&\textbf{1.00$\pm$0.00}&\textbf{3600$\pm$   0}\\
  planning-relax&5&0.86$\pm$0.13&3&\textbf{0.97$\pm$0.03}&\textbf{0.60$\pm$0.55}&\textbf{3075$\pm$1016}&0.93$\pm$0.05&1.00$\pm$0.00&3600$\pm$   0\\
  breast-cancer-prognostic&2&\textbf{0.87$\pm$0.02}&3&0.85$\pm$0.02&\textbf{0.00$\pm$0.00}&\textbf{ 865$\pm$ 283}&0.86$\pm$0.03&0.83$\pm$0.08&3600$\pm$   0\\
  breast-cancer-prognostic&3&\textbf{0.93$\pm$0.02}&1&0.91$\pm$0.02&\textbf{1.00$\pm$0.00}&\textbf{3600$\pm$   0}&0.92$\pm$0.02&\textbf{1.00$\pm$0.00}&\textbf{3600$\pm$   0}\\
  breast-cancer-prognostic&4&\textbf{0.97$\pm$0.04}&2&0.94$\pm$0.02&\textbf{1.00$\pm$0.00}&\textbf{3600$\pm$   0}&0.94$\pm$0.02&\textbf{1.00$\pm$0.00}&\textbf{3600$\pm$   0}\\
  breast-cancer-prognostic&5&\textbf{0.99$\pm$0.01}&2&0.95$\pm$0.04&\textbf{1.00$\pm$0.00}&\textbf{3600$\pm$   0}&0.93$\pm$0.03&\textbf{1.00$\pm$0.00}&\textbf{3600$\pm$   0}\\
  parkinsons&2&\textbf{0.95$\pm$0.01}&0&0.91$\pm$0.01&\textbf{0.00$\pm$0.00}&\textbf{  36$\pm$   8}&0.93$\pm$0.01&\textbf{0.00$\pm$0.00}& 291$\pm$ 102\\
  parkinsons&3&\textbf{0.99$\pm$0.01}&2&0.97$\pm$0.01&\textbf{1.00$\pm$0.00}&\textbf{3600$\pm$   0}&0.98$\pm$0.01&\textbf{1.00$\pm$0.00}&\textbf{3600$\pm$   0}\\
  parkinsons&4&0.99$\pm$0.01&3&\textbf{0.99$\pm$0.00}&0.80$\pm$0.45&3255$\pm$ 772&0.99$\pm$0.01&\textbf{0.60$\pm$0.55}&\textbf{2817$\pm$1123}\\
  parkinsons&5&0.99$\pm$0.01&3&\textbf{1.00$\pm$0.00}&\textbf{0.00$\pm$0.00}&\textbf{ 426$\pm$ 609}&0.99$\pm$0.01&0.60$\pm$0.55&2251$\pm$1849\\
  connectionist-bench-sonar&2&\textbf{0.84$\pm$0.01}&2&0.83$\pm$0.01&\textbf{0.54$\pm$0.42}&\textbf{3600$\pm$   0}&0.83$\pm$0.02&1.00$\pm$0.00&\textbf{3600$\pm$   0}\\
  connectionist-bench-sonar&3&0.88$\pm$0.03&3&\textbf{0.89$\pm$0.02}&\textbf{1.00$\pm$0.00}&\textbf{3600$\pm$   0}&\textbf{0.89$\pm$0.02}&\textbf{1.00$\pm$0.00}&\textbf{3600$\pm$   0}\\
  connectionist-bench-sonar&4&0.91$\pm$0.06&3&\textbf{0.95$\pm$0.03}&\textbf{1.00$\pm$0.00}&\textbf{3600$\pm$   0}&0.94$\pm$0.02&\textbf{1.00$\pm$0.00}&\textbf{3600$\pm$   0}\\
  connectionist-bench-sonar&5&0.96$\pm$0.04&3&\textbf{0.98$\pm$0.02}&\textbf{0.80$\pm$0.45}&\textbf{2995$\pm$1354}&0.89$\pm$0.03&1.00$\pm$0.00&3600$\pm$   0\\ \hline
\end{tabular}%
}
}
% \end{sidewaystable*}
\end{table}
\end{landscape}

% \begin{sidewaystable*}[t]
\begin{landscape}
\begin{table}
\OneAndAHalfSpacedXII
\caption{\revision{Average in-sample accuracy and standard deviation of accuracy across 5 samples for the case of $\lambda = 0$ on mixed-feature datasets (part 2).
The best performance achieved in a given dataset and depth is reported in \textbf{bold}.}}
\small{
\label{tab:in_sample_non_categorical_data_no_reg_part_2}
\setlength{\tabcolsep}{4pt}
% \resizebox{1.3\textheight}{!}{%
\scalebox{0.9}{
\color{black}\begin{tabular}{lc||cc|ccc|ccc}
\hline
Dataset &
  Depth &
  \multicolumn{2}{c|}{\texttt{OCT}} &
  \multicolumn{3}{c|}{\bendersFiveBuckets} &
  \multicolumn{3}{c}{\bendersTenBuckets} \\
 & & Train-acc & Numerical Issues  & Train-acc & Gap & Time & Train-acc & Gap & Time \\
  \hline
  seeds&2&\textbf{0.96$\pm$0.01}&1&0.90$\pm$0.02&\textbf{0.00$\pm$0.00}&\textbf{   3$\pm$   1}&0.93$\pm$0.01&\textbf{0.00$\pm$0.00}&   9$\pm$   1\\
  seeds&3&\textbf{0.98$\pm$0.01}&4&0.94$\pm$0.02&\textbf{0.04$\pm$0.06}&\textbf{1654$\pm$1799}&0.97$\pm$0.02&0.42$\pm$0.39&2977$\pm$ 856\\
  seeds&4&\textbf{0.99$\pm$0.01}&3&0.97$\pm$0.01&1.00$\pm$0.00&3600$\pm$   0&\textbf{0.99$\pm$0.01}&\textbf{0.80$\pm$0.45}&\textbf{3586$\pm$  30}\\
  seeds&5&\textbf{0.99$\pm$0.01}&3&0.98$\pm$0.01&1.00$\pm$0.00&3600$\pm$   0&\textbf{0.99$\pm$0.01}&\textbf{0.40$\pm$0.55}&\textbf{1938$\pm$1537}\\
  cylinder-bands&2&0.74$\pm$0.02&1&\textbf{0.76$\pm$0.01}&\textbf{0.26$\pm$0.27}&\textbf{3515$\pm$ 191}&0.75$\pm$0.02&0.98$\pm$0.03&3600$\pm$   0\\
  cylinder-bands&3&0.81$\pm$0.02&0&0.81$\pm$0.01&\textbf{1.00$\pm$0.00}&\textbf{3600$\pm$   0}&\textbf{0.82$\pm$0.02}&\textbf{1.00$\pm$0.00}&\textbf{3600$\pm$   0}\\
  cylinder-bands&4&\textbf{0.85$\pm$0.02}&1&0.79$\pm$0.05&\textbf{1.00$\pm$0.00}&\textbf{3600$\pm$   0}&0.81$\pm$0.03&\textbf{1.00$\pm$0.00}&\textbf{3600$\pm$   0}\\
  cylinder-bands&5&\textbf{0.89$\pm$0.01}&3&0.82$\pm$0.07&\textbf{1.00$\pm$0.00}&\textbf{3600$\pm$   0}&0.82$\pm$0.05&\textbf{1.00$\pm$0.00}&\textbf{3600$\pm$   0}\\
  heart-cleveland&2&\textbf{0.63$\pm$0.02}&2&\textbf{0.63$\pm$0.02}&\textbf{0.00$\pm$0.00}&\textbf{ 118$\pm$  87}&\textbf{0.63$\pm$0.02}&\textbf{0.00$\pm$0.00}& 237$\pm$  69\\
  heart-cleveland&3&0.68$\pm$0.02&1&0.69$\pm$0.02&\textbf{0.97$\pm$0.02}&\textbf{3600$\pm$   0}&\textbf{0.71$\pm$0.02}&1.00$\pm$0.00&\textbf{3600$\pm$   0}\\
  heart-cleveland&4&0.71$\pm$0.02&3&\textbf{0.76$\pm$0.03}&\textbf{1.00$\pm$0.00}&\textbf{3600$\pm$   0}&0.75$\pm$0.04&\textbf{1.00$\pm$0.00}&\textbf{3600$\pm$   0}\\
  heart-cleveland&5&0.73$\pm$0.03&4&\textbf{0.80$\pm$0.05}&\textbf{1.00$\pm$0.00}&\textbf{3600$\pm$   0}&0.76$\pm$0.07&\textbf{1.00$\pm$0.00}&\textbf{3600$\pm$   0}\\
  ionosphere&2&\textbf{0.92$\pm$0.01}&4&0.89$\pm$0.01&\textbf{0.00$\pm$0.00}&\textbf{ 522$\pm$ 466}&0.90$\pm$0.01&0.32$\pm$0.45&2999$\pm$ 782\\
  ionosphere&3&0.94$\pm$0.01&4&0.94$\pm$0.01&\textbf{1.00$\pm$0.00}&\textbf{3600$\pm$   0}&\textbf{0.95$\pm$0.01}&\textbf{1.00$\pm$0.00}&\textbf{3600$\pm$   0}\\
  ionosphere&4&0.95$\pm$0.01&3&\textbf{0.96$\pm$0.01}&\textbf{1.00$\pm$0.00}&\textbf{3600$\pm$   0}&\textbf{0.96$\pm$0.01}&\textbf{1.00$\pm$0.00}&\textbf{3600$\pm$   0}\\
  ionosphere&5&0.96$\pm$0.01&4&\textbf{0.98$\pm$0.01}&\textbf{1.00$\pm$0.00}&\textbf{3600$\pm$   0}&0.97$\pm$0.02&\textbf{1.00$\pm$0.00}&\textbf{3600$\pm$   0}\\
  thoracic-surgery&2&\textbf{0.88$\pm$0.01}&1&0.87$\pm$0.01&\textbf{0.00$\pm$0.00}&\textbf{  41$\pm$  12}&\textbf{0.88$\pm$0.01}&\textbf{0.00$\pm$0.00}& 188$\pm$  49\\
  thoracic-surgery&3&0.89$\pm$0.02&1&\textbf{0.89$\pm$0.01}&\textbf{0.99$\pm$0.02}&\textbf{3600$\pm$   0}&0.89$\pm$0.02&1.00$\pm$0.00&\textbf{3600$\pm$   0}\\
  thoracic-surgery&4&0.89$\pm$0.01&4&\textbf{0.90$\pm$0.02}&\textbf{1.00$\pm$0.00}&\textbf{3600$\pm$   0}&\textbf{0.90$\pm$0.02}&\textbf{1.00$\pm$0.00}&\textbf{3600$\pm$   0}\\
  thoracic-surgery&5&0.91$\pm$0.02&3&\textbf{0.92$\pm$0.01}&\textbf{1.00$\pm$0.00}&\textbf{3600$\pm$   0}&\textbf{0.92$\pm$0.01}&\textbf{1.00$\pm$0.00}&\textbf{3600$\pm$   0}\\
  climate&2&0.87$\pm$0.16&2&0.93$\pm$0.01&\textbf{0.00$\pm$0.00}&\textbf{ 654$\pm$ 224}&\textbf{0.94$\pm$0.01}&0.70$\pm$0.18&3600$\pm$   0\\
  climate&3&0.90$\pm$0.12&3&\textbf{0.95$\pm$0.01}&\textbf{1.00$\pm$0.00}&\textbf{3600$\pm$   0}&\textbf{0.95$\pm$0.01}&\textbf{1.00$\pm$0.00}&\textbf{3600$\pm$   0}\\
  climate&4&0.95$\pm$0.03&4&\textbf{0.97$\pm$0.01}&\textbf{1.00$\pm$0.00}&\textbf{3600$\pm$   0}&0.96$\pm$0.01&\textbf{1.00$\pm$0.00}&\textbf{3600$\pm$   0}\\
  climate&5&0.94$\pm$0.06&3&0.97$\pm$0.02&\textbf{1.00$\pm$0.00}&\textbf{3600$\pm$   0}&\textbf{0.97$\pm$0.01}&\textbf{1.00$\pm$0.00}&\textbf{3600$\pm$   0}\\
  breast-cancer-diagnostic&2&\textbf{0.97$\pm$0.01}&2&0.95$\pm$0.01&\textbf{0.00$\pm$0.00}&\textbf{ 344$\pm$ 162}&0.96$\pm$0.01&0.10$\pm$0.22&1948$\pm$1102\\
  breast-cancer-diagnostic&3&\textbf{0.98$\pm$0.00}&3&0.97$\pm$0.01&\textbf{1.00$\pm$0.00}&\textbf{3600$\pm$   0}&0.98$\pm$0.01&\textbf{1.00$\pm$0.00}&\textbf{3600$\pm$   0}\\
  breast-cancer-diagnostic&4&\textbf{0.99$\pm$0.01}&3&0.98$\pm$0.01&\textbf{1.00$\pm$0.00}&\textbf{3600$\pm$   0}&\textbf{0.99$\pm$0.01}&\textbf{1.00$\pm$0.00}&\textbf{3600$\pm$   0}\\
  breast-cancer-diagnostic&5&0.99$\pm$0.01&3&0.99$\pm$0.01&\textbf{0.60$\pm$0.55}&\textbf{3056$\pm$ 747}&\textbf{0.99$\pm$0.00}&1.00$\pm$0.00&3600$\pm$   0\\
  indian-liver-patient&2&\textbf{0.76$\pm$0.02}&0&0.74$\pm$0.02&\textbf{0.00$\pm$0.00}&\textbf{ 248$\pm$  99}&0.75$\pm$0.02&\textbf{0.00$\pm$0.00}&1326$\pm$ 174\\
  indian-liver-patient&3&\textbf{0.78$\pm$0.01}&4&0.77$\pm$0.02&\textbf{0.99$\pm$0.02}&\textbf{3600$\pm$   0}&\textbf{0.78$\pm$0.01}&1.00$\pm$0.00&\textbf{3600$\pm$   0}\\
  indian-liver-patient&4&\textbf{0.80$\pm$0.02}&4&\textbf{0.80$\pm$0.02}&\textbf{1.00$\pm$0.00}&\textbf{3600$\pm$   0}&\textbf{0.80$\pm$0.02}&\textbf{1.00$\pm$0.00}&\textbf{3600$\pm$   0}\\
  indian-liver-patient&5&0.83$\pm$0.02&5&0.82$\pm$0.02&\textbf{1.00$\pm$0.00}&\textbf{3600$\pm$   0}&\textbf{0.84$\pm$0.01}&\textbf{1.00$\pm$0.00}&\textbf{3600$\pm$   0}\\
  credit-approval&2&0.87$\pm$0.01&5&\textbf{0.88$\pm$0.01}&\textbf{0.00$\pm$0.00}&\textbf{ 274$\pm$ 162}&\textbf{0.88$\pm$0.01}&\textbf{0.00$\pm$0.00}&1070$\pm$ 266\\
  credit-approval&3&0.85$\pm$0.09&5&\textbf{0.89$\pm$0.01}&\textbf{1.00$\pm$0.00}&\textbf{3600$\pm$   0}&\textbf{0.89$\pm$0.01}&\textbf{1.00$\pm$0.00}&\textbf{3600$\pm$   0}\\
  credit-approval&4&0.84$\pm$0.05&5&\textbf{0.91$\pm$0.01}&\textbf{1.00$\pm$0.00}&\textbf{3600$\pm$   0}&0.90$\pm$0.01&\textbf{1.00$\pm$0.00}&\textbf{3600$\pm$   0}\\
  credit-approval&5&0.83$\pm$0.04&5&0.91$\pm$0.01&\textbf{1.00$\pm$0.00}&\textbf{3600$\pm$   0}&\textbf{0.92$\pm$0.02}&\textbf{1.00$\pm$0.00}&\textbf{3600$\pm$   0}\\ \hline
\end{tabular}%
}
}
% \end{sidewaystable*}
\end{table}
\end{landscape}

% \begin{sidewaystable*}[t]
\begin{landscape}
\begin{table}
\OneAndAHalfSpacedXII
\caption{\revision{Average in-sample accuracy and standard deviation of accuracy across 5 samples for the case of $\lambda = 0$ on mixed-feature datasets (part 3).
The best performance achieved in a given dataset and depth is reported in \textbf{bold}.}}
\small{
\label{tab:in_sample_non_categorical_data_no_reg_part_3}
\setlength{\tabcolsep}{4pt}
% \resizebox{1.3\textheight}{!}{%
\scalebox{0.9}{
\color{black}\begin{tabular}{lc||cc|ccc|ccc}
\hline
Dataset &
  Depth &
  \multicolumn{2}{c|}{\texttt{OCT}} &
  \multicolumn{3}{c|}{\bendersFiveBuckets} &
  \multicolumn{3}{c}{\bendersTenBuckets} \\
 & & Train-acc & Numerical Issues  & Train-acc & Gap & Time & Train-acc & Gap & Time \\
  \hline
  blood-transfusion&2&\textbf{0.79$\pm$0.01}&1&0.77$\pm$0.01&\textbf{0.00$\pm$0.00}&\textbf{  17$\pm$   4}&0.78$\pm$0.01&\textbf{0.00$\pm$0.00}&  68$\pm$  24\\
  blood-transfusion&3&\textbf{0.81$\pm$0.01}&4&0.79$\pm$0.01&\textbf{0.68$\pm$0.11}&\textbf{3600$\pm$   0}&\textbf{0.81$\pm$0.01}&0.81$\pm$0.03&\textbf{3600$\pm$   0}\\
  blood-transfusion&4&0.81$\pm$0.03&5&0.80$\pm$0.01&0.96$\pm$0.02&\textbf{3600$\pm$   0}&\textbf{0.82$\pm$0.01}&\textbf{0.93$\pm$0.02}&\textbf{3600$\pm$   0}\\
  blood-transfusion&5&0.81$\pm$0.03&5&0.81$\pm$0.01&0.96$\pm$0.02&\textbf{3600$\pm$   0}&\textbf{0.82$\pm$0.01}&\textbf{0.95$\pm$0.02}&\textbf{3600$\pm$   0}\\
  diabetes&2&\textbf{0.78$\pm$0.01}&2&0.77$\pm$0.00&\textbf{0.00$\pm$0.00}&\textbf{ 216$\pm$ 173}&0.77$\pm$0.00&\textbf{0.00$\pm$0.00}&1059$\pm$ 290\\
  diabetes&3&\textbf{0.80$\pm$0.01}&3&0.79$\pm$0.01&\textbf{1.00$\pm$0.00}&\textbf{3600$\pm$   0}&0.79$\pm$0.01&\textbf{1.00$\pm$0.00}&\textbf{3600$\pm$   0}\\
  diabetes&4&\textbf{0.81$\pm$0.02}&4&0.80$\pm$0.01&\textbf{1.00$\pm$0.00}&\textbf{3600$\pm$   0}&0.79$\pm$0.02&\textbf{1.00$\pm$0.00}&\textbf{3600$\pm$   0}\\
  diabetes&5&\textbf{0.83$\pm$0.02}&4&0.82$\pm$0.01&\textbf{1.00$\pm$0.00}&\textbf{3600$\pm$   0}&0.81$\pm$0.02&\textbf{1.00$\pm$0.00}&\textbf{3600$\pm$   0}\\
  qsar-biodegradation&2&\textbf{0.80$\pm$0.01}&2&0.80$\pm$0.02&\textbf{0.04$\pm$0.08}&\textbf{2686$\pm$ 587}&0.80$\pm$0.02&0.95$\pm$0.06&3600$\pm$   0\\
  qsar-biodegradation&3&\textbf{0.83$\pm$0.01}&3&0.83$\pm$0.02&\textbf{1.00$\pm$0.00}&\textbf{3600$\pm$   0}&0.83$\pm$0.02&\textbf{1.00$\pm$0.00}&\textbf{3600$\pm$   0}\\
  qsar-biodegradation&4&0.85$\pm$0.02&4&\textbf{0.86$\pm$0.02}&\textbf{1.00$\pm$0.00}&\textbf{3600$\pm$   0}&0.83$\pm$0.03&\textbf{1.00$\pm$0.00}&\textbf{3600$\pm$   0}\\
  qsar-biodegradation&5&\textbf{0.87$\pm$0.03}&4&0.86$\pm$0.02&\textbf{1.00$\pm$0.00}&\textbf{3600$\pm$   0}&0.84$\pm$0.03&\textbf{1.00$\pm$0.00}&\textbf{3600$\pm$   0}\\
  banknote-authentication&2&0.55$\pm$0.17&5&0.90$\pm$0.01&\textbf{0.00$\pm$0.00}&\textbf{  19$\pm$   3}&\textbf{0.92$\pm$0.01}&\textbf{0.00$\pm$0.00}& 153$\pm$ 110\\
  banknote-authentication&3&0.66$\pm$0.29&3&0.95$\pm$0.00&\textbf{0.00$\pm$0.00}&\textbf{1655$\pm$ 352}&\textbf{0.98$\pm$0.01}&1.00$\pm$0.00&3600$\pm$   0\\
  banknote-authentication&4&0.65$\pm$0.29&5&0.97$\pm$0.00&\textbf{1.00$\pm$0.00}&\textbf{3600$\pm$   0}&\textbf{0.99$\pm$0.00}&\textbf{1.00$\pm$0.00}&\textbf{3600$\pm$   0}\\
  banknote-authentication&5&0.63$\pm$0.21&5&0.98$\pm$0.01&\textbf{1.00$\pm$0.00}&\textbf{3600$\pm$   0}&\textbf{0.99$\pm$0.00}&\textbf{1.00$\pm$0.00}&\textbf{3600$\pm$   0}\\
  ozone-level-detection-one&2&\textbf{0.97$\pm$0.01}&0&\textbf{0.97$\pm$0.01}&\textbf{0.98$\pm$0.03}&\textbf{3600$\pm$   0}&\textbf{0.97$\pm$0.01}&1.00$\pm$0.00&\textbf{3600$\pm$   0}\\
  ozone-level-detection-one&3&\textbf{0.98$\pm$0.01}&2&0.97$\pm$0.01&\textbf{1.00$\pm$0.00}&\textbf{3600$\pm$   0}&0.97$\pm$0.01&\textbf{1.00$\pm$0.00}&\textbf{3600$\pm$   0}\\
  ozone-level-detection-one&4&\textbf{0.97$\pm$0.01}&2&\textbf{0.97$\pm$0.01}&\textbf{1.00$\pm$0.00}&\textbf{3600$\pm$   0}&\textbf{0.97$\pm$0.01}&\textbf{1.00$\pm$0.00}&\textbf{3600$\pm$   0}\\
  ozone-level-detection-one&5&0.97$\pm$0.01&2&0.97$\pm$0.01&\textbf{1.00$\pm$0.00}&\textbf{3600$\pm$   0}&\textbf{0.97$\pm$0.00}&\textbf{1.00$\pm$0.00}&\textbf{3600$\pm$   0}\\
  image-segmentation&2&0.15$\pm$0.00&5&0.57$\pm$0.01&\textbf{0.08$\pm$0.18}&\textbf{1645$\pm$1172}&\textbf{0.58$\pm$0.01}&0.26$\pm$0.15&3292$\pm$ 689\\
  image-segmentation&3&0.14$\pm$0.01&5&\textbf{0.76$\pm$0.05}&\textbf{1.00$\pm$0.00}&\textbf{3600$\pm$   0}&0.52$\pm$0.05&\textbf{1.00$\pm$0.00}&\textbf{3600$\pm$   0}\\
  image-segmentation&4&0.14$\pm$0.01&5&\textbf{0.82$\pm$0.02}&\textbf{1.00$\pm$0.00}&\textbf{3600$\pm$   0}&0.57$\pm$0.04&\textbf{1.00$\pm$0.00}&3600$\pm$   1\\
  image-segmentation&5&0.15$\pm$0.00&5&\textbf{0.78$\pm$0.03}&\textbf{1.00$\pm$0.00}&\textbf{3600$\pm$   0}&0.76$\pm$0.13&\textbf{1.00$\pm$0.00}&\textbf{3600$\pm$   0}\\
  seismic-bumps&2&0.73$\pm$0.37&5&\textbf{0.94$\pm$0.01}&\textbf{0.00$\pm$0.00}&\textbf{ 789$\pm$ 441}&\textbf{0.94$\pm$0.01}&0.19$\pm$0.12&3331$\pm$ 603\\
  seismic-bumps&3&0.80$\pm$0.20&5&\textbf{0.94$\pm$0.01}&\textbf{1.00$\pm$0.00}&\textbf{3600$\pm$   0}&\textbf{0.94$\pm$0.01}&\textbf{1.00$\pm$0.00}&\textbf{3600$\pm$   0}\\
  seismic-bumps&4&0.92$\pm$0.04&5&\textbf{0.94$\pm$0.01}&\textbf{1.00$\pm$0.00}&\textbf{3600$\pm$   0}&\textbf{0.94$\pm$0.01}&\textbf{1.00$\pm$0.00}&\textbf{3600$\pm$   0}\\
  seismic-bumps&5&0.79$\pm$0.28&5&\textbf{0.94$\pm$0.01}&\textbf{1.00$\pm$0.00}&\textbf{3600$\pm$   0}&\textbf{0.94$\pm$0.01}&\textbf{1.00$\pm$0.00}&\textbf{3600$\pm$   0}\\
  thyroid-disease-ann-thyroid&2&\textbf{0.98$\pm$0.01}&0&0.92$\pm$0.00&\textbf{0.00$\pm$0.00}& 616$\pm$ 250&0.97$\pm$0.00&\textbf{0.00$\pm$0.00}&\textbf{ 349$\pm$ 116}\\
  thyroid-disease-ann-thyroid&3&\textbf{0.99$\pm$0.00}&0&0.93$\pm$0.00&\textbf{0.97$\pm$0.04}&\textbf{3600$\pm$   0}&0.98$\pm$0.00&1.00$\pm$0.00&\textbf{3600$\pm$   0}\\
  thyroid-disease-ann-thyroid&4&0.98$\pm$0.01&1&0.94$\pm$0.00&\textbf{1.00$\pm$0.00}&\textbf{3600$\pm$   0}&\textbf{0.99$\pm$0.00}&\textbf{1.00$\pm$0.00}&\textbf{3600$\pm$   0}\\
  thyroid-disease-ann-thyroid&5&0.98$\pm$0.01&1&0.94$\pm$0.01&\textbf{1.00$\pm$0.00}&\textbf{3600$\pm$   0}&\textbf{0.99$\pm$0.00}&\textbf{1.00$\pm$0.00}&\textbf{3600$\pm$   0}\\
  spambase&2&0.52$\pm$0.12&5&\textbf{0.86$\pm$0.01}&\textbf{0.12$\pm$0.26}&\textbf{3072$\pm$ 633}&0.83$\pm$0.01&0.89$\pm$0.02&3600$\pm$   0\\
  spambase&3&0.48$\pm$0.12&5&\textbf{0.86$\pm$0.02}&\textbf{0.98$\pm$0.01}&\textbf{3600$\pm$   0}&0.81$\pm$0.02&0.99$\pm$0.00&\textbf{3600$\pm$   0}\\
  spambase&4&0.44$\pm$0.09&5&\textbf{0.87$\pm$0.01}&\textbf{0.99$\pm$0.01}&\textbf{3600$\pm$   0}&0.83$\pm$0.03&1.00$\pm$0.00&\textbf{3600$\pm$   0}\\
  spambase&5&0.57$\pm$0.09&5&\textbf{0.86$\pm$0.03}&\textbf{0.99$\pm$0.01}&\textbf{3600$\pm$   1}&0.83$\pm$0.03&1.00$\pm$0.00&3601$\pm$   0\\
  wall-following-robot-2&2&\textbf{0.64$\pm$0.10}&0&0.63$\pm$0.06&\textbf{0.98$\pm$0.02}&\textbf{3600$\pm$   0}&0.60$\pm$0.09&1.00$\pm$0.00&3601$\pm$   1\\
  wall-following-robot-2&3&\textbf{0.68$\pm$0.05}&0&0.66$\pm$0.04&\textbf{1.00$\pm$0.00}&\textbf{3600$\pm$   0}&0.58$\pm$0.11&\textbf{1.00$\pm$0.00}&\textbf{3600$\pm$   0}\\
  wall-following-robot-2&4&\textbf{0.70$\pm$0.09}&2&0.63$\pm$0.12&\textbf{1.00$\pm$0.00}&\textbf{3600$\pm$   0}&0.56$\pm$0.03&\textbf{1.00$\pm$0.00}&\textbf{3600$\pm$   0}\\
  wall-following-robot-2&5&0.63$\pm$0.11&0&0.62$\pm$0.06&\textbf{1.00$\pm$0.00}&\textbf{3600$\pm$   0}&\textbf{0.64$\pm$0.11}&\textbf{1.00$\pm$0.00}&3602$\pm$   3\\ \hline

\end{tabular}%
}
}
% \end{sidewaystable*}
\end{table}
\end{landscape}

% \begin{sidewaystable*}[t]
% \begin{landscape}
\begin{table}[ht]
\center{
\OneAndAHalfSpacedXII
\caption{\revision{In-sample results including the average and standard deviation of optimality gap and solving time across 45 instances (5 samples and 9 values of $\lambda$) for the case of $\lambda > 0$ on mixed-feature datasets (part 1). The best performance achieved in a given dataset and depth is reported in \textbf{bold}.}}
\small{
\label{tab:in_sample_non_categorical_data_with_reg_part_1}
\setlength{\tabcolsep}{4pt}
% \resizebox{0.7\textheight}{!}{%
\scalebox{0.9}{
\color{black}\begin{tabular}{lc||cc|cc}
\hline
Dataset &
  Depth &
  \multicolumn{2}{c|}{\bendersFiveBuckets} &
  \multicolumn{2}{c}{\bendersTenBuckets}\\
 &  & Gap & Time  & Gap & Time \\
  \hline
  echocardiogram&2&\textbf{0.00$\pm$0.00}&\textbf{   1$\pm$   1}&\textbf{0.00$\pm$0.00}&\textbf{   1$\pm$   1}\\
  echocardiogram&3&\textbf{0.00$\pm$0.00}&   2$\pm$   2&\textbf{0.00$\pm$0.00}&\textbf{   1$\pm$   1}\\
  echocardiogram&4&\textbf{0.00$\pm$0.00}&\textbf{   3$\pm$   2}&\textbf{0.00$\pm$0.00}&   4$\pm$   4\\
  echocardiogram&5&\textbf{0.00$\pm$0.00}&   7$\pm$   5&\textbf{0.00$\pm$0.00}&\textbf{   5$\pm$   5}\\
  hepatitis&2&\textbf{0.00$\pm$0.00}&\textbf{   6$\pm$  10}&\textbf{0.00$\pm$0.00}&  14$\pm$  21\\
  hepatitis&3&0.00$\pm$0.03&\textbf{ 188$\pm$ 623}&\textbf{0.00$\pm$0.01}& 244$\pm$ 697\\
  hepatitis&4&\textbf{0.00$\pm$0.00}& 223$\pm$ 597&\textbf{0.00$\pm$0.00}&\textbf{ 150$\pm$ 364}\\
  hepatitis&5&0.00$\pm$0.02& 238$\pm$ 701&\textbf{0.00$\pm$0.00}&\textbf{ 148$\pm$ 271}\\
  fertility&2&\textbf{0.00$\pm$0.00}&\textbf{   4$\pm$   2}&\textbf{0.00$\pm$0.00}&   4$\pm$   3\\
  fertility&3&0.04$\pm$0.14&1179$\pm$1416&\textbf{0.04$\pm$0.13}&\textbf{1168$\pm$1385}\\
  fertility&4&\textbf{0.14$\pm$0.19}&\textbf{1756$\pm$1741}&0.15$\pm$0.20&1768$\pm$1737\\
  fertility&5&\textbf{0.17$\pm$0.22}&\textbf{1851$\pm$1755}&0.18$\pm$0.22&1859$\pm$1744\\
  iris&2&\textbf{0.00$\pm$0.00}&\textbf{   2$\pm$   0}&\textbf{0.00$\pm$0.00}&   3$\pm$   1\\
  iris&3&\textbf{0.00$\pm$0.00}&\textbf{  16$\pm$  22}&\textbf{0.00$\pm$0.00}& 107$\pm$ 222\\
  iris&4&\textbf{0.02$\pm$0.08}&\textbf{ 407$\pm$ 960}&0.04$\pm$0.12& 705$\pm$1321\\
  iris&5&0.07$\pm$0.16&\textbf{ 787$\pm$1435}&\textbf{0.07$\pm$0.14}& 825$\pm$1456\\
  wine&2&\textbf{0.00$\pm$0.00}&\textbf{  12$\pm$   4}&\textbf{0.00$\pm$0.00}&  49$\pm$  23\\
  wine&3&\textbf{0.01$\pm$0.04}&\textbf{ 565$\pm$ 797}&0.08$\pm$0.17& 997$\pm$1433\\
  wine&4&0.05$\pm$0.13&\textbf{1075$\pm$1189}&\textbf{0.03$\pm$0.13}&1223$\pm$1414\\
  wine&5&0.05$\pm$0.12&1401$\pm$1381&\textbf{0.05$\pm$0.10}&\textbf{1176$\pm$1394}\\
  planning-relax&2&\textbf{0.00$\pm$0.00}&\textbf{  80$\pm$  43}&\textbf{0.00$\pm$0.00}&1426$\pm$ 850\\
  planning-relax&3&\textbf{0.60$\pm$0.36}&\textbf{3003$\pm$1301}&0.67$\pm$0.33&3155$\pm$1158\\
  planning-relax&4&\textbf{0.62$\pm$0.34}&\textbf{3063$\pm$1267}&0.67$\pm$0.32&3189$\pm$1136\\
  planning-relax&5&\textbf{0.62$\pm$0.33}&\textbf{3066$\pm$1258}&0.68$\pm$0.32&3197$\pm$1103\\
  breast-cancer-prognostic&2&\textbf{0.00$\pm$0.00}&\textbf{1188$\pm$ 717}&0.53$\pm$0.35&3112$\pm$1147\\
  breast-cancer-prognostic&3&\textbf{0.62$\pm$0.32}&\textbf{3136$\pm$1196}&0.69$\pm$0.30&3220$\pm$1088\\
  breast-cancer-prognostic&4&\textbf{0.61$\pm$0.31}&\textbf{3167$\pm$1126}&0.68$\pm$0.29&3242$\pm$1024\\
  breast-cancer-prognostic&5&\textbf{0.63$\pm$0.31}&\textbf{3186$\pm$1074}&0.69$\pm$0.30&3284$\pm$ 912\\
  parkinsons&2&\textbf{0.00$\pm$0.00}&\textbf{  51$\pm$  26}&\textbf{0.00$\pm$0.00}& 442$\pm$ 295\\
  parkinsons&3&\textbf{0.34$\pm$0.29}&\textbf{2738$\pm$1472}&0.46$\pm$0.28&3079$\pm$1216\\
  parkinsons&4&\textbf{0.38$\pm$0.26}&\textbf{2837$\pm$1416}&0.52$\pm$0.27&3076$\pm$1239\\
  parkinsons&5&\textbf{0.38$\pm$0.25}&\textbf{2876$\pm$1360}&0.52$\pm$0.26&3116$\pm$1146\\
  connectionist-bench-sonar&2&\textbf{0.62$\pm$0.30}&\textbf{3391$\pm$ 626}&0.79$\pm$0.22&3600$\pm$   0\\
  connectionist-bench-sonar&3&\textbf{0.75$\pm$0.25}&\textbf{3600$\pm$   0}&0.80$\pm$0.17&\textbf{3600$\pm$   0}\\
  connectionist-bench-sonar&4&\textbf{0.75$\pm$0.22}&\textbf{3600$\pm$   0}&0.80$\pm$0.17&\textbf{3600$\pm$   0}\\
  connectionist-bench-sonar&5&\textbf{0.75$\pm$0.24}&\textbf{3600$\pm$   0}&0.81$\pm$0.17&\textbf{3600$\pm$   0}\\
  seeds&2&\textbf{0.00$\pm$0.00}&\textbf{   5$\pm$   1}&\textbf{0.00$\pm$0.00}&  12$\pm$   4\\
  seeds&3&\textbf{0.01$\pm$0.05}&\textbf{1010$\pm$1271}&0.15$\pm$0.21&2109$\pm$1589\\
  seeds&4&\textbf{0.35$\pm$0.30}&\textbf{2610$\pm$1533}&0.37$\pm$0.27&2838$\pm$1445\\
  seeds&5&0.41$\pm$0.31&\textbf{2731$\pm$1474}&\textbf{0.38$\pm$0.28}&2823$\pm$1454\\
  cylinder-bands&2&\textbf{0.75$\pm$0.23}&\textbf{3548$\pm$ 352}&0.86$\pm$0.15&3604$\pm$  27\\
  cylinder-bands&3&\textbf{0.87$\pm$0.14}&\textbf{3600$\pm$   0}&0.89$\pm$0.11&\textbf{3600$\pm$   0}\\
  cylinder-bands&4&\textbf{0.86$\pm$0.14}&\textbf{3600$\pm$   0}&0.87$\pm$0.14&3600$\pm$   1\\
  cylinder-bands&5&\textbf{0.86$\pm$0.15}&\textbf{3600$\pm$   0}&0.88$\pm$0.12&3670$\pm$ 445\\
  heart-cleveland&2&\textbf{0.00$\pm$0.00}&\textbf{ 109$\pm$  33}&\textbf{0.00$\pm$0.00}& 437$\pm$ 141\\
  heart-cleveland&3&\textbf{0.77$\pm$0.24}&\textbf{3600$\pm$   0}&0.82$\pm$0.20&\textbf{3600$\pm$   0}\\
  heart-cleveland&4&\textbf{0.84$\pm$0.19}&\textbf{3600$\pm$   0}&0.85$\pm$0.18&\textbf{3600$\pm$   0}\\
  heart-cleveland&5&\textbf{0.85$\pm$0.18}&\textbf{3600$\pm$   0}&0.86$\pm$0.17&\textbf{3600$\pm$   0}\\
  ionosphere&2&\textbf{0.00$\pm$0.00}&\textbf{ 488$\pm$ 305}&0.32$\pm$0.37&3031$\pm$ 821\\
  ionosphere&3&\textbf{0.69$\pm$0.30}&\textbf{3276$\pm$ 931}&0.73$\pm$0.24&3600$\pm$   0\\
  ionosphere&4&\textbf{0.71$\pm$0.29}&\textbf{3354$\pm$ 742}&0.73$\pm$0.22&3600$\pm$   0\\
  ionosphere&5&\textbf{0.71$\pm$0.29}&\textbf{3430$\pm$ 607}&0.73$\pm$0.22&3600$\pm$   0\\
  thoracic-surgery&2&\textbf{0.00$\pm$0.00}&\textbf{  69$\pm$  40}&\textbf{0.00$\pm$0.00}& 167$\pm$  78\\
  thoracic-surgery&3&\textbf{0.70$\pm$0.31}&\textbf{3213$\pm$1107}&0.74$\pm$0.30&3221$\pm$1083\\
  thoracic-surgery&4&\textbf{0.74$\pm$0.30}&\textbf{3218$\pm$1092}&0.75$\pm$0.30&3243$\pm$1032\\
  thoracic-surgery&5&\textbf{0.75$\pm$0.30}&\textbf{3227$\pm$1070}&0.78$\pm$0.30&3242$\pm$1030\\ \hline
\end{tabular}%
}
}
}
% \end{sidewaystable*}
\end{table}
% \end{landscape}

% \begin{sidewaystable*}[t]
% \begin{landscape}
\begin{table}[ht]
\center{
\OneAndAHalfSpacedXII
\caption{  
\revision{In-sample results including the average and standard deviation of optimality gap and solving time across 45 instances (5 samples and 9 values of $\lambda$) for the case of $\lambda > 0$ on mixed-feature datasets (part 2). The best performance achieved in a given dataset and depth is reported in \textbf{bold}.}}
\small{
\label{tab:in_sample_non_categorical_data_with_reg_part_2}
\setlength{\tabcolsep}{4pt}
% \resizebox{0.7\textheight}{!}{%
\scalebox{0.9}{
\color{black}\begin{tabular}{lc||cc|cc}
\hline
Dataset &
  Depth &
  \multicolumn{2}{c|}{\bendersFiveBuckets} &
  \multicolumn{2}{c}{\bendersTenBuckets}\\
 &  & Gap & Time  & Gap & Time \\
  \hline
  climate&2&\textbf{0.00$\pm$0.00}&\textbf{ 563$\pm$ 323}&0.63$\pm$0.33&3163$\pm$1083\\
  climate&3&\textbf{0.66$\pm$0.33}&\textbf{3162$\pm$1145}&0.72$\pm$0.30&3225$\pm$1074\\
  climate&4&\textbf{0.69$\pm$0.32}&\textbf{3236$\pm$1044}&0.71$\pm$0.30&3258$\pm$ 988\\
  climate&5&\textbf{0.69$\pm$0.31}&\textbf{3208$\pm$1120}&0.72$\pm$0.30&3226$\pm$1072\\
  breast-cancer-diagnostic&2&\textbf{0.00$\pm$0.00}&\textbf{ 405$\pm$ 208}&0.09$\pm$0.22&2602$\pm$ 927\\
  breast-cancer-diagnostic&3&\textbf{0.67$\pm$0.30}&\textbf{3293$\pm$ 884}&0.70$\pm$0.26&3530$\pm$ 470\\
  breast-cancer-diagnostic&4&\textbf{0.70$\pm$0.28}&\textbf{3310$\pm$ 841}&0.73$\pm$0.26&3526$\pm$ 496\\
  breast-cancer-diagnostic&5&\textbf{0.71$\pm$0.28}&\textbf{3379$\pm$ 702}&0.72$\pm$0.27&3528$\pm$ 485\\
  indian-liver-patient&2&\textbf{0.00$\pm$0.00}&\textbf{ 111$\pm$  32}&\textbf{0.00$\pm$0.00}&1940$\pm$ 635\\
  indian-liver-patient&3&\textbf{0.82$\pm$0.19}&\textbf{3600$\pm$   0}&0.89$\pm$0.13&\textbf{3600$\pm$   0}\\
  indian-liver-patient&4&\textbf{0.86$\pm$0.17}&\textbf{3600$\pm$   0}&0.91$\pm$0.11&\textbf{3600$\pm$   0}\\
  indian-liver-patient&5&\textbf{0.88$\pm$0.15}&\textbf{3600$\pm$   0}&0.91$\pm$0.11&\textbf{3600$\pm$   0}\\
  credit-approval&2&\textbf{0.00$\pm$0.00}&\textbf{ 296$\pm$ 101}&\textbf{0.00$\pm$0.00}&1298$\pm$ 587\\
  credit-approval&3&\textbf{0.83$\pm$0.21}&\textbf{3600$\pm$   0}&0.86$\pm$0.17&\textbf{3600$\pm$   0}\\
  credit-approval&4&\textbf{0.86$\pm$0.16}&\textbf{3600$\pm$   0}&0.87$\pm$0.15&\textbf{3600$\pm$   0}\\
  credit-approval&5&\textbf{0.87$\pm$0.15}&\textbf{3600$\pm$   0}&0.88$\pm$0.13&\textbf{3600$\pm$   0}\\
  blood-transfusion&2&\textbf{0.00$\pm$0.00}&\textbf{  22$\pm$   5}&\textbf{0.00$\pm$0.00}&  96$\pm$  44\\
  blood-transfusion&3&\textbf{0.52$\pm$0.22}&\textbf{3342$\pm$ 743}&0.67$\pm$0.16&3600$\pm$   0\\
  blood-transfusion&4&\textbf{0.75$\pm$0.18}&\textbf{3600$\pm$   0}&0.78$\pm$0.16&\textbf{3600$\pm$   0}\\
  blood-transfusion&5&\textbf{0.80$\pm$0.17}&\textbf{3600$\pm$   0}&0.82$\pm$0.15&\textbf{3600$\pm$   0}\\
  diabetes&2&\textbf{0.00$\pm$0.00}&\textbf{ 123$\pm$  56}&\textbf{0.00$\pm$0.00}&1305$\pm$ 310\\
  diabetes&3&\textbf{0.86$\pm$0.14}&\textbf{3600$\pm$   0}&0.91$\pm$0.10&\textbf{3600$\pm$   0}\\
  diabetes&4&\textbf{0.90$\pm$0.12}&\textbf{3600$\pm$   0}&0.92$\pm$0.09&\textbf{3600$\pm$   0}\\
  diabetes&5&\textbf{0.91$\pm$0.11}&\textbf{3600$\pm$   0}&0.93$\pm$0.09&\textbf{3600$\pm$   0}\\
  qsar-biodegradation&2&\textbf{0.17$\pm$0.21}&\textbf{3345$\pm$ 392}&0.92$\pm$0.08&3600$\pm$   0\\
  qsar-biodegradation&3&\textbf{0.93$\pm$0.07}&\textbf{3600$\pm$   0}&0.95$\pm$0.06&\textbf{3600$\pm$   0}\\
  qsar-biodegradation&4&\textbf{0.94$\pm$0.07}&\textbf{3600$\pm$   0}&0.95$\pm$0.06&\textbf{3600$\pm$   0}\\
  qsar-biodegradation&5&\textbf{0.95$\pm$0.06}&\textbf{3600$\pm$   0}&\textbf{0.95$\pm$0.06}&\textbf{3600$\pm$   0}\\
  banknote-authentication&2&\textbf{0.00$\pm$0.00}&\textbf{  21$\pm$   4}&\textbf{0.00$\pm$0.00}& 163$\pm$  55\\
  banknote-authentication&3&\textbf{0.11$\pm$0.23}&\textbf{2430$\pm$1020}&0.71$\pm$0.23&3600$\pm$   0\\
  banknote-authentication&4&\textbf{0.75$\pm$0.20}&\textbf{3600$\pm$   0}&0.77$\pm$0.16&\textbf{3600$\pm$   0}\\
  banknote-authentication&5&\textbf{0.76$\pm$0.18}&\textbf{3600$\pm$   0}&0.80$\pm$0.15&\textbf{3600$\pm$   0}\\
  ozone-level-detection-one&2&\textbf{0.78$\pm$0.28}&\textbf{3372$\pm$ 752}&0.83$\pm$0.25&3553$\pm$ 317\\
  ozone-level-detection-one&3&\textbf{0.82$\pm$0.25}&\textbf{3527$\pm$ 493}&0.86$\pm$0.19&3583$\pm$ 115\\
  ozone-level-detection-one&4&\textbf{0.85$\pm$0.21}&\textbf{3568$\pm$ 218}&\textbf{0.85$\pm$0.21}&3569$\pm$ 210\\
  ozone-level-detection-one&5&\textbf{0.84$\pm$0.23}&\textbf{3538$\pm$ 417}&0.87$\pm$0.16&3611$\pm$  37\\
  image-segmentation&2&\textbf{0.20$\pm$0.16}&\textbf{3543$\pm$ 171}&0.52$\pm$0.17&3572$\pm$ 187\\
  image-segmentation&3&\textbf{0.97$\pm$0.03}&\textbf{3600$\pm$   0}&0.99$\pm$0.01&\textbf{3600$\pm$   0}\\
  image-segmentation&4&\textbf{0.98$\pm$0.02}&\textbf{3600$\pm$   0}&0.99$\pm$0.01&\textbf{3600$\pm$   0}\\
  image-segmentation&5&\textbf{0.98$\pm$0.02}&\textbf{3600$\pm$   0}&0.99$\pm$0.02&\textbf{3600$\pm$   0}\\
  seismic-bumps&2&\textbf{0.00$\pm$0.00}&\textbf{1040$\pm$ 414}&0.16$\pm$0.17&3209$\pm$ 700\\
  seismic-bumps&3&\textbf{0.88$\pm$0.14}&\textbf{3600$\pm$   0}&0.90$\pm$0.12&\textbf{3600$\pm$   0}\\
  seismic-bumps&4&\textbf{0.90$\pm$0.13}&\textbf{3600$\pm$   0}&0.91$\pm$0.11&\textbf{3600$\pm$   0}\\
  seismic-bumps&5&\textbf{0.91$\pm$0.12}&\textbf{3600$\pm$   0}&0.92$\pm$0.10&\textbf{3600$\pm$   0}\\
  thyroid-disease-ann-thyroid&2&\textbf{0.00$\pm$0.00}&\textbf{ 320$\pm$  98}&\textbf{0.00$\pm$0.00}& 617$\pm$ 322\\
  thyroid-disease-ann-thyroid&3&0.88$\pm$0.13&\textbf{3600$\pm$   0}&\textbf{0.85$\pm$0.16}&\textbf{3600$\pm$   0}\\
  thyroid-disease-ann-thyroid&4&0.93$\pm$0.08&\textbf{3600$\pm$   0}&\textbf{0.87$\pm$0.13}&\textbf{3600$\pm$   0}\\
  thyroid-disease-ann-thyroid&5&0.94$\pm$0.07&\textbf{3600$\pm$   0}&\textbf{0.89$\pm$0.11}&\textbf{3600$\pm$   0}\\
  spambase&2&\textbf{0.23$\pm$0.25}&\textbf{3177$\pm$ 563}&0.89$\pm$0.04&3600$\pm$   0\\
  spambase&3&\textbf{0.96$\pm$0.03}&\textbf{3600$\pm$   0}&0.98$\pm$0.02&\textbf{3600$\pm$   0}\\
  spambase&4&\textbf{0.96$\pm$0.03}&\textbf{3600$\pm$   0}&0.98$\pm$0.01&\textbf{3600$\pm$   0}\\
  spambase&5&\textbf{0.97$\pm$0.02}&\textbf{3600$\pm$   1}&0.98$\pm$0.02&3601$\pm$   1\\
  wall-following-robot-2&2&\textbf{0.95$\pm$0.05}&\textbf{3600$\pm$   0}&0.99$\pm$0.01&\textbf{3600$\pm$   0}\\
  wall-following-robot-2&3&\textbf{0.99$\pm$0.01}&\textbf{3600$\pm$   0}&1.00$\pm$0.00&3601$\pm$   2\\
  wall-following-robot-2&4&\textbf{0.99$\pm$0.01}&\textbf{3600$\pm$   1}&1.00$\pm$0.00&3601$\pm$   1\\
  wall-following-robot-2&5&1.00$\pm$0.01&\textbf{3600$\pm$   1}&\textbf{1.00$\pm$0.00}&\textbf{3600$\pm$   1}\\ \hline
\end{tabular}%
}
}
}
% \end{sidewaystable*}
\end{table}
% \end{landscape}

\begin{table}[!ht]
\OneAndAHalfSpacedXII
\begin{center}
\caption{\revision{Average out-of-sample accuracy and standard deviation of accuracy across 5 samples on mixed-feature datasets (part 1).  The highest accuracy achieved in a given dataset and depth is reported in \textbf{bold}.}}
\label{tab:out_of_sample_non_categorical_data_part_1}
\small{
\setlength{\tabcolsep}{3pt}
\color{black}\begin{tabular}{l||c|c|c|c}
\hline
dataset&depth&\texttt{OCT}&\bendersFiveBuckets{}&\bendersTenBuckets{}\\
\hline
echocardiogram&2&0.92$\pm$0.07&\textbf{0.95$\pm$0.08}&0.91$\pm$0.06\\
echocardiogram&3&0.92$\pm$0.07&\textbf{0.95$\pm$0.08}&0.91$\pm$0.06\\
echocardiogram&4&0.92$\pm$0.07&\textbf{0.95$\pm$0.08}&0.94$\pm$0.00\\
echocardiogram&5&\textbf{0.96$\pm$0.03}&0.95$\pm$0.08&0.91$\pm$0.07\\
hepatitis&2&0.75$\pm$0.06&\textbf{0.77$\pm$0.10}&0.76$\pm$0.10\\
hepatitis&3&0.75$\pm$0.09&0.75$\pm$0.10&\textbf{0.77$\pm$0.09}\\
hepatitis&4&0.75$\pm$0.06&0.77$\pm$0.08&\textbf{0.81$\pm$0.11}\\
hepatitis&5&0.72$\pm$0.10&0.76$\pm$0.11&\textbf{0.78$\pm$0.12}\\
fertility&2&\textbf{0.90$\pm$0.06}&\textbf{0.90$\pm$0.06}&\textbf{0.90$\pm$0.06}\\
fertility&3&\textbf{0.90$\pm$0.06}&\textbf{0.90$\pm$0.06}&\textbf{0.90$\pm$0.06}\\
fertility&4&\textbf{0.90$\pm$0.06}&\textbf{0.90$\pm$0.06}&\textbf{0.90$\pm$0.06}\\
fertility&5&\textbf{0.90$\pm$0.06}&\textbf{0.90$\pm$0.06}&\textbf{0.90$\pm$0.06}\\
iris&2&\textbf{0.94$\pm$0.03}&0.91$\pm$0.05&0.90$\pm$0.05\\
iris&3&\textbf{0.94$\pm$0.03}&0.89$\pm$0.05&0.88$\pm$0.05\\
iris&4&\textbf{0.95$\pm$0.03}&0.88$\pm$0.04&0.91$\pm$0.05\\
iris&5&\textbf{0.94$\pm$0.03}&0.88$\pm$0.04&0.91$\pm$0.05\\
wine&2&\textbf{0.94$\pm$0.03}&0.90$\pm$0.07&0.86$\pm$0.05\\
wine&3&\textbf{0.95$\pm$0.04}&0.93$\pm$0.02&0.91$\pm$0.07\\
wine&4&0.90$\pm$0.07&0.91$\pm$0.07&\textbf{0.92$\pm$0.05}\\
wine&5&\textbf{0.92$\pm$0.07}&0.91$\pm$0.08&0.90$\pm$0.06\\
planning-relax&2&0.64$\pm$0.07&\textbf{0.67$\pm$0.07}&0.64$\pm$0.08\\
planning-relax&3&0.43$\pm$0.17&\textbf{0.67$\pm$0.07}&0.64$\pm$0.08\\
planning-relax&4&0.63$\pm$0.13&0.63$\pm$0.13&\textbf{0.64$\pm$0.07}\\
planning-relax&5&0.54$\pm$0.19&\textbf{0.67$\pm$0.07}&0.62$\pm$0.10\\
breast-cancer-prognostic&2&0.73$\pm$0.05&\textbf{0.75$\pm$0.07}&\textbf{0.75$\pm$0.07}\\
breast-cancer-prognostic&3&0.73$\pm$0.05&0.72$\pm$0.04&\textbf{0.74$\pm$0.06}\\
breast-cancer-prognostic&4&0.74$\pm$0.06&0.73$\pm$0.07&\textbf{0.75$\pm$0.06}\\
breast-cancer-prognostic&5&0.70$\pm$0.06&\textbf{0.76$\pm$0.06}&0.72$\pm$0.06\\
parkinsons&2&0.82$\pm$0.05&\textbf{0.87$\pm$0.02}&0.83$\pm$0.03\\
parkinsons&3&\textbf{0.84$\pm$0.08}&0.84$\pm$0.02&0.84$\pm$0.05\\
parkinsons&4&0.87$\pm$0.04&0.84$\pm$0.02&\textbf{0.89$\pm$0.05}\\
parkinsons&5&0.79$\pm$0.04&\textbf{0.87$\pm$0.03}&0.86$\pm$0.04\\
connectionist-bench-sonar&2&0.75$\pm$0.10&\textbf{0.76$\pm$0.06}&0.70$\pm$0.06\\
connectionist-bench-sonar&3&0.67$\pm$0.07&0.67$\pm$0.05&\textbf{0.72$\pm$0.06}\\
connectionist-bench-sonar&4&\textbf{0.76$\pm$0.03}&0.74$\pm$0.10&0.75$\pm$0.06\\
connectionist-bench-sonar&5&\textbf{0.74$\pm$0.02}&0.72$\pm$0.06&0.73$\pm$0.06\\
seeds&2&0.88$\pm$0.02&\textbf{0.89$\pm$0.04}&0.88$\pm$0.03\\
seeds&3&0.88$\pm$0.02&\textbf{0.89$\pm$0.04}&0.88$\pm$0.03\\
seeds&4&0.89$\pm$0.03&\textbf{0.89$\pm$0.04}&0.89$\pm$0.03\\
seeds&5&\textbf{0.91$\pm$0.04}&0.88$\pm$0.02&0.90$\pm$0.02\\
cylinder-bands&2&0.65$\pm$0.06&\textbf{0.72$\pm$0.04}&0.64$\pm$0.06\\
cylinder-bands&3&0.68$\pm$0.04&0.67$\pm$0.04&\textbf{0.69$\pm$0.03}\\
cylinder-bands&4&0.65$\pm$0.03&\textbf{0.74$\pm$0.05}&0.70$\pm$0.03\\
cylinder-bands&5&0.61$\pm$0.14&0.68$\pm$0.08&\textbf{0.70$\pm$0.04}\\
heart-cleveland&2&0.53$\pm$0.03&\textbf{0.54$\pm$0.01}&\textbf{0.54$\pm$0.01}\\
heart-cleveland&3&\textbf{0.55$\pm$0.04}&0.52$\pm$0.04&0.52$\pm$0.03\\
heart-cleveland&4&0.56$\pm$0.04&\textbf{0.59$\pm$0.04}&0.55$\pm$0.07\\
heart-cleveland&5&0.53$\pm$0.02&\textbf{0.55$\pm$0.04}&0.53$\pm$0.05\\
ionosphere&2&\textbf{0.88$\pm$0.09}&0.88$\pm$0.06&0.87$\pm$0.03\\
ionosphere&3&0.84$\pm$0.06&0.90$\pm$0.02&\textbf{0.90$\pm$0.03}\\
ionosphere&4&0.86$\pm$0.11&\textbf{0.90$\pm$0.02}&\textbf{0.90$\pm$0.02}\\
ionosphere&5&0.70$\pm$0.25&\textbf{0.89$\pm$0.03}&0.87$\pm$0.05\\
thoracic-surgery&2&\textbf{0.84$\pm$0.02}&\textbf{0.84$\pm$0.02}&\textbf{0.84$\pm$0.02}\\
thoracic-surgery&3&\textbf{0.84$\pm$0.02}&\textbf{0.84$\pm$0.02}&\textbf{0.84$\pm$0.02}\\
thoracic-surgery&4&\textbf{0.84$\pm$0.02}&\textbf{0.84$\pm$0.02}&\textbf{0.84$\pm$0.02}\\
thoracic-surgery&5&0.84$\pm$0.02&\textbf{0.84$\pm$0.03}&0.84$\pm$0.02\\
\hline
\end{tabular}
}
\end{center}
\end{table}

\begin{table}[!ht]
\OneAndAHalfSpacedXII
\begin{center}
\caption{\revision{Average out-of-sample accuracy and standard deviation of accuracy across 5 samples on mixed-feature datasets (part 2).  The highest accuracy achieved in a given dataset and depth is reported in \textbf{bold}.}}
\label{tab:out_of_sample_non_categorical_data_part_2}
\small{
\setlength{\tabcolsep}{3pt}
\color{black}\begin{tabular}{l||c|c|c|c}
\hline
dataset&depth&\texttt{OCT}&\bendersFiveBuckets{}&\bendersTenBuckets{}\\
\hline
climate&2&0.71$\pm$0.36&\textbf{0.92$\pm$0.03}&\textbf{0.92$\pm$0.03}\\
climate&3&0.57$\pm$0.46&\textbf{0.93$\pm$0.02}&0.92$\pm$0.02\\
climate&4&0.72$\pm$0.38&\textbf{0.92$\pm$0.02}&0.91$\pm$0.02\\
climate&5&0.77$\pm$0.36&0.92$\pm$0.02&\textbf{0.93$\pm$0.03}\\
breast-cancer-diagnostic&2&0.93$\pm$0.03&0.93$\pm$0.03&\textbf{0.95$\pm$0.02}\\
breast-cancer-diagnostic&3&\textbf{0.95$\pm$0.03}&0.93$\pm$0.02&0.94$\pm$0.02\\
breast-cancer-diagnostic&4&0.93$\pm$0.01&0.93$\pm$0.03&\textbf{0.94$\pm$0.02}\\
breast-cancer-diagnostic&5&0.91$\pm$0.03&0.93$\pm$0.02&\textbf{0.95$\pm$0.02}\\
indian-liver-patient&2&0.73$\pm$0.02&\textbf{0.73$\pm$0.03}&\textbf{0.73$\pm$0.03}\\
indian-liver-patient&3&0.72$\pm$0.03&\textbf{0.72$\pm$0.05}&0.71$\pm$0.03\\
indian-liver-patient&4&\textbf{0.72$\pm$0.05}&0.69$\pm$0.05&0.71$\pm$0.04\\
indian-liver-patient&5&\textbf{0.73$\pm$0.02}&0.71$\pm$0.04&0.70$\pm$0.05\\
credit-approval&2&\textbf{0.86$\pm$0.03}&\textbf{0.86$\pm$0.03}&\textbf{0.86$\pm$0.03}\\
credit-approval&3&0.81$\pm$0.12&\textbf{0.86$\pm$0.04}&0.86$\pm$0.03\\
credit-approval&4&0.79$\pm$0.09&\textbf{0.86$\pm$0.03}&\textbf{0.86$\pm$0.03}\\
credit-approval&5&0.69$\pm$0.18&\textbf{0.86$\pm$0.03}&\textbf{0.86$\pm$0.03}\\
blood-transfusion&2&\textbf{0.77$\pm$0.01}&0.76$\pm$0.03&\textbf{0.77$\pm$0.01}\\
blood-transfusion&3&\textbf{0.79$\pm$0.02}&0.77$\pm$0.02&0.78$\pm$0.01\\
blood-transfusion&4&0.77$\pm$0.01&\textbf{0.78$\pm$0.02}&\textbf{0.78$\pm$0.02}\\
blood-transfusion&5&\textbf{0.78$\pm$0.02}&\textbf{0.78$\pm$0.02}&0.76$\pm$0.03\\
diabetes&2&\textbf{0.75$\pm$0.03}&0.74$\pm$0.02&0.74$\pm$0.02\\
diabetes&3&\textbf{0.76$\pm$0.02}&0.73$\pm$0.02&0.75$\pm$0.02\\
diabetes&4&\textbf{0.75$\pm$0.02}&0.74$\pm$0.03&0.73$\pm$0.02\\
diabetes&5&\textbf{0.75$\pm$0.03}&0.74$\pm$0.03&0.73$\pm$0.02\\
qsar-biodegradation&2&0.76$\pm$0.02&\textbf{0.78$\pm$0.02}&\textbf{0.78$\pm$0.02}\\
qsar-biodegradation&3&0.77$\pm$0.04&\textbf{0.81$\pm$0.02}&0.77$\pm$0.03\\
qsar-biodegradation&4&0.74$\pm$0.04&0.76$\pm$0.02&\textbf{0.77$\pm$0.02}\\
qsar-biodegradation&5&0.72$\pm$0.07&\textbf{0.76$\pm$0.03}&0.75$\pm$0.05\\
banknote-authentication&2&0.50$\pm$0.07&0.89$\pm$0.02&\textbf{0.91$\pm$0.01}\\
banknote-authentication&3&0.50$\pm$0.07&0.93$\pm$0.00&\textbf{0.97$\pm$0.01}\\
banknote-authentication&4&0.54$\pm$0.05&\textbf{0.97$\pm$0.01}&\textbf{0.97$\pm$0.01}\\
banknote-authentication&5&0.48$\pm$0.07&\textbf{0.97$\pm$0.01}&\textbf{0.97$\pm$0.01}\\
ozone-level-detection-one&2&\textbf{0.97$\pm$0.00}&\textbf{0.97$\pm$0.00}&\textbf{0.97$\pm$0.00}\\
ozone-level-detection-one&3&\textbf{0.97$\pm$0.00}&\textbf{0.97$\pm$0.00}&\textbf{0.97$\pm$0.00}\\
ozone-level-detection-one&4&\textbf{0.97$\pm$0.00}&\textbf{0.97$\pm$0.00}&\textbf{0.97$\pm$0.00}\\
ozone-level-detection-one&5&0.97$\pm$0.00&\textbf{0.97$\pm$0.01}&0.97$\pm$0.00\\
image-segmentation&2&0.14$\pm$0.00&\textbf{0.55$\pm$0.02}&0.43$\pm$0.06\\
image-segmentation&3&0.14$\pm$0.01&\textbf{0.52$\pm$0.03}&0.51$\pm$0.03\\
image-segmentation&4&0.14$\pm$0.01&\textbf{0.64$\pm$0.10}&0.49$\pm$0.09\\
image-segmentation&5&0.14$\pm$0.01&0.65$\pm$0.13&\textbf{0.66$\pm$0.08}\\
seismic-bumps&2&0.76$\pm$0.39&\textbf{0.93$\pm$0.01}&\textbf{0.93$\pm$0.01}\\
seismic-bumps&3&0.81$\pm$0.26&\textbf{0.93$\pm$0.01}&\textbf{0.93$\pm$0.01}\\
seismic-bumps&4&\textbf{0.93$\pm$0.01}&\textbf{0.93$\pm$0.01}&\textbf{0.93$\pm$0.01}\\
seismic-bumps&5&0.77$\pm$0.37&\textbf{0.93$\pm$0.01}&\textbf{0.93$\pm$0.01}\\
thyroid-disease-ann-thyroid&2&\textbf{0.96$\pm$0.01}&0.93$\pm$0.00&0.96$\pm$0.00\\
thyroid-disease-ann-thyroid&3&0.96$\pm$0.02&0.94$\pm$0.00&\textbf{0.97$\pm$0.01}\\
thyroid-disease-ann-thyroid&4&0.96$\pm$0.02&0.94$\pm$0.01&\textbf{0.98$\pm$0.01}\\
thyroid-disease-ann-thyroid&5&0.96$\pm$0.01&0.94$\pm$0.01&\textbf{0.97$\pm$0.01}\\
spambase&2&0.47$\pm$0.11&\textbf{0.85$\pm$0.01}&0.81$\pm$0.02\\
spambase&3&0.47$\pm$0.11&\textbf{0.83$\pm$0.03}&0.82$\pm$0.02\\
spambase&4&0.51$\pm$0.11&\textbf{0.82$\pm$0.02}&0.81$\pm$0.03\\
spambase&5&0.60$\pm$0.01&\textbf{0.80$\pm$0.06}&0.75$\pm$0.08\\
wall-following-robot-2&2&0.55$\pm$0.07&\textbf{0.64$\pm$0.06}&0.58$\pm$0.12\\
wall-following-robot-2&3&0.61$\pm$0.06&\textbf{0.61$\pm$0.11}&0.59$\pm$0.10\\
wall-following-robot-2&4&\textbf{0.63$\pm$0.05}&0.57$\pm$0.09&0.54$\pm$0.10\\
wall-following-robot-2&5&0.51$\pm$0.8&\textbf{0.54$\pm$0.05}&0.49$\pm$0.09\\
\hline
\end{tabular}
}
\end{center}
\end{table}

}

%%%%%%%%%%%%%%%%%%%%%%%%%%%%%%%%%%%%%%%%%%%%%%%%%%
%%%%%%%%%%%%%%%%%%%%%%%%%%%%%%%%%%%%%%%%%%%%%%%%%%
\section{Additional Results}
\label{appendix_sec:additional_results}
%%%%%%%%%%%%%%%%%%%%%%%%%%%%%%%%%%%%%%%%%%%%%%%%%%
%%%%%%%%%%%%%%%%%%%%%%%%%%%%%%%%%%%%%%%%%%%%%%%%%%

\revision{In this section, we report and discuss additional numerical experiments conducted on the categorical datasets.}

\subsection{\benders{}'s Variants}
\label{appendix_sec:benders_variants}

\revision{In this section, we evaluate three implementation variants of \benders{} that differ in the strategy for adding the cut-set inequalities. Recall from \revision{Remark~\ref{remark:benders_implementation},} that in the implementation of \benders{}, we first add as many cut-set inequalities as possible at the root node of the branch-and-bound tree by using Gurobi to solve each subproblem. After that, we only separate the cut-set inequalities at the integral solutions by solving the min-cut subproblems using Algorithm~\ref{alg:cut_reg}. The first variant,~\bendersMIPSolAlg{}, is similar to \benders{} with the only difference being that we no longer add the cut-set inequalities at the root node of the branch-and-bound tree.
In the second variant,~\bendersMIPSolLP{}, we again separate the cut-set inequalities only at the integral solutions but instead of using Algorithm~\ref{alg:cut_reg} to solve the min-cut subproblems, we use Gurobi to solve \revision{them}. In the third variant,~\bendersAllSolLP{}, we separate the cut-set inequalities at all solutions including both fractional and integral ones by solving \revision{each subproblem using Gurobi}. 
Figure~\ref{fig:performace-Benders_All} summarizes the in-sample performance of all four methods. 

From the left part (time axis) of Figure~\ref{fig:performace-Benders_All}, we observe that \bendersAllSolLP{} can solve $1159$ instances (out of 2400) to optimality within the time limit.~\bendersMIPSolLP{} can solve the same number of instances in $110$ seconds, resulting in a $\floor{\frac{3600}{110}}=33\times$ speedup. From this observation, we conclude that it is better to separate the cut-set inequalities only at the integral solutions.~\bendersMIPSolAlg{} solves the same number of instances in $53$ seconds, which means that using the separation procedure described in Algorithm~\ref{alg:cut_reg}, instead of solving the corresponding LO, results in a $\floor{\frac{110}{53}}=2\times$ speedup. However, \benders{} slightly outperforms~\bendersMIPSolAlg{} as it can solves more instances (1536 vs 1522) within the time limit by adding some extra cuts at the root node of the branch-and-bound search tree. 
The right part (gap axis) of Figure~\ref{fig:performace-Benders_All} summarizes the optimality gap of each approach at the time limit. As we can see, \benders{},~\bendersMIPSolAlg{}, and \bendersMIPSolLP{} have a similar performance and all outperform \bendersAllSolLP.
\begin{figure}[t!]
%\vskip 0.2in
\begin{center}
\includegraphics[width = 0.55\textwidth]{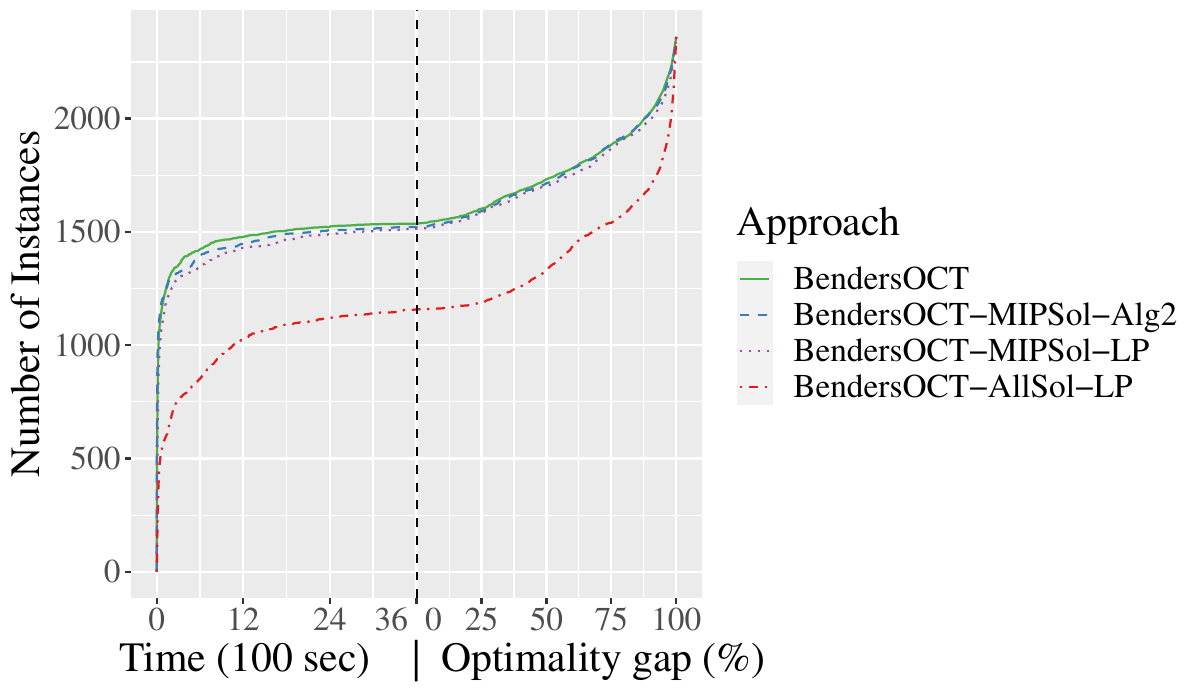}
\caption{Number of instances solved to optimality by each approach within the given time on the time axis, and number of instances with optimality gap no larger than each given value at the time limit on the optimality gap axis.}
\label{fig:performace-Benders_All}
\end{center}
\end{figure}
}

\subsection{Worst-case Accuracy}
\label{appendix_sec:flow_worst}

\begin{figure}[t!]
%\vskip 0.2in
\begin{center}
\includegraphics[width = 0.75\textwidth]{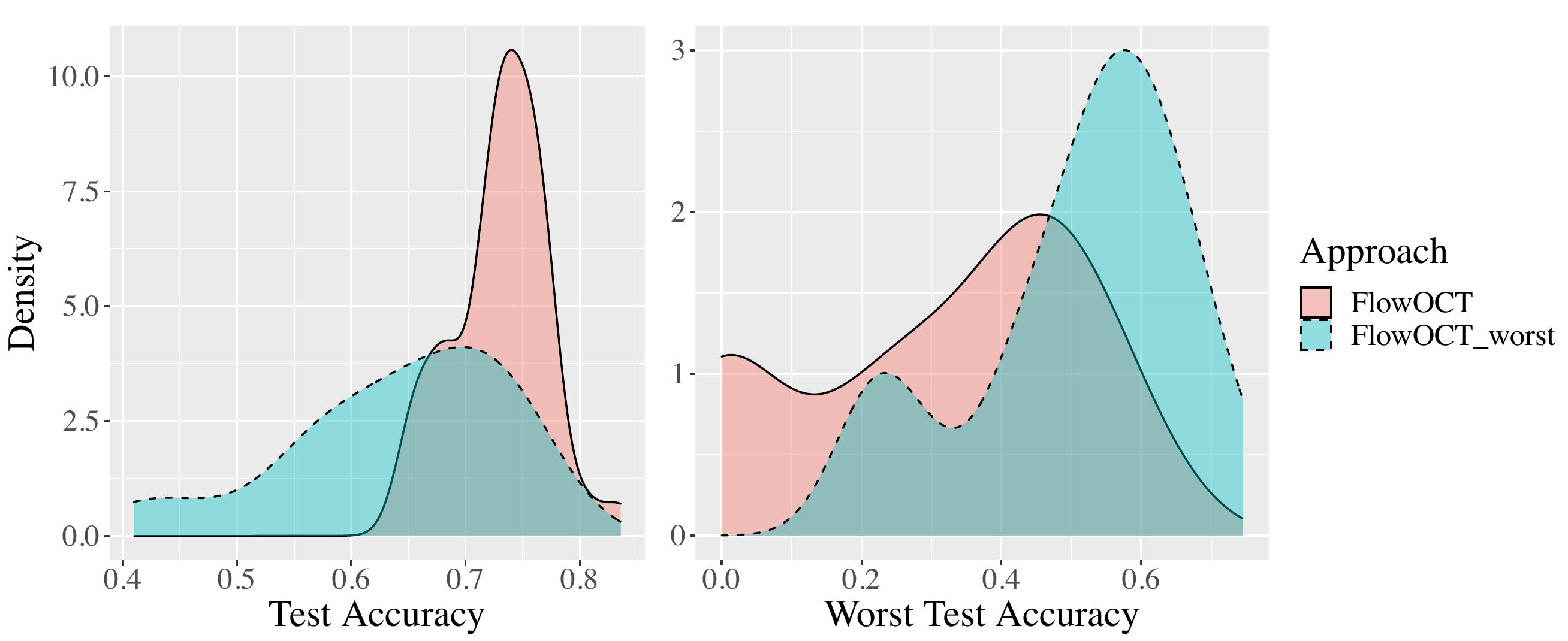}
\caption{The left (resp.\ right) figure depicts the density of out-of-sample accuracy (resp.\ worst accuracy), among all class labels, for each approach.}
\label{fig:flowOCT-worst}
\end{center}
\end{figure}

\begin{figure}[t!]
%\vskip 0.2in
\begin{center}
\includegraphics[width = 0.55\textwidth]{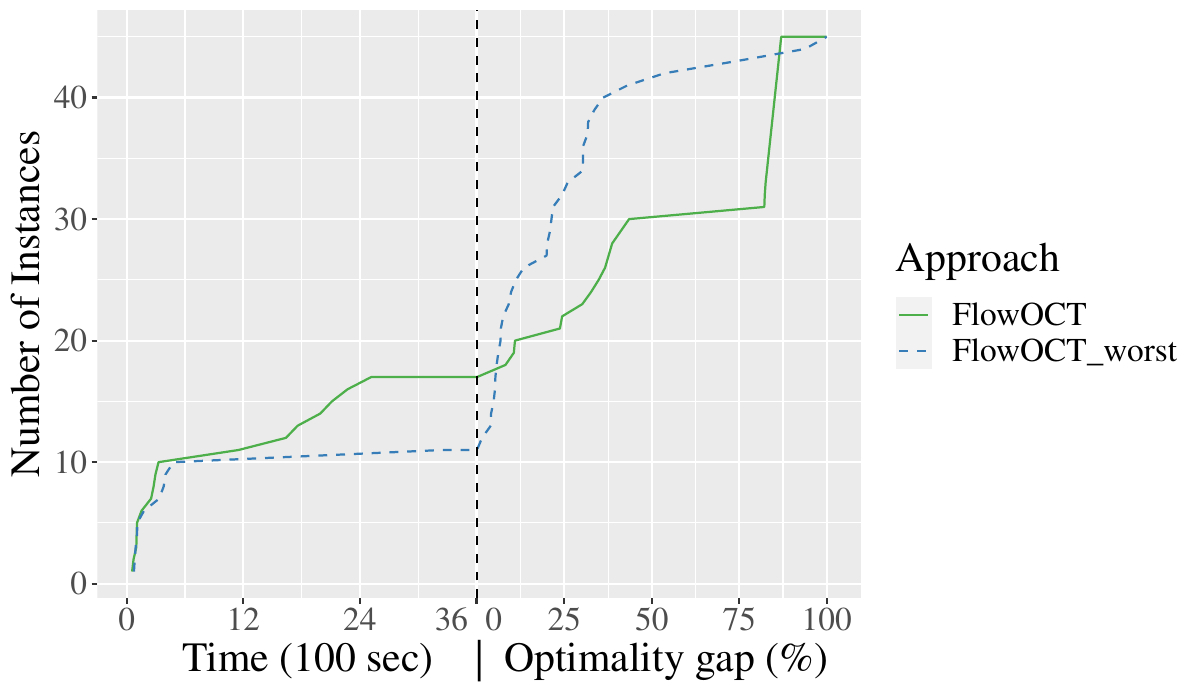}
\caption{Number of instances solved to optimality by each approach within the given time on the time axis, and number of instances with optimality gap no larger than each given value at the time limit on the optimality gap axis.}
\label{fig:flowOCT-worst_2}
\end{center}
\end{figure}

In this section, we study a variant of \flow{}, called~\FlowOCTworst{}, where we maximize the worst-case accuracy objective defined in Section~\ref{sec:imbalanced_datasets}. For this purpose we implement formulation~\eqref{eq:flow_reg_all}. We compare \flow{} and \FlowOCTworst{} on three imbalanced datasets (car-evaluation, spect and breast-cancer).
Figures~\ref{fig:flowOCT-worst} and~\ref{fig:flowOCT-worst_2} summerize the numerical results.
In Figure~\ref{fig:flowOCT-worst} (right) we show the density of the worst out-of-sample accuracy, among all class labels, for each approach. As we can see, \FlowOCTworst{} achieves better worst-case accuracy. However as it is shown in~Figure~\ref{fig:flowOCT-worst} (left), \FlowOCTworst{} has a worst performance in out-of-sample classification accuracy. So there's a trade-off between the total accuracy and worst-case accuracy.
On one hand, from the first part (time axis) of Figure~\ref{fig:flowOCT-worst_2}, we observe that \FlowOCTworst{} can solve 11 instances to optimality within the timelimit. However, \flow{} can solve the same number of instances in 1150 seconds resulting in a $\floor{\frac{3600}{1150}}=3\times$ speedup. On the other hand, from the second part (gap axis) of Figure~\ref{fig:flowOCT-worst_2}, we see that \FlowOCTworst{} tends to have lower optimality gap with respect to \flow{}. 

% \notesa{Isn't that weird the \flow{} can solve more instances to optimality but has larger optimality gap in general?}

\subsection{LO Relaxation}
\label{appendix_sec:LO_relaxation}

In this section we compare the strength, i.e., LO relaxation optimal value, of the various formulations.
For this purpose, for all 2400 instances \revision{involving the categorical datasets}, we solve the LO relaxation of the approaches \flow{}, \texttt{OCT}, and \texttt{BinOCT}. For all the approaches, the objective value reflects number of misclassified datapoints. In this setting, a trivial value for the objective value of the relaxed problem would be zero, i.e., correctly classifying all datapoints. However neither of \texttt{OCT} nor \texttt{BinOCT} outputs a non-trivial objective value. But in $44\%$ of the instances, \flow{} outputs a non-trivial objective value, i.e., an objective value less than or equal to $-1$ and also gives a root improvement of a factor of $8$, where root improvement is defined as the ratio of the MIO objective value to the LO relaxation objective value. 
\revision{Tables~\ref{tab:LO_relaxation_with_regularization} and~\ref{tab:LO_relaxation_no_regularization} report the detailed results.}

Furthermore, we look at the number of branch-and-bound nodes explored by each approach.
For this purpose, we only consider $115$ (out of $240$) instances where all the approaches can solve to optimality. Table~\eqref{tab:explored_nodes} summarizes the distribution of the number of explored branch-and-bound nodes for each approach. As it is shown in Table~\eqref{tab:explored_nodes}, \flow{} and \benders{} find the optimal solution by exploring far less branch-and-bound nodes in comparison to \texttt{OCT} and \texttt{BinOCT}. On average, \flow{} (resp.\ \benders) explores $60\%$ and $98\%$ \revision{(resp.\ $43\%$ and $97\%$)} less nodes in comparison to \texttt{OCT} and \texttt{BinOCT}, respectively.

\begin{table}[]
\OneAndAHalfSpacedXII
\small{
\caption{\revision{Number of branch-and-bound nodes explored by each approach during the solving process of 115 instances (out of 240 instances comprising 12 categorical datasets, 5 samples, and 4 different depths), which were successfully solved to optimality by all approaches.}}
\label{tab:explored_nodes}
\begin{center}
% \resizebox{0.7\textwidth}{!}{%
\begin{tabular}{l||llllll}
\hline
Approach       & Min  & 1st Quarter & Median & Mean     & 3rd Quarter & Max        \\ \hline
FlowOCT        & 0.0  & 129.5       & 681.0  & 4480.0   & 4556.0      & 79175.0    \\ 
BendersOCT     & 1.0  & 251         & 1029.0 & 6566.3   & 4060.0      & 180663.0   \\ 
OCT            & 1.0  & 706.0       & 3234.0 & 11449.0  & 9785.0      & 270265.0   \\
BinOCT         & 60.0 & 1508.0      & 6808.0 & 243229.0 & 45659.0     & 10755902.0 \\ \hline
\end{tabular}%
% }
\end{center}
}
\end{table}

\revision{\begin{table}[ht]
\center{
\OneAndAHalfSpacedXII
\caption{  
\revision{In-sample results of the LO relaxation including the average and standard deviation of objective value, root improvement and solving time across 45 instances (5 samples and 9 values of $\lambda$) for the case of $\lambda > 0$ on categorical datasets. The best performance achieved in a given dataset and depth is reported in \textbf{bold}.}}
\small{
\label{tab:LO_relaxation_with_regularization}
\setlength{\tabcolsep}{4pt}
\scalebox{0.9}{
\color{black}\begin{tabular}{lc||ccc|ccc}
\hline
Dataset &
  Depth &
  \multicolumn{3}{c|}{\texttt{OCT}} &
  \multicolumn{3}{c}{\flow}\\
 &  & Obj Value & Root Improvement & Time  & Obj Value & Root Improvement & Time \\
  \hline
soybean-small&2&  0.00$\pm$0.00& 0.00$\pm$ 0.00&\textbf{   0$\pm$   0}&\textbf{ -1.32$\pm$0.62}&\textbf{ 0.67$\pm$ 0.48}&\textbf{   0$\pm$   0}\\
soybean-small&3&  0.00$\pm$0.00& 0.00$\pm$ 0.00&\textbf{   0$\pm$   0}&\textbf{ -1.32$\pm$0.62}&\textbf{ 0.67$\pm$ 0.48}&\textbf{   0$\pm$   0}\\
soybean-small&4&  0.00$\pm$0.00& 0.00$\pm$ 0.00&\textbf{   0$\pm$   0}&\textbf{ -1.32$\pm$0.62}&\textbf{ 0.67$\pm$ 0.48}&\textbf{   0$\pm$   0}\\
soybean-small&5&  0.00$\pm$0.00& 0.00$\pm$ 0.00&\textbf{   0$\pm$   0}&\textbf{ -1.32$\pm$0.62}&\textbf{ 0.67$\pm$ 0.48}&   2$\pm$   1\\
monk3&2&  0.00$\pm$0.00& 0.00$\pm$ 0.00&\textbf{   0$\pm$   0}&\textbf{ -0.75$\pm$0.39}&\textbf{ 0.65$\pm$ 0.95}&\textbf{   0$\pm$   0}\\
monk3&3&  0.00$\pm$0.00& 0.00$\pm$ 0.00&\textbf{   0$\pm$   0}&\textbf{ -0.75$\pm$0.39}&\textbf{ 0.65$\pm$ 0.95}&\textbf{   0$\pm$   0}\\
monk3&4&  0.00$\pm$0.00& 0.00$\pm$ 0.00&\textbf{   0$\pm$   0}&\textbf{ -0.75$\pm$0.39}&\textbf{ 0.65$\pm$ 0.95}&\textbf{   0$\pm$   0}\\
monk3&5&  0.00$\pm$0.00& 0.00$\pm$ 0.00&\textbf{   0$\pm$   0}&\textbf{ -0.75$\pm$0.39}&\textbf{ 0.65$\pm$ 0.95}&\textbf{   0$\pm$   0}\\
monk1&2&  0.00$\pm$0.00& 0.00$\pm$ 0.00&\textbf{   0$\pm$   0}&\textbf{ -0.75$\pm$0.39}&\textbf{ 1.10$\pm$ 1.70}&\textbf{   0$\pm$   0}\\
monk1&3&  0.00$\pm$0.00& 0.00$\pm$ 0.00&\textbf{   0$\pm$   0}&\textbf{ -0.75$\pm$0.39}&\textbf{ 1.03$\pm$ 1.56}&\textbf{   0$\pm$   0}\\
monk1&4&  0.00$\pm$0.00& 0.00$\pm$ 0.00&\textbf{   0$\pm$   0}&\textbf{ -0.75$\pm$0.39}&\textbf{ 1.01$\pm$ 1.52}&\textbf{   0$\pm$   0}\\
monk1&5&  0.00$\pm$0.00& 0.00$\pm$ 0.00&\textbf{   0$\pm$   0}&\textbf{ -0.75$\pm$0.39}&\textbf{ 1.01$\pm$ 1.52}&\textbf{   0$\pm$   0}\\
hayes-roth&2&  0.00$\pm$0.00& 0.00$\pm$ 0.00&\textbf{   0$\pm$   0}&\textbf{ -1.00$\pm$0.52}&\textbf{ 3.71$\pm$ 4.34}&\textbf{   0$\pm$   0}\\
hayes-roth&3&  0.00$\pm$0.00& 0.00$\pm$ 0.00&\textbf{   0$\pm$   0}&\textbf{ -1.00$\pm$0.52}&\textbf{ 3.18$\pm$ 3.50}&\textbf{   0$\pm$   0}\\
hayes-roth&4&  0.00$\pm$0.00& 0.00$\pm$ 0.00&\textbf{   0$\pm$   0}&\textbf{ -1.00$\pm$0.52}&\textbf{ 2.96$\pm$ 3.17}&\textbf{   0$\pm$   0}\\
hayes-roth&5&  0.00$\pm$0.00& 0.00$\pm$ 0.00&\textbf{   0$\pm$   0}&\textbf{ -1.00$\pm$0.52}&\textbf{ 2.84$\pm$ 2.98}&\textbf{   0$\pm$   0}\\
monk2&2&  0.00$\pm$0.00& 0.00$\pm$ 0.00&\textbf{   0$\pm$   0}&\textbf{ -0.75$\pm$0.39}&\textbf{ 1.83$\pm$ 3.02}&\textbf{   0$\pm$   0}\\
monk2&3&  0.00$\pm$0.00& 0.00$\pm$ 0.00&\textbf{   0$\pm$   0}&\textbf{ -0.75$\pm$0.39}&\textbf{ 1.78$\pm$ 2.90}&\textbf{   0$\pm$   0}\\
monk2&4&  0.00$\pm$0.00& 0.00$\pm$ 0.00&\textbf{   0$\pm$   0}&\textbf{ -0.75$\pm$0.39}&\textbf{ 1.78$\pm$ 2.90}&\textbf{   0$\pm$   0}\\
monk2&5&  0.00$\pm$0.00& 0.00$\pm$ 0.00&\textbf{   0$\pm$   0}&\textbf{ -0.75$\pm$0.39}&\textbf{ 1.79$\pm$ 2.92}&\textbf{   0$\pm$   0}\\
house-votes-84&2&  0.00$\pm$0.00& 0.00$\pm$ 0.00&\textbf{   0$\pm$   0}&\textbf{ -0.81$\pm$0.39}&\textbf{ 0.58$\pm$ 0.73}&\textbf{   0$\pm$   0}\\
house-votes-84&3&  0.00$\pm$0.00& 0.00$\pm$ 0.00&\textbf{   0$\pm$   0}&\textbf{ -0.81$\pm$0.39}&\textbf{ 0.58$\pm$ 0.73}&\textbf{   0$\pm$   0}\\
house-votes-84&4&  0.00$\pm$0.00& 0.00$\pm$ 0.00&\textbf{   0$\pm$   0}&\textbf{ -0.81$\pm$0.39}&\textbf{ 0.58$\pm$ 0.73}&\textbf{   0$\pm$   0}\\
house-votes-84&5&  0.00$\pm$0.00& 0.00$\pm$ 0.00&\textbf{   0$\pm$   0}&\textbf{ -0.81$\pm$0.39}&\textbf{ 0.58$\pm$ 0.73}&\textbf{   0$\pm$   0}\\
spect&2&  0.00$\pm$0.00& 0.00$\pm$ 0.00&\textbf{   0$\pm$   0}&\textbf{ -6.23$\pm$2.24}&\textbf{ 1.96$\pm$ 0.51}&\textbf{   0$\pm$   0}\\
spect&3&  0.00$\pm$0.00& 0.00$\pm$ 0.00&\textbf{   0$\pm$   0}&\textbf{ -4.98$\pm$1.35}&\textbf{ 2.08$\pm$ 0.64}&\textbf{   0$\pm$   0}\\
spect&4&  0.00$\pm$0.00& 0.00$\pm$ 0.00&\textbf{   0$\pm$   0}&\textbf{ -4.41$\pm$0.98}&\textbf{ 1.91$\pm$ 0.51}&   1$\pm$   0\\
spect&5&  0.00$\pm$0.00& 0.00$\pm$ 0.00&\textbf{   0$\pm$   0}&\textbf{ -4.02$\pm$0.90}&\textbf{ 1.89$\pm$ 0.58}&   1$\pm$   0\\
breast-cancer&2&  0.00$\pm$0.00& 0.00$\pm$ 0.00&\textbf{   0$\pm$   0}&\textbf{ -0.65$\pm$0.31}&\textbf{ 0.49$\pm$ 1.66}&\textbf{   0$\pm$   0}\\
breast-cancer&3&  0.00$\pm$0.00& 0.00$\pm$ 0.00&\textbf{   0$\pm$   0}&\textbf{ -0.65$\pm$0.31}&\textbf{ 0.64$\pm$ 1.89}&\textbf{   0$\pm$   0}\\
breast-cancer&4&  0.00$\pm$0.00& 0.00$\pm$ 0.00&\textbf{   0$\pm$   0}&\textbf{ -0.65$\pm$0.31}&\textbf{ 0.49$\pm$ 1.66}&   2$\pm$   1\\
breast-cancer&5&  0.00$\pm$0.00& 0.00$\pm$ 0.00&\textbf{   0$\pm$   0}&\textbf{ -0.65$\pm$0.31}&\textbf{ 0.49$\pm$ 1.66}&   3$\pm$   1\\
balance-scale&2&  0.00$\pm$0.00& 0.00$\pm$ 0.00&\textbf{   0$\pm$   0}&\textbf{-19.80$\pm$9.32}&\textbf{ 2.37$\pm$ 0.18}&\textbf{   0$\pm$   0}\\
balance-scale&3&  0.00$\pm$0.00& 0.00$\pm$ 0.00&\textbf{   0$\pm$   0}&\textbf{ -3.92$\pm$0.54}&\textbf{ 9.66$\pm$ 3.54}&   2$\pm$   0\\
balance-scale&4&  0.00$\pm$0.00& 0.00$\pm$ 0.00&\textbf{   0$\pm$   0}&\textbf{ -1.67$\pm$0.87}&\textbf{13.38$\pm$14.04}&  17$\pm$   5\\
balance-scale&5&  0.00$\pm$0.00& 0.00$\pm$ 0.00&\textbf{   1$\pm$   0}&\textbf{ -1.67$\pm$0.87}&\textbf{10.67$\pm$11.32}&  21$\pm$  15\\
tic-tac-toe&2&  0.00$\pm$0.00& 0.00$\pm$ 0.00&\textbf{   0$\pm$   0}&\textbf{ -0.71$\pm$0.37}&\textbf{ 4.01$\pm$ 8.19}&   1$\pm$   0\\
tic-tac-toe&3&  0.00$\pm$0.00& 0.00$\pm$ 0.00&\textbf{   0$\pm$   0}&\textbf{ -0.71$\pm$0.37}&\textbf{ 3.65$\pm$ 7.36}&   5$\pm$   1\\
tic-tac-toe&4&  0.00$\pm$0.00& 0.00$\pm$ 0.00&\textbf{   1$\pm$   0}&\textbf{ -0.71$\pm$0.37}&\textbf{ 3.62$\pm$ 7.23}&  18$\pm$  13\\
tic-tac-toe&5&  0.00$\pm$0.00& 0.00$\pm$ 0.00&\textbf{   2$\pm$   0}&\textbf{ -0.71$\pm$0.37}&\textbf{ 3.68$\pm$ 7.39}&  14$\pm$  13\\
car-evaluation&2&  0.00$\pm$0.00& 0.00$\pm$ 0.00&\textbf{   0$\pm$   0}&\textbf{ -9.09$\pm$3.26}&\textbf{ 9.68$\pm$ 2.30}&   2$\pm$   0\\
car-evaluation&3&  0.00$\pm$0.00& 0.00$\pm$ 0.00&\textbf{   0$\pm$   0}&\textbf{ -1.50$\pm$0.78}&\textbf{24.05$\pm$26.93}&   9$\pm$   2\\
car-evaluation&4&  0.00$\pm$0.00& 0.00$\pm$ 0.00&\textbf{   1$\pm$   0}&\textbf{ -1.50$\pm$0.78}&\textbf{24.45$\pm$27.69}& 104$\pm$  27\\
car-evaluation&5&  0.00$\pm$0.00& 0.00$\pm$ 0.00&\textbf{   3$\pm$   0}&\textbf{ -1.50$\pm$0.78}&\textbf{27.34$\pm$31.74}&1219$\pm$ 230\\
kr-vs-kp&2&  0.00$\pm$0.00&\textbf{ 0.00$\pm$ 0.00}&\textbf{   1$\pm$   0}&\textbf{ -0.51$\pm$0.26}&\textbf{ 0.00$\pm$ 0.00}&  18$\pm$   5\\
kr-vs-kp&3&  0.00$\pm$0.00&\textbf{ 0.00$\pm$ 0.00}&\textbf{   2$\pm$   0}&\textbf{ -0.51$\pm$0.26}&\textbf{ 0.00$\pm$ 0.00}& 186$\pm$  97\\
kr-vs-kp&4&  0.00$\pm$0.00&\textbf{ 0.00$\pm$ 0.00}&\textbf{   6$\pm$   2}&\textbf{ -0.51$\pm$0.26}&\textbf{ 0.00$\pm$ 0.00}&1520$\pm$1076\\
kr-vs-kp&5&  0.00$\pm$0.00&\textbf{ 0.00$\pm$ 0.00}&\textbf{  16$\pm$   2}&\textbf{-Inf}&\textbf{ 0.00$\pm$ 0.00}&1557$\pm$1516\\ \hline
\end{tabular}%
}
}
}
\end{table}

\begin{landscape}
\begin{table}
\OneAndAHalfSpacedXII
\caption{\revision{In-sample results of the LO relaxation including the average and standard deviation of objective value, root improvement and solving time across 
5 samples for the case of $\lambda=0$ on categorical datasets. The best performance achieved in a given dataset and depth is reported in \textbf{bold}.}}
\small{
\label{tab:LO_relaxation_no_regularization}
\setlength{\tabcolsep}{4pt}
% \resizebox{1.3\textheight}{!}{%
\scalebox{0.8}{
\color{black}\begin{tabular}{lc||ccc|ccc|ccc}
\hline
Dataset &
  Depth &
  \multicolumn{3}{c|}{\texttt{OCT}} &
  \multicolumn{3}{c|}{\texttt{BinOCT}} &
  \multicolumn{3}{c}{\flow} \\
 &  & Obj Value & Root Improvement & Time  & Obj Value & Root Improvement & Time & Obj Value & Root Improvement \\
  \hline
  soybean-small&2&\textbf{  0.00$\pm$0.00}&\textbf{ 0.00$\pm$0.00}&\textbf{  0$\pm$  0}&\textbf{  0.00$\pm$0.00}&\textbf{ 0.00$\pm$0.00}&\textbf{  0$\pm$  0}&\textbf{  0.00$\pm$0.00}&\textbf{ 0.00$\pm$0.00}&\textbf{  0$\pm$  0}\\
  soybean-small&3&\textbf{  0.00$\pm$0.00}&\textbf{ 0.00$\pm$0.00}&\textbf{  0$\pm$  0}&\textbf{  0.00$\pm$0.00}&\textbf{ 0.00$\pm$0.00}&\textbf{  0$\pm$  0}&\textbf{  0.00$\pm$0.00}&\textbf{ 0.00$\pm$0.00}&\textbf{  0$\pm$  0}\\
  soybean-small&4&\textbf{  0.00$\pm$0.00}&\textbf{ 0.00$\pm$0.00}&\textbf{  0$\pm$  0}&\textbf{  0.00$\pm$0.00}&\textbf{ 0.00$\pm$0.00}&\textbf{  0$\pm$  0}&\textbf{  0.00$\pm$0.00}&\textbf{ 0.00$\pm$0.00}&\textbf{  0$\pm$  0}\\
  soybean-small&5&\textbf{  0.00$\pm$0.00}&\textbf{ 0.00$\pm$0.00}&\textbf{  0$\pm$  0}&\textbf{  0.00$\pm$0.00}&\textbf{ 0.00$\pm$0.00}&\textbf{  0$\pm$  0}&\textbf{  0.00$\pm$0.00}&\textbf{ 0.00$\pm$0.00}&\textbf{  0$\pm$  0}\\
  monk3&2&\textbf{  0.00$\pm$0.00}&\textbf{ 0.00$\pm$0.00}&\textbf{  0$\pm$  0}&\textbf{  0.00$\pm$0.00}&\textbf{ 0.00$\pm$0.00}&\textbf{  0$\pm$  0}&\textbf{  0.00$\pm$0.00}&\textbf{ 0.00$\pm$0.00}&\textbf{  0$\pm$  0}\\
  monk3&3&\textbf{  0.00$\pm$0.00}&\textbf{ 0.00$\pm$0.00}&\textbf{  0$\pm$  0}&\textbf{  0.00$\pm$0.00}&\textbf{ 0.00$\pm$0.00}&\textbf{  0$\pm$  0}&\textbf{  0.00$\pm$0.00}&\textbf{ 0.00$\pm$0.00}&\textbf{  0$\pm$  0}\\
  monk3&4&\textbf{  0.00$\pm$0.00}&\textbf{ 0.00$\pm$0.00}&\textbf{  0$\pm$  0}&\textbf{  0.00$\pm$0.00}&\textbf{ 0.00$\pm$0.00}&\textbf{  0$\pm$  0}&\textbf{  0.00$\pm$0.00}&\textbf{ 0.00$\pm$0.00}&\textbf{  0$\pm$  0}\\
  monk3&5&\textbf{  0.00$\pm$0.00}&\textbf{ 0.00$\pm$0.00}&\textbf{  0$\pm$  0}&\textbf{  0.00$\pm$0.00}&\textbf{ 0.00$\pm$0.00}&\textbf{  0$\pm$  0}&\textbf{  0.00$\pm$0.00}&\textbf{ 0.00$\pm$0.00}&\textbf{  0$\pm$  0}\\
  monk1&2&\textbf{  0.00$\pm$0.00}&\textbf{ 0.00$\pm$0.00}&\textbf{  0$\pm$  0}&\textbf{  0.00$\pm$0.00}&\textbf{ 0.00$\pm$0.00}&\textbf{  0$\pm$  0}&\textbf{  0.00$\pm$0.00}&\textbf{ 0.00$\pm$0.00}&\textbf{  0$\pm$  0}\\
  monk1&3&\textbf{  0.00$\pm$0.00}&\textbf{ 0.00$\pm$0.00}&\textbf{  0$\pm$  0}&\textbf{  0.00$\pm$0.00}&\textbf{ 0.00$\pm$0.00}&\textbf{  0$\pm$  0}&\textbf{  0.00$\pm$0.00}&\textbf{ 0.00$\pm$0.00}&\textbf{  0$\pm$  0}\\
  monk1&4&\textbf{  0.00$\pm$0.00}&\textbf{ 0.00$\pm$0.00}&\textbf{  0$\pm$  0}&\textbf{  0.00$\pm$0.00}&\textbf{ 0.00$\pm$0.00}&\textbf{  0$\pm$  0}&\textbf{  0.00$\pm$0.00}&\textbf{ 0.00$\pm$0.00}&\textbf{  0$\pm$  0}\\
  monk1&5&\textbf{  0.00$\pm$0.00}&\textbf{ 0.00$\pm$0.00}&\textbf{  0$\pm$  0}&\textbf{  0.00$\pm$0.00}&\textbf{ 0.00$\pm$0.00}&\textbf{  0$\pm$  0}&\textbf{  0.00$\pm$0.00}&\textbf{ 0.00$\pm$0.00}&\textbf{  0$\pm$  0}\\
  hayes-roth&2&\textbf{  0.00$\pm$0.00}&\textbf{ 0.00$\pm$0.00}&\textbf{  0$\pm$  0}&\textbf{  0.00$\pm$0.00}&\textbf{ 0.00$\pm$0.00}&\textbf{  0$\pm$  0}&\textbf{  0.00$\pm$0.00}&\textbf{ 0.00$\pm$0.00}&\textbf{  0$\pm$  0}\\
  hayes-roth&3&\textbf{  0.00$\pm$0.00}&\textbf{ 0.00$\pm$0.00}&\textbf{  0$\pm$  0}&\textbf{  0.00$\pm$0.00}&\textbf{ 0.00$\pm$0.00}&\textbf{  0$\pm$  0}&\textbf{  0.00$\pm$0.00}&\textbf{ 0.00$\pm$0.00}&\textbf{  0$\pm$  0}\\
  hayes-roth&4&\textbf{  0.00$\pm$0.00}&\textbf{ 0.00$\pm$0.00}&\textbf{  0$\pm$  0}&\textbf{  0.00$\pm$0.00}&\textbf{ 0.00$\pm$0.00}&\textbf{  0$\pm$  0}&\textbf{  0.00$\pm$0.00}&\textbf{ 0.00$\pm$0.00}&\textbf{  0$\pm$  0}\\
  hayes-roth&5&\textbf{  0.00$\pm$0.00}&\textbf{ 0.00$\pm$0.00}&\textbf{  0$\pm$  0}&\textbf{  0.00$\pm$0.00}&\textbf{ 0.00$\pm$0.00}&\textbf{  0$\pm$  0}&\textbf{  0.00$\pm$0.00}&\textbf{ 0.00$\pm$0.00}&\textbf{  0$\pm$  0}\\
  monk2&2&\textbf{  0.00$\pm$0.00}&\textbf{ 0.00$\pm$0.00}&\textbf{  0$\pm$  0}&\textbf{  0.00$\pm$0.00}&\textbf{ 0.00$\pm$0.00}&\textbf{  0$\pm$  0}&\textbf{  0.00$\pm$0.00}&\textbf{ 0.00$\pm$0.00}&\textbf{  0$\pm$  0}\\
  monk2&3&\textbf{  0.00$\pm$0.00}&\textbf{ 0.00$\pm$0.00}&\textbf{  0$\pm$  0}&\textbf{  0.00$\pm$0.00}&\textbf{ 0.00$\pm$0.00}&\textbf{  0$\pm$  0}&\textbf{  0.00$\pm$0.00}&\textbf{ 0.00$\pm$0.00}&\textbf{  0$\pm$  0}\\
  monk2&4&\textbf{  0.00$\pm$0.00}&\textbf{ 0.00$\pm$0.00}&\textbf{  0$\pm$  0}&\textbf{  0.00$\pm$0.00}&\textbf{ 0.00$\pm$0.00}&\textbf{  0$\pm$  0}&\textbf{  0.00$\pm$0.00}&\textbf{ 0.00$\pm$0.00}&\textbf{  0$\pm$  0}\\
  monk2&5&\textbf{  0.00$\pm$0.00}&\textbf{ 0.00$\pm$0.00}&\textbf{  0$\pm$  0}&\textbf{  0.00$\pm$0.00}&\textbf{ 0.00$\pm$0.00}&\textbf{  0$\pm$  0}&\textbf{  0.00$\pm$0.00}&\textbf{ 0.00$\pm$0.00}&\textbf{  0$\pm$  0}\\
  house-votes-84&2&\textbf{  0.00$\pm$0.00}&\textbf{ 0.00$\pm$0.00}&\textbf{  0$\pm$  0}&\textbf{  0.00$\pm$0.00}&\textbf{ 0.00$\pm$0.00}&\textbf{  0$\pm$  0}&\textbf{  0.00$\pm$0.00}&\textbf{ 0.00$\pm$0.00}&\textbf{  0$\pm$  0}\\
  house-votes-84&3&\textbf{  0.00$\pm$0.00}&\textbf{ 0.00$\pm$0.00}&\textbf{  0$\pm$  0}&\textbf{  0.00$\pm$0.00}&\textbf{ 0.00$\pm$0.00}&\textbf{  0$\pm$  0}&\textbf{  0.00$\pm$0.00}&\textbf{ 0.00$\pm$0.00}&\textbf{  0$\pm$  0}\\
  house-votes-84&4&\textbf{  0.00$\pm$0.00}&\textbf{ 0.00$\pm$0.00}&\textbf{  0$\pm$  0}&\textbf{  0.00$\pm$0.00}&\textbf{ 0.00$\pm$0.00}&\textbf{  0$\pm$  0}&\textbf{  0.00$\pm$0.00}&\textbf{ 0.00$\pm$0.00}&  1$\pm$  0\\
  house-votes-84&5&\textbf{  0.00$\pm$0.00}&\textbf{ 0.00$\pm$0.00}&\textbf{  0$\pm$  0}&\textbf{  0.00$\pm$0.00}&\textbf{ 0.00$\pm$0.00}&\textbf{  0$\pm$  0}&\textbf{  0.00$\pm$0.00}&\textbf{ 0.00$\pm$0.00}&  1$\pm$  0\\
  spect&2&  0.00$\pm$0.00& 0.00$\pm$0.00&\textbf{  0$\pm$  0}&  0.00$\pm$0.00& 0.00$\pm$0.00&\textbf{  0$\pm$  0}&\textbf{-10.26$\pm$1.04}&\textbf{ 2.51$\pm$0.28}&\textbf{  0$\pm$  0}\\
  spect&3&  0.00$\pm$0.00& 0.00$\pm$0.00&\textbf{  0$\pm$  0}&  0.00$\pm$0.00& 0.00$\pm$0.00&\textbf{  0$\pm$  0}&\textbf{ -7.00$\pm$1.20}&\textbf{ 2.72$\pm$0.55}&\textbf{  0$\pm$  0}\\
  spect&4&  0.00$\pm$0.00& 0.00$\pm$0.00&\textbf{  0$\pm$  0}&  0.00$\pm$0.00& 0.00$\pm$0.00&\textbf{  0$\pm$  0}&\textbf{ -5.17$\pm$1.26}&\textbf{ 2.41$\pm$0.66}&  1$\pm$  0\\
  spect&5&  0.00$\pm$0.00& 0.00$\pm$0.00&\textbf{  0$\pm$  0}&  0.00$\pm$0.00& 0.00$\pm$0.00&\textbf{  0$\pm$  0}&\textbf{ -3.76$\pm$1.40}&\textbf{ 2.22$\pm$1.07}&  3$\pm$  1\\
  breast-cancer&2&\textbf{  0.00$\pm$0.00}&\textbf{ 0.00$\pm$0.00}&\textbf{  0$\pm$  0}&\textbf{  0.00$\pm$0.00}&\textbf{ 0.00$\pm$0.00}&\textbf{  0$\pm$  0}&\textbf{  0.00$\pm$0.00}&\textbf{ 0.00$\pm$0.00}&\textbf{  0$\pm$  0}\\
  breast-cancer&3&\textbf{  0.00$\pm$0.00}&\textbf{ 0.00$\pm$0.00}&\textbf{  0$\pm$  0}&\textbf{  0.00$\pm$0.00}&\textbf{ 0.00$\pm$0.00}&\textbf{  0$\pm$  0}&\textbf{  0.00$\pm$0.00}&\textbf{ 0.00$\pm$0.00}&  1$\pm$  0\\
  breast-cancer&4&\textbf{  0.00$\pm$0.00}&\textbf{ 0.00$\pm$0.00}&\textbf{  0$\pm$  0}&\textbf{  0.00$\pm$0.00}&\textbf{ 0.00$\pm$0.00}&\textbf{  0$\pm$  0}&\textbf{  0.00$\pm$0.00}&\textbf{ 0.00$\pm$0.00}&  4$\pm$  1\\
  breast-cancer&5&\textbf{  0.00$\pm$0.00}&\textbf{ 0.00$\pm$0.00}&\textbf{  0$\pm$  0}&\textbf{  0.00$\pm$0.00}&\textbf{ 0.00$\pm$0.00}&\textbf{  0$\pm$  0}&\textbf{  0.00$\pm$0.00}&\textbf{ 0.00$\pm$0.00}&  4$\pm$  4\\
  balance-scale&2&  0.00$\pm$0.00& 0.00$\pm$0.00&\textbf{  0$\pm$  0}&  0.00$\pm$0.00& 0.00$\pm$0.00&\textbf{  0$\pm$  0}&\textbf{-37.60$\pm$1.39}&\textbf{ 2.48$\pm$0.15}&\textbf{  0$\pm$  0}\\
  balance-scale&3&  0.00$\pm$0.00& 0.00$\pm$0.00&\textbf{  0$\pm$  0}&  0.00$\pm$0.00& 0.00$\pm$0.00&\textbf{  0$\pm$  0}&\textbf{ -4.84$\pm$0.50}&\textbf{15.20$\pm$0.97}&  2$\pm$  1\\
  balance-scale&4&\textbf{  0.00$\pm$0.00}&\textbf{ 0.00$\pm$0.00}&\textbf{  0$\pm$  0}&\textbf{  0.00$\pm$0.00}&\textbf{ 0.00$\pm$0.00}&\textbf{  0$\pm$  0}&\textbf{  0.00$\pm$0.00}&\textbf{ 0.00$\pm$0.00}& 41$\pm$  4\\
  balance-scale&5&\textbf{  0.00$\pm$0.00}&\textbf{ 0.00$\pm$0.00}&  1$\pm$  0&\textbf{  0.00$\pm$0.00}&\textbf{ 0.00$\pm$0.00}&\textbf{  0$\pm$  0}&\textbf{  0.00$\pm$0.00}&\textbf{ 0.00$\pm$0.00}& 20$\pm$  3\\
  tic-tac-toe&2&\textbf{  0.00$\pm$0.00}&\textbf{ 0.00$\pm$0.00}&\textbf{  0$\pm$  0}&\textbf{  0.00$\pm$0.00}&\textbf{ 0.00$\pm$0.00}&\textbf{  0$\pm$  0}&\textbf{  0.00$\pm$0.00}&\textbf{ 0.00$\pm$0.00}&  1$\pm$  0\\
  tic-tac-toe&3&\textbf{  0.00$\pm$0.00}&\textbf{ 0.00$\pm$0.00}&\textbf{  0$\pm$  0}&\textbf{  0.00$\pm$0.00}&\textbf{ 0.00$\pm$0.00}&\textbf{  0$\pm$  0}&\textbf{  0.00$\pm$0.00}&\textbf{ 0.00$\pm$0.00}& 10$\pm$  2\\
  tic-tac-toe&4&\textbf{  0.00$\pm$0.00}&\textbf{ 0.00$\pm$0.00}&  1$\pm$  0&\textbf{  0.00$\pm$0.00}&\textbf{ 0.00$\pm$0.00}&\textbf{  0$\pm$  0}&\textbf{  0.00$\pm$0.00}&\textbf{ 0.00$\pm$0.00}& 29$\pm$ 42\\
  tic-tac-toe&5&\textbf{  0.00$\pm$0.00}&\textbf{ 0.00$\pm$0.00}&  2$\pm$  0&\textbf{  0.00$\pm$0.00}&\textbf{ 0.00$\pm$0.00}&\textbf{  1$\pm$  0}&\textbf{  0.00$\pm$0.00}&\textbf{ 0.00$\pm$0.00}& 20$\pm$ 12\\
  car-evaluation&2&  0.00$\pm$0.00& 0.00$\pm$0.00&\textbf{  0$\pm$  0}&  0.00$\pm$0.00& 0.00$\pm$0.00&\textbf{  0$\pm$  0}&\textbf{-15.20$\pm$1.15}&\textbf{12.34$\pm$0.71}&  2$\pm$  0\\
  car-evaluation&3&\textbf{  0.00$\pm$0.00}&\textbf{ 0.00$\pm$0.00}&\textbf{  0$\pm$  0}&\textbf{  0.00$\pm$0.00}&\textbf{ 0.00$\pm$0.00}&\textbf{  0$\pm$  0}&\textbf{  0.00$\pm$0.00}&\textbf{ 0.00$\pm$0.00}&  5$\pm$  1\\
  car-evaluation&4&\textbf{  0.00$\pm$0.00}&\textbf{ 0.00$\pm$0.00}&\textbf{  1$\pm$  0}&\textbf{  0.00$\pm$0.00}&\textbf{ 0.00$\pm$0.00}&\textbf{  1$\pm$  0}&\textbf{  0.00$\pm$0.00}&\textbf{ 0.00$\pm$0.00}& 16$\pm$  5\\
  car-evaluation&5&\textbf{  0.00$\pm$0.00}&\textbf{ 0.00$\pm$0.00}&  2$\pm$  0&\textbf{  0.00$\pm$0.00}&\textbf{ 0.00$\pm$0.00}&\textbf{  1$\pm$  0}&\textbf{  0.00$\pm$0.00}&\textbf{ 0.00$\pm$0.00}&233$\pm$ 71\\
  kr-vs-kp&2&\textbf{  0.00$\pm$0.00}&\textbf{ 0.00$\pm$0.00}&  1$\pm$  0&\textbf{  0.00$\pm$0.00}&\textbf{ 0.00$\pm$0.00}&\textbf{  0$\pm$  0}&\textbf{  0.00$\pm$0.00}&\textbf{ 0.00$\pm$0.00}&  4$\pm$  1\\
  kr-vs-kp&3&\textbf{  0.00$\pm$0.00}&\textbf{ 0.00$\pm$0.00}&  2$\pm$  0&\textbf{  0.00$\pm$0.00}&\textbf{ 0.00$\pm$0.00}&\textbf{  1$\pm$  0}&\textbf{  0.00$\pm$0.00}&\textbf{ 0.00$\pm$0.00}&188$\pm$148\\
  kr-vs-kp&4&\textbf{  0.00$\pm$0.00}&\textbf{ 0.00$\pm$0.00}&  6$\pm$  0&\textbf{  0.00$\pm$0.00}&\textbf{ 0.00$\pm$0.00}&\textbf{  2$\pm$  1}&\textbf{  0.00$\pm$0.00}&\textbf{ 0.00$\pm$0.00}&115$\pm$113\\
  kr-vs-kp&5&\textbf{  0.00$\pm$0.00}&\textbf{ 0.00$\pm$0.00}& 16$\pm$  2&\textbf{  0.00$\pm$0.00}&\textbf{ 0.00$\pm$0.00}&\textbf{  6$\pm$  2}&\textbf{  0.00$\pm$0.00}&\textbf{ 0.00$\pm$0.00}&210$\pm$133\\ \hline

\end{tabular}%
}
}
\end{table}
\end{landscape}
}

\end{document}